\documentclass[10pt,letterpaper]{article}
\usepackage[top=0.85in,left=2.75in,footskip=0.75in]{geometry}

\usepackage{fixltx2e}
\usepackage{stfloats}
\usepackage{subfloat}
\usepackage[caption=false,font=footnotesize]{subfig}
\usepackage{caption}
\usepackage{amsmath,amssymb}
\usepackage{algorithm,algorithmicx,algpseudocode}
\usepackage{multirow}
\usepackage{footnote}
\usepackage{changepage}

\usepackage[utf8x]{inputenc}

\usepackage{textcomp,marvosym}

\usepackage{cite}

\usepackage{nameref,hyperref}

\usepackage[right]{lineno}

\usepackage{microtype}
\DisableLigatures[f]{encoding = *, family = * }

\usepackage[table]{xcolor}
\usepackage{booktabs}
\usepackage{array}

\newcolumntype{+}{!{\vrule width 2pt}}

\newlength\savedwidth

\raggedright
\usepackage[aboveskip=1pt,labelfont=bf,labelsep=period,justification=raggedright,singlelinecheck=off]{caption}
\renewcommand{\figurename}{Fig}

\makeatletter
\renewcommand{\@biblabel}[1]{\quad#1.}
\makeatother

\usepackage{lastpage,fancyhdr,graphicx}
\usepackage{epstopdf}
\graphicspath{{./Figures/}}
\fancyheadoffset[L]{2.25in}
\fancyfootoffset[L]{2.25in}
\newcommand{\etal}{\textit{et al.}}

\begin{document}
\vspace*{0.2in}

\begin{flushleft}
{\Large
\textbf{{Simulation-based multi-criteria comparison of mono-articular and bi-articular exoskeletons during walking with and without load}} 
}
\newline
\\
Ali KhalilianMotamed Bonab\textsuperscript{1*},
Volkan Patoglu\textsuperscript{1*}
\\
\bigskip
\textsuperscript{\textbf{1}}Faculty of Engineering and Natural Sciences, Sabanc{\i} University, \.{I}stanbul, Turkey.
\\
*\{alik, volkan.patoglu\}@sabanciuniv.edu

\end{flushleft}


\section*{Abstract}
Wearable robotic assistive devices possess the potential to improve the metabolic efficiency of human locomotion. Developing exoskeletons that can reduce the metabolic cost of assisted subjects is challenging, since a systematic design approach is required to capture the effects of device dynamics and the assistance torques on human performance. Conducting such investigations through human subject experiments with physical devices is generally infeasible. On the other hand, design studies that rely on musculoskeletal models hold high promise in providing effective design guidelines, as the effect of various devices and different assistance torque profiles on muscle recruitment and metabolic cost can be studied systematically. In this paper, we present a simulation-based multi-criteria design approach to systematically study the effect of different device kinematics and corresponding optimal assistive torque profiles under actuator saturation on the metabolic cost, muscle activation, and joint reaction forces of subjects walking under different loading conditions. For the multi-criteria comparison of mono-articular and bi-articular exoskeletons, we introduce a Pareto optimization approach to simultaneously optimize the exoskeleton power consumption and the human metabolic rate reduction during walking, under different loading conditions. We further superpose the effects of device inertia and electrical regeneration on the metabolic rate and power consumption, respectively.  Our simulation results explain the effects of heavy loads on the optimal assistance profiles of the exoskeletons and provide guidelines on choosing optimal device configurations under actuator torque limitations, device inertia, and regeneration effects. The multi-criteria comparison of devices indicates that despite the similar assistance levels that can be provided by both types of exoskeletons, mono-articular exoskeletons demonstrate better performance on reducing the peak reaction forces, while the power consumption of bi-articular exoskeletons is less sensitive to the loading. Furthermore, for the bi-articular exoskeletons, the device inertia has lower detrimental effects on the metabolic cost of subjects and does not affect Pareto-optimality of solutions, while non-dominated configurations are significantly affected by the device inertia for the mono-articular exoskeletons.

\bigskip
\section*{Introduction}
The versatility and bipedalism of human locomotion are both unique and important characteristics of humans among the mammalians~\cite{Rodman1980}. While bipedal locomotion has a low energy cost of transport~\cite{Rodman1980}, the human musculoskeletal system is not optimal for performing all locomotion tasks~\cite{Uchida2016_idealexo_running}.

Bipedal locomotion can lose its efficiency through aging, disease and injury, which can profoundly affect the quality of life due to a loss of independence and mobility~\cite{Schalock2004}. Even though training can improve the efficiency of locomotion by increasing the stiffness of the tendons~\cite{Kubo2002}, and rehabilitation can help patients to achieve near-normal locomotion  in the long term~\cite{Duncan2011}, the muscle and tendon tissues fundamentally constrain the dynamic properties of the muscles. Musculoskeletal system compromises between enhancing the efficiency of the desired task and its adaptability~\cite{Uchida2016_idealexo_running}. Persistence of neuromotor deficits even after the rehabilitation constrains patients from completely resolving their gait issues and reaching complete independence~\cite{Duncan2011}. By taking advantage of assistive devices that are not bounded by any fundamental biological limitations, a musculotendon system can be customized to increase the efficiency of performing locomotion tasks. Assistive devices can be used for patients to improve their quality of life by recovering their normal gait pattern and decreasing their dependency. Assistive devices can also be employed to reduce the risk of injuries due to repetitive tasks, such as walking with heavy loads~\cite{Ruby2003,Knapik2004}.

The main objective of many lower extremity assistive devices is  to reduce the metabolic cost of locomotion and, in particular, to decrease the metabolic energy required for walking and/or running. Although efforts on designing exoskeletons to reach this goal were initiated decades ago, the researchers have succeeded in accomplishing this goal relatively recently. In 2013, Malcolm~\etal~\cite{Malcolm2013} reported that a 6\%~$\pm$~2\% metabolic cost reduction was achieved by a tethered ankle exoskeleton. Later, Mooney~\etal~\cite{Mooney2014_a} and Collins~\etal~\cite{Collins2015} reported that  their untethered ankle exoskeletons reduced the metabolic energy consumption during walking by 8\% and 7.2\%, respectively. The metabolic cost reduction for running was achieved by Lee~\etal~\cite{Lee2017_softexo_running} using a tethered hip exoskeleton in which they were able to improve the metabolic cost by 5.4\% using simulation-optimized assistance torque profiles.

Several assistive devices have improved the metabolic cost of locomotion since these pioneering studies~\cite{Sawicki2020}. Most of these devices assist the hip or the ankle joints, with the exception of an untethered active knee exoskeleton in~\cite{Mhairi2019}, which has been reported to reduce the metabolic burden of incline walking by 4.2\% while carrying loads. Among these devices, tethered exoskeletons have been reported to improve metabolic burden by 5.4\%--17.4\%, while untethered devices reduced the metabolic cost by 3.3\%--19.8\%. Hip exoskeletons showed superior performance compared to ankle devices:  the metabolic cost improvements of tethered and untethered hip devices were reported as 17.4\% and 19.8\%, respectively, while tethered and untethered ankle exoskeleton improved the walking efficiency by 12\% and 11\%. All devices that have been reported to improve the metabolic cost of human running were developed for the hip joint: passive hip devices reduced the metabolic burden by 6.4\%--8\%, while active hip devices reduced metabolic cost by 3.9\%--5.4\%, respectively~\cite{Sawicki2020}. In addition to assisting healthy subjects, some assistive devices have been successfully employed in improving the walking economy of elderly~\cite{Kim2018_exo_elders,H_Lee_hipexo_elders} and patients with gait abnormality~\cite{Awad2017}. For instance, in~\cite{Awad2017},  a 32\%~$\pm$~9\% metabolic cost reduction is reported for post-stroke patients during walking using a tethered exosuit. Readers are referred to comprehensive survey papers for the details of these studies~\cite{Sawicki2020,Dollar2008,Young2017}.

Despite the progress that has been made on designing assistive exoskeletons, there exists no systematic mechatronic design approach for exoskeletons to improve metabolic cost of human locomotion. A systematic design approach necessitates a means for a rigorous and fair comparison of the effects of different exoskeleton designs and assistance torque profiles on the human metabolic cost of locomotion. Studies based on comprehensive human subject experiments to compare a large range of exoskeleton designs are generally not feasible. Such human-in-the-loop studies are challenging as multiple device prototypes need to be implemented,  appropriate volunteers need to be recruited, the safety of trials need to be established and a sufficient number of trials need to be administered to achieve statistically reliable results. Furthermore, inability to collect certain data without introducing sensors inside the body~\cite{Uchida2016_idealexo_running}, difficulty or impossibility of some measurements~\cite{Seth2018}, and effects of training and fatigue on the performance of subjects~\cite{Selinger2015,Gordon2007} pose as some other important limitations of human-in-the-loop studies.

Simulation-based studies to design assistive devices can complement the experimental design approach  to overcome many of these challenges. Studying assistive devices through musculoskeletal simulations facilitates the research by reducing the need for physical prototyping and enables researchers to carry out fast, automated, and repeatable studies in a controlled environment, where different analyses can be conducted on human musculoskeletal models without the risk of injuries.

Simulation-based approaches have been used to design, optimize, and study different types of assistive devices~\cite{Grabke2019,Smith2021review}. Through these investigations, researchers have suggested assistance profiles different than scaled biomimetic joint moments~\cite{Dembia2017,Uchida2016_idealexo_running}. Simulation-based studies have also helped researchers to explain some of the experimentally observed behaviors~\cite{Jackson2017,Farris2014}. For instance, simulations showed that exoskeletons can have a negative effect on the muscle fiber mechanics~\cite{Farris2014,Sawicki2016}. Simulations have also been used to suggest assistance strategies for subjects with impaired gait cycles~\cite{Lim2016,Lim2017}. Readers are referred to~\cite{Grabke2019} and~\cite{Smith2021review}, which provide comprehensive reviews of simulation-based studies.

OpenSim is an open-source software that has been extensively utilized in movement science related fields~\cite{Seth2018,Delp2007}. Despite the limitations on musculoskeletal modeling and simulations~\cite{Hicks2015}, OpenSim enables researchers  to investigate human movements by providing them with biomechanical models and simulation tools~\cite{Seth2018,Dembia2017}. OpenSim has been widely used to design and study assistive devices~\cite{Uchida2016_idealexo_running,Dembia2017,Gordon2018,Ong2016,Aftabi2021simulation,Bianco2021}. For instance, Uchida~\etal~\cite{Uchida2016_idealexo_running} simulated several combinations of \emph{ideal} assistive devices on subjects running at 2~m/s and 5~m/s and reported that muscle activations can be decreased even in the muscles that do not cross the assisted joints, while they can be increased in some other muscles, based on the configuration of the assistive device. Their simulation results confirmed and offered clarification on some of the similar phenomena observed in experimental studies. Effects of several \emph{ideal} assistive devices on the metabolic cost of subjects carrying heavy loads have been studied by Dembia~\etal~\cite{Dembia2017}. Their study suggests effective joint configurations for ideal assistance and provides a perspective on how an assistive device can change the muscular activities of subjects while carrying loads.

Recently, predictive simulations or simulation-based dynamic optimization approaches are emerging to study assistive devices. These approaches can capture the effect of assistive devices  on the musculoskeletal kinematics and kinetics~\cite{Dembia2019_Moco,Geijtenbeek2019}. For instance,  Nguyen~\etal~\cite{Nguyen2019} studied the effect of an ankle exoskeleton on subjects walking at  normal speeds through predictive simulations, which enabled them not only to study the effect of the exoskeleton on the metabolic cost, but also to investigate how the exoskeleton affects subject's kinematics and ground reaction forces. Similarly, an active ankle powered prosthesis has been studied via predictive simulation framework by LaPre~\etal~\cite{LaPre2014}. Predictive simulations have also been employed to simulate knee~\cite{Zhao2013} and ankle prostheses to investigate the effects of various control approaches~\cite{Handford2016}. Moreover, passive ankle prosthesis and ankle-foot orthosis stiffness have been optimized using this strategy~\cite{Sreenivasa2017,Fey2012}. Finally, a dynamic optimization approach has been used by Handford and Srinivasan~\cite{Handford2016} to conduct a Pareto optimization of a robotic lower limb prosthesis, by simultaneously optimizing the metabolic and prosthesis cost rates.

\bigskip
\subsection*{Contributions}

We introduce a simulation-based design approach to systematically design exoskeletons that reduce the metabolic cost of locomotion. Along these lines, we propose a Pareto optimization approach that enables rigorous and fair comparison of effects of different exoskeletons designs and assistance torques on the metabolic cost of locomotion. We demonstrate the proposed systematic mechatronics system design approach by introducing a bi-articular exoskeleton design and comparing its efficiency with the commonly used mono-articular exoskeletons. In particular, we simultaneously optimize the power consumption of and the metabolic rate reduction due to assistance provided by bi-articular and mono-articular hip-knee exoskeletons during no load and loaded walking conditions, and present a rigourous comparison among such devices.

The proposed bi-articular exoskeleton design is motivated by human anatomy, where bi-articulation is known to improve the efficiency of human bipedal locomotion~\cite{Junius2017}. While a mono-articular hip-knee exoskeleton assists each joint directly by actuators mounted at these joints, a bi-articular hip-knee exoskeleton has all actuators mounted at the hip, mimicking bi-articular muscles in the human musculoskeletal system to improve locomotion performance, by enabling power transformation from proximal to distal joints~\cite{Schenau1989_a,Ron1996_a} and promoting power regeneration between adjacent joints~\cite{Boris1996,Wells1988}, facilitating the coupling of joint movements~\cite{Junius2017}, and improving the distribution of muscle weight and leg inertia~\cite{Junius2017}.

Our multi-criteria optimization results subsume the single objective optima. In particular, our results include the solution to a commonly addressed single objective optimization problem that aims to maximize the metabolic rate reduction of exoskeletons without considering their power consumption or torque capabilities, commonly referred to as \emph{ideal exoskeleton} optimization~\cite{Uchida2016_idealexo_running,Dembia2017}. Our ideal exoskeleton optimization results verify that, without any limits on the actuator torques and the power consumption, both ideal mono-articular and bi-articular devices can achieve the same level of metabolic rate reduction under the same total power consumption. Moreover, we show that the assistance torques not only improve the metabolic rate, but can also considerably decrease the peak reaction forces and moments at the knee, patellofemoral, and hip joints. Furthermore, in addition to the direct effect of ideal exoskeletons on the muscular activities of the hip and knee joints in the sagittal plane, we verify that the assistance provided by the exoskeletons can also indirectly affect the activity of muscles at the ankle joint and the muscles related to the hip abduction. In addition, we show that loading subjects with a heavy backpack results in a predictable change in the assistance profiles, by causing a proportional increase in the magnitude and a time shift. 

While ideal exoskeleton optimizations can provide several useful insights, different exoskeleton designs cannot be meaningfully compared without introducing some realistic physical limits on the actuator torques and considering the power consumption of these devices. Along these lines, we introduce different levels of peak torque constraints on the actuators of both mono-articular and bi-articular exoskeletons and compare their performance by considering the metabolic cost reduction and the power consumption metrics, \emph{simultaneously}. We show that introducing sufficiently large torque limits to the actuators of both devices does not have a significant impact on the metabolic cost reduction, while such limits can cause a statistically significant and meaningful reductions in the average and maximum positive power consumption of the exoskeletons. We also show that both devices can reach similar performance levels for different assignments of peak torques to the hip and knee joints, while larger peak torque limits are required for mono-articular exoskeletons compared to the bi-articular devices. Despite the similar assistance levels provided by both exoskeletons, mono-articular exoskeletons demonstrate better performance on reducing the peak joint reaction moments and forces, while the power consumption of bi-articular exoskeletons is less affected by loading. 

We also demonstrate the effect of regeneration on the power consumption of exoskeletons by superposing this effect on the Pareto-front curves. We show that regeneration can have a significant impact on the power efficiency of exoskeletons, as it can improve the power consumption of these devices from 6.54\%~$\pm$~2.60\% to 25.76\%~$\pm$~4.34\%, depending on the efficiency of regeneration system, the kinematics of the exoskeleton, and the torque/power limitations of the actuators.  Our results indicate that the knee actuators of mono-articular devices have more regeneration potential, while all actuators of the bi-articular devices possess similar regeneration potential.
%

We also show the detrimental effects of the inertial properties of exoskeletons on the metabolic cost of locomotion, by superposing this effect on the Pareto-front curves. In particular, we estimate the effect of additional mass and inertia on metabolic rate of subjects for mono-articular and bi-articular exoskeletons based on~\cite{Browning2007} and quantify these effects through a modified version of the augmentation factor proposed in~\cite{Mooney2014_a}. Our results indicate that the added device inertia causes optimal mono-articular and bi-articular devices loose their efficiency by 42.51\%~($\pm$~0.17\%)--55.51\%~($\pm$~0.11\%), and 35.12\%~($\pm$~0.21\%)--49.67\%~($\pm$~0.21\%), respectively. For bi-articular exoskeletons, the inertial effects simply shift the Pareto-front curve, without significantly affecting the Pareto-optimal configurations, while the non-dominated solutions are significantly affected by the device inertia for the mono-articular exoskeletons.

%

\subsection*{Outline}
The rest of the paper is organized as follows:  \emph{Musculoskeletal simulations of assisted subjects} section discusses the kinematics of the proposed bi-articular and mono-articular exoskeletons, explaining the relationship between these assistive device configurations. This section also presents the musculoskeletal model used for the simulations in the OpenSim framework and explains the procedures used for the simulations and analyses.  \emph{Multi-criteria design optimization and comparison} section explains the Pareto optimization and comparison approach based on the OpenSim framework and details the objective functions and performance metrics considered. \emph{Multi-criteria comparisons and selection of an optimal solution among all non-dominated solutions} subsection presents the  secondary criteria considered to select among non-dominated solutions. We also discuss methods of statistical analyses to conclude this section.

The results and their discussion are separated into seven main subsections under the \emph{Results and discussion} section. \emph{Validation of simulations} subsection discusses the procedures of validating the performed simulations. \emph{Ideal exoskeleton results} subsection presents and discusses the results of simulations performed for exoskeletons under ideal conditions, without any constraints on their torque/power output. \emph{Simulation-based multi-criteria optimization results} subsection discusses the Pareto-front curves of the mono-articular and bi-articular exoskeletons and presents comparisons among ideal and torque limited devices. \emph{Inclusion of regeneration effects on the Pareto solutions} subsection presents the effect of regeneration of the negative power on the energy efficiency of the torque limited devices and discusses the regeneration potential of actuators for different device configurations. \emph{Inclusion of interial effects on the Pareto solutions} subsection discusses the effect of inertial properties of devices on the Pareto-front solutions. The case studies of selected non-dominated solutions are discussed through \emph{Selection of optimal exoskeleton among the non-dominated solutions} subsection. We conclude  \emph{Results and discussion} section by discussing the general shortcomings of simulation-based studies and the specific limitations of the study we have conducted. Finally, \emph{Conclusions and future work} section concludes the paper and provides future research directions.

\section*{Musculoskeletal simulations of assisted subjects}


Two types of exoskeletons (referred to as exoskeleton configurations) are studied through musculoskeletal simulations by conducting simulations of seven subjects walking at their chosen speed, with no load and while carrying a 38 kg load on their torso. The experimental data used in this study has been experimentally collected, processed, and made publicly available by Dembia~\etal~\cite{Dembia2017}.

The musculoskeletal model used in the simulation is  based on the three-dimensional model developed by Rajagopal~\etal~\cite{Rajagopal2016} with 39 degrees of freedom, in which the lower limbs are actuated using 80 massless musculotendon actuators, and the upper limb is actuated by 17 torque actuators. This three-dimensional musculoskeletal model is adapted by locking some unnecessary degrees of freedom for both normal walking and walking with a heavy load scenarios and modeling the extra load on the torso of the musculoskeletal model for the walking with a heavy load condition as in~\cite{Dembia2017}.

Since this research is built upon the study performed by~\cite{Dembia2017}, we follow a similar terminology to enable easy comparisons. Therefore, the \textit{loaded} condition refers to subjects walking while carrying a 38 kg load on their torso, while the \textit{noload} condition refers to subjects walking  at their chosen speed without any extra load.

\begin{figure*}[b!]
	\centering
	\subfloat[\small{bi-articular exoskeleton}]{\includegraphics[width=2.45in]{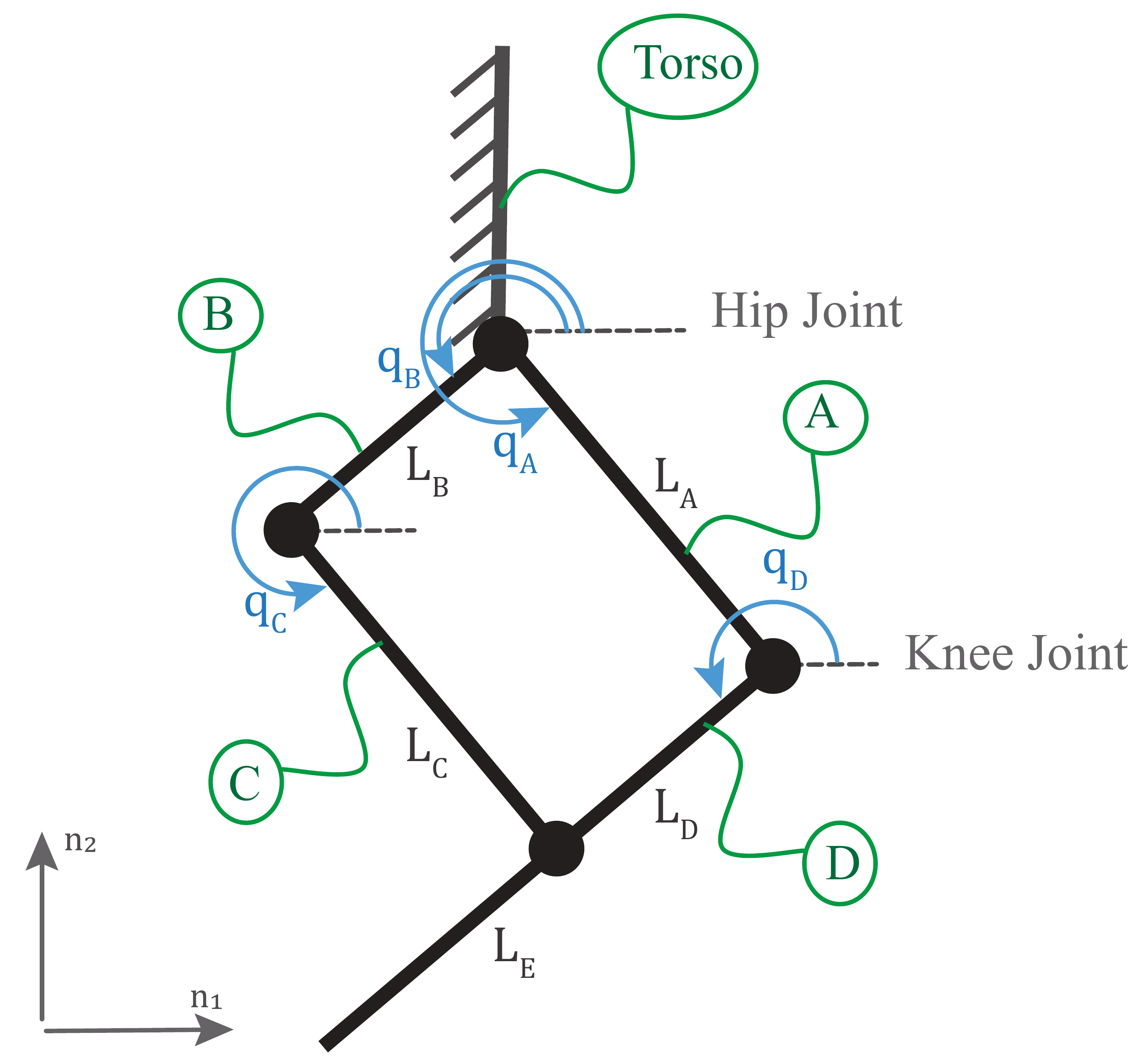}
		\label{Fig_biarticular_Exo_Mechanism}}
	\hfil
	\subfloat[\small{mono-articular exoskeleton}]{\includegraphics[width=2.45in]{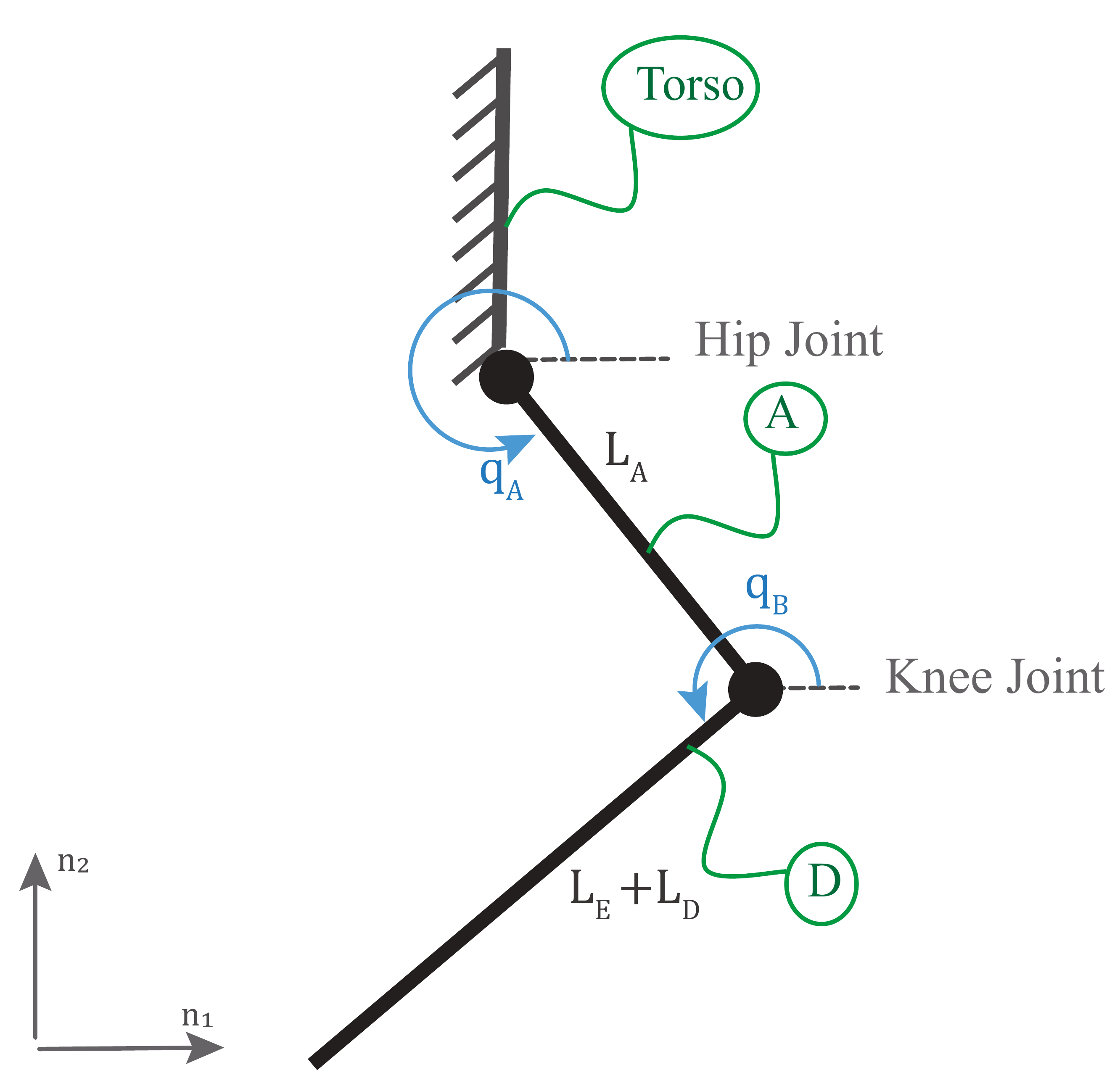}
		\label{Fig_monoarticular_Exo_Mechanism}}
	\caption{\small{\textbf{Kinematics of assistive devices.} A parallelogram is used to implement the kinematics of the bi-articular exoskeleton, while the mono-articular exoskeleton is modeled as a two-link serial manipulator. Symbols $q_A$ to $q_D$ represent the hip and knee joint angles, respectively. Bodies $Torso$, $A$, $D$ represent the torso (depicted as grounded for simplicity of analysis), the upper leg, and the lower leg of the user. Bodies $B$ and $C$ are the links used to implement the symmetric parallelogram mechanism. Symbols $L_A$ to $L_E$ stand for the lengths of the upper leg and the lower leg, respectively.}} 
	\label{Fig_Exos_Kinematics_Model}
\end{figure*}

\subsection*{Modeling of assistive devices}

Mono-articular and bi-articular exoskeleton configurations are proposed to assist the hip and knee joints. The bi-articular exoskeleton is inspired by bi-articular muscles and  aims to keep the large portion of the device weight around the proximal hip joint, while delivering the required power to the distal knee joint. A parallelogram mechanism is utilized for the bi-articular exoskeleton to accomplish this goal. The underlying kinematics of the purposed bi-articular exoskeleton is depicted in Fig~\ref{Fig_Exos_Kinematics_Model}\subref{Fig_biarticular_Exo_Mechanism}.

A mono-articular exoskeleton is modeled as a two-link serial manipulator as shown in Fig~\ref{Fig_Exos_Kinematics_Model}\subref{Fig_monoarticular_Exo_Mechanism}, in which each actuator is directly attached to the assisted joint. The kinematic models of the mono-articular and bi-articular exoskeletons at both the configuration and motion levels are presented in~\nameref{S1_Appendix}.

A linear mapping between the kinematics of mono-articular and the bi-articular exoskeleton can be established to relate these two devices through a linear mapping $J$ as
\begin{equation}\label{Eqn_Mono_Bi_Jacobian}
\begin{aligned}
\omega_{\mathrm{mono-articular}} &= J \hspace{2mm} \omega_{\mathrm{bi-articular}}\\
\left\lbrack \begin{array}{c}
{}^{\mathrm{torso}} {\omega_{\mathrm{mono}} }^{\mathrm{femur}} \\
{}^{\mathrm{femur}} {\omega_{\mathrm{mono}} }^{\mathrm{tibia}}
\end{array}\right\rbrack &=\left\lbrack \begin{array}{cc}
1 & 0\\
-1 & 1
\end{array}\right\rbrack \left\lbrack \begin{array}{c}
{}^{\mathrm{torso}} {\omega_{\mathrm{bi}} }^{\mathrm{femur}} \\
{}^{\mathrm{torso}} {\omega_{\mathrm{bi}} }^{\mathrm{tibia}}
\end{array}\right\rbrack
\end{aligned}
\end{equation}
\noindent where ${}^{\mathrm{torso}} {\omega_{\mathrm{mono}} }^{\mathrm{femur}}$ and ${}^{\mathrm{femur}} {\omega_{\mathrm{mono}} }^{\mathrm{tibia}}$ represent the angular velocities of the hip and knee actuators of the mono-articular exoskeleton, while ${}^{\mathrm{torso}} {\omega_{\mathrm{bi}} }^{\mathrm{femur}}$ and ${}^{\mathrm{torso}} {\omega_{\mathrm{bi}} }^{\mathrm{tibia}}$ stand for the angular velocities of the bi-articular hip and knee exoskeleton, respectively.

\begin{figure*}[b!]
	\centering
	\subfloat[\small{bi-articular exoskeleton}]{\includegraphics[width=2.0in]{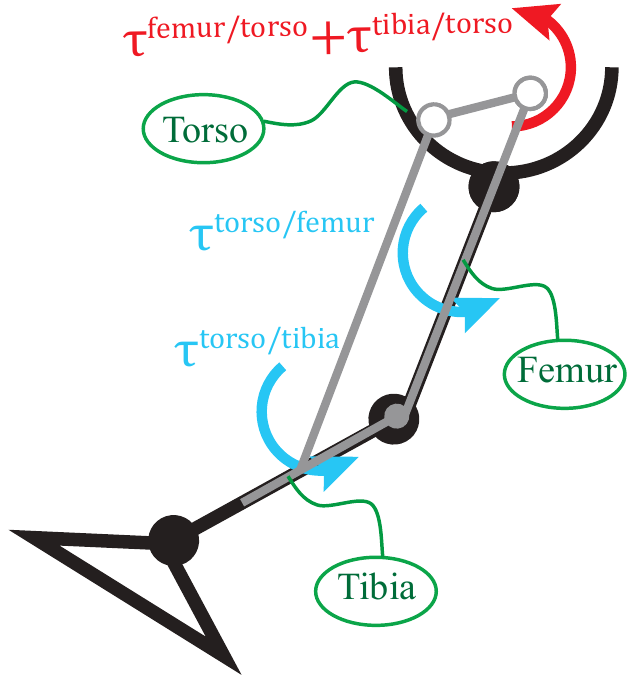}
		\label{Fig_Bi_Exo_Model_Opensim}}
	\hfil
	\subfloat[\small{mono-articular exoskeleton}]{\includegraphics[width=2.0in]{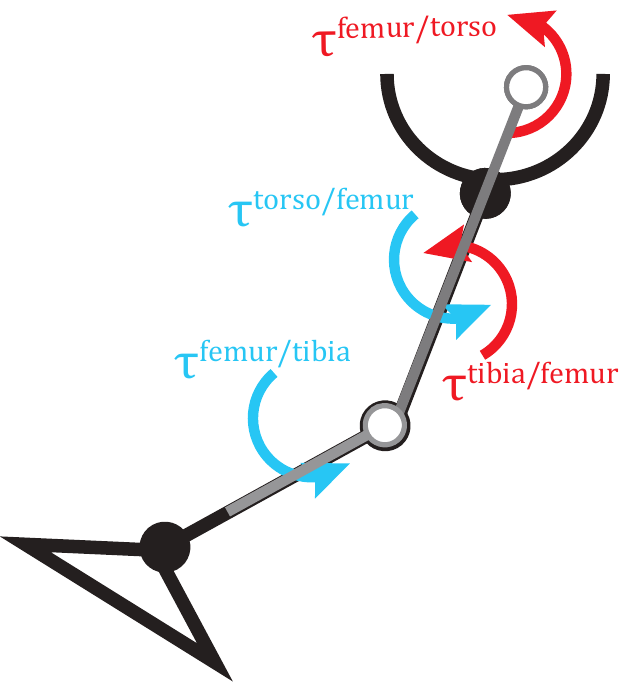}
		\label{Fig_Mono_Exo_Model_Opensim}}
	\vspace{1mm}
	\caption{\small{\textbf{Assistive actuator model of assistive devices.} The blue and red arrows represent the action and reaction torques of the assistive actuators on the assistive devices, respectively.}} 
	\label{Fig_Exos_Model_Opensim}
\end{figure*}

In order to model the ideal exoskeletons in the OpenSim framework, ``torque actuators'' provided by OpenSim API are utilized. The action and reaction torques of the bi-articular and mono-articular exoskeletons are assigned, as shown in Fig~\ref{Fig_Exos_Model_Opensim}.

As presented in Fig~\ref{Fig_Exos_Model_Opensim} \subref{Fig_Bi_Exo_Model_Opensim}, both torque actuators of the bi-articular exoskeleton are located at the torso; the reaction forces of the actuators are then applied to the torso, which matches the kinematics and dynamics model of the bi-articular exoskeleton.
\begin{eqnarray}\label{Eqn_biarticular_Torque_Act}
\tau^{hip}_{bi} = \tau^{torso/femur} \nonumber\\
\tau^{knee}_{bi} = \tau^{torso/tibia} \nonumber
\end{eqnarray}

\noindent where ${{\tau_{\mathrm{bi}}^{\mathrm{torso}/\mathrm{femur}} }}$ and ${{\tau_{\mathrm{bi}}^{\mathrm{torso}/\mathrm{tibia}} }}$ stand for torques of the bi-articular hip and knee actuators, respectively.

The mono-articular exoskeleton depicted in Fig~\ref{Fig_Exos_Model_Opensim} \subref{Fig_Mono_Exo_Model_Opensim} is modeled by assigning the action torque of the hip joint actuator from the torso to the femur and the action torque of the knee joint actuator  from the femur to the tibia
\begin{eqnarray}\label{Eqn_monoarticular_Torque_Act}
\tau^{hip}_{mono} = \tau^{torso/femur} \nonumber\\
\tau^{knee}_{mono} = \tau^{femur/tibia}\nonumber
\end{eqnarray}

\noindent  where  ${{\tau_{\mathrm{mono}}^{\mathrm{torso}/\mathrm{femur}} }}$ and ${{\tau_{\mathrm{mono}}^{\mathrm{femur}/\mathrm{tibia}} }}$ represent the torques of the mono-articular hip and knee actuators, respectively.

Transpose of the linear map $J$ in Eq~\eqref{Eqn_Mono_Bi_Jacobian} provides the mapping  between the torques provided by exoskeletons as
\begin{equation}\label{Mono_Bi_Torque_Mapping}
\tau_{\mathrm{bi-articular}} =J^T \hspace{2mm} \tau_{\mathrm{mono-articular}}
\end{equation}

If no saturation limits exist on the actuator velocities or torques, the linear mapping between the exoskeletons dictates that they can provide the same level of assistance with the same amount of power consumption, under ideal conditions. However, this linear mapping between the mono-articular and the bi-articular exoskeletons no longer holds if the hip and knee actuators have velocity or torque saturation.

\begin{figure*}[b!]
	\includegraphics[width=\linewidth]{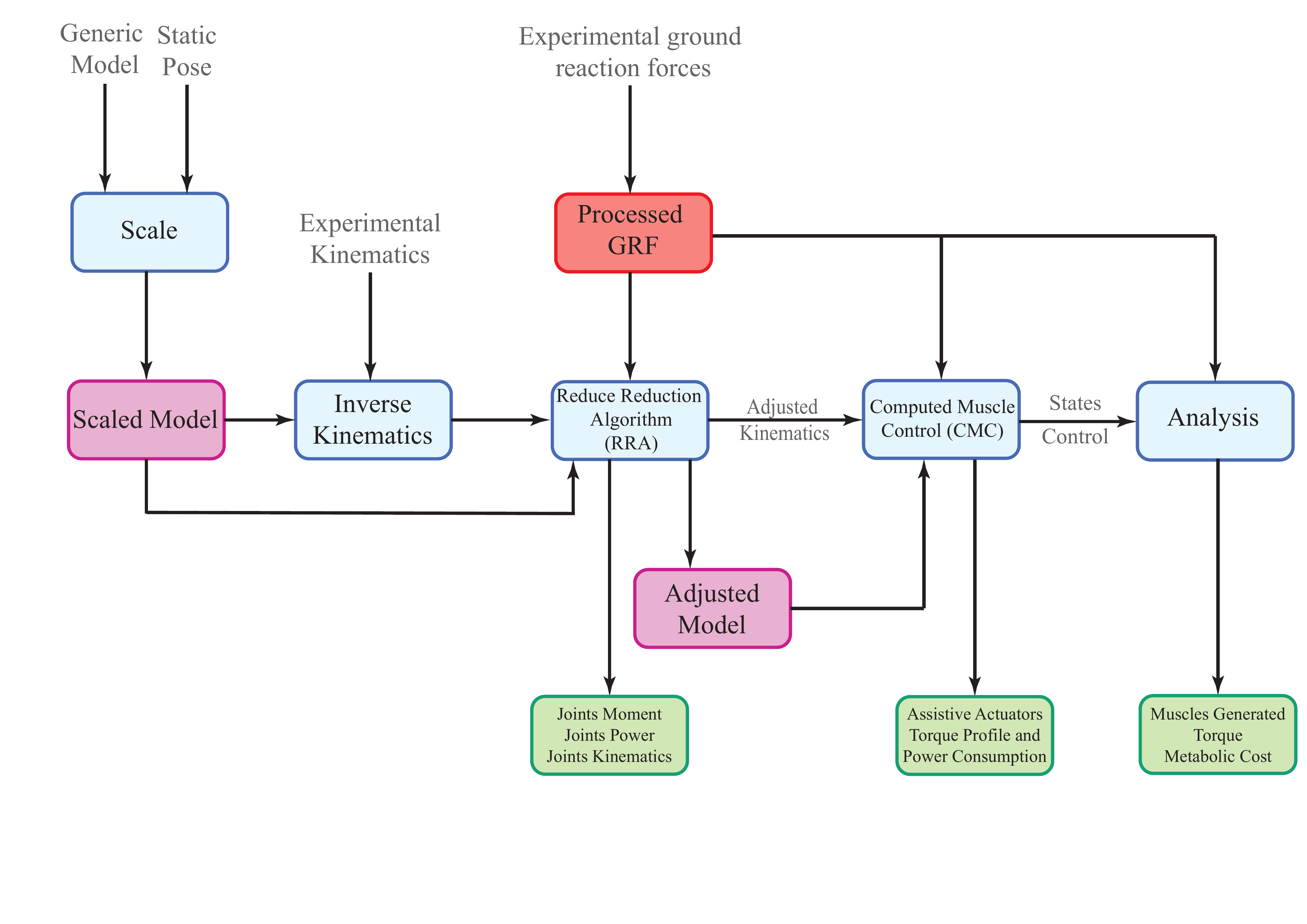}
	\vspace{-4\baselineskip}
	\caption{\small{\textbf{Block diagram of OpenSim simulation.} Green blocks stand for the outputs, blue blocks represent OpenSim simulations or analyses, purple blocks signify models that have been used for simulations and analyses, and red blocks represent the experimental data.}}
	\label{Fig_OpenSim_Sim_Procedure}
\end{figure*}
\subsection*{Musculoskeletal simulation workflow}

Fig~\ref{Fig_OpenSim_Sim_Procedure} presents an overview of the musculoskeletal simulation workflow using OpenSim. The first step of conducting the simulations for each subject is scaling the generic dynamic model to acquire a musculoskeletal model that matches with the anthropometry of each subject. This scaling is performed using OpenSim Scale Tool. Similarly, the maximum isometric forces of the muscles are scaled according to the mass and height of each subject. After obtaining the scaled model for each subject, the inverse kinematics for each subject is computed using OpenSim Inverse Kinematics Tool to map experimentally collected motion capture data to trajectories of joint angles.

At the second step of the simulation workflow, the incompatibility between the experimentally collected data (ground reaction forces and motion capture data) and the musculoskeletal model are minimized by slightly adjusting the inertial properties and kinematics of the scaled model of the simulated subject. OpenSim residual reduction algorithm (RRA) is utilized for this purpose~\cite{Delp2007}.

As the third step, the muscle recruitment computations for the scaled and adjusted musculoskeletal model are performed using the computed muscle control (CMC) algorithm in OpenSim~\cite{Thelen2003_b}. The CMC algorithm resolves the muscle redundancy by a least-square based static optimization approach to compute  optimal muscle recruitment to track the adjusted kinematics~\cite{Hicks2015}. The CMC algorithm assumes that, throughout  the simulation, the kinematics and dynamics of the subject remain consistent with the experimentally captured data; hence, and any changes in kinematics or inertial properties of the subject may cause a systematic error in the results. The effort-based objective function used by the OpenSim CMC algorithm is given as

\begin{equation}\label{Eqn_CMC_Objective}
J = \sum_{i\in nMuscles} a_{i}^{2} + \sum_{i \in nReserves} (\frac{\tau_{r,i}}{w_{r,i}})^2.
\end{equation}

To investigate the performance of the assistive devices and their effect on the human musculoskeletal system through the OpenSim simulation framework, the CMC algorithm is used with a modified objective function. The objective function of the original CMC algorithm  depends on the sum of squared of muscle activations and reserve actuators as in Eq~\ref{Eqn_CMC_Objective}. In this equation, $w_{r,i}$ determines the weight of the reserve actuators, which is generally selected as a small number to highly penalize the use of reserve actuators, as the utilization of these actuators is required only to compensate for modeled passive structures and potential muscle weakness~\cite{Dembia2017}.

In order to include effects of assistive devices into musculoskeletal simulations, assistive torques provided by the actuators of the exoskeleton need to be considered as part of the CMC algorithm. Along these lines, assistive torques provided by OpenSim Torque Actuators need to be added to the objective function of CMC algorithm. To enable inclusion of assistive torques,  the objective function is adjusted as
\begin{equation}\label{Eqn_CMC_Assisted_Obj_Func}
	J \hspace{2mm}  = \sum_{i\in nMuscles} a_{i}^{2} + \sum_{i \in nExo} \left(\frac{\tau_{exo,i}}{w_{exo,i}}\right)^{2} +  \sum_{i \in nReserves} \left(\frac{\tau_{r,i}}{w_{r,i}}\right)^2.
\end{equation}

In the adjusted objective function, $w_{exo,i}$ represents the weights of each torque actuator (called as optimal force in OpenSim) and these weights are used to penalize the utilization of the torque actuators. Large values assigned to $w_{exo,i}$ result in a low penalty, promoting dominant the use of torque actuators in the CMC optimization. Small values assigned to $w_{exo,i}$ result in a high penalty and the CMC algorithms minimizes the use of exoskeleton actuators. In this study, in order to utilize exoskeleton assistance as much as possible, the weights of  torque actuators are set to a high value (1000 N-m), promoting the optimizer to use the assistive actuators during the gait cycle.

Finally, the output of the CMC algorithm is used to compute the metabolic power consumption, muscle moments, and joint reaction forces of the subject, utilizing the Analysis Tool of OpenSim.

Please note that, in this study, we rely on experimentally collected data and adjusted models provided by~\cite{Dembia2017} for the first two steps of the musculoskeletal simulations.


\subsection*{Metabolic model} To estimate the instantaneous metabolic power of subjects, Umberger's muscle energetic model~\cite{Umberger2003} modified by Uchida~\etal~\cite{Uchida2016_metabolic_model} is employed. The average power consumption of a muscle during a gait cycle is calculated as

\begin{equation}\label{Eqn_avg_muscle_power}
	P_{avg} = \frac{m}{t_1 - t_0}\int_{t_0}^{t_1} \dot{E}(t) dt,
\end{equation}

\noindent where $m$ is the muscle mass, and $\dot{E}(t)$ is the normalized metabolic power consumed by the muscle. The whole body metabolic power is calculated by summing the metabolic power of all muscles. Finally, to compute the gross metabolic power consumption of each subject,  the metabolic power over the gait cycle is integrated and normalized by the mass of each subject.

Assisted subjects are analyzed similar to the unassisted case: the instantaneous metabolic power of the subjects is computed using the energetic model of Uchida~\etal~\cite{Uchida2016_metabolic_model} and the metabolic rate of each subject is estimated through integration of the metabolic power over the gait cycle.

In order to compute the power consumption of the assistive actuators, the power profiles of the actuators are extracted and their absolute power profiles are integrated over the gait cycle and normalized by the subject mass. Similarly, the negative power or regeneratable power through a gait cycle is calculated by obtaining the negative power profiles and integrating them over the gait cycle and normalizing them by the mass of the subject.


\subsection*{Computation of joint reaction forces/moments} Since the equations of motion of the musculoskeletal model are formulated in terms of generalized coordinates and generalized forces, the internal forces and moments are not directly solved for. Consequently, a joint reaction analysis needs to be utilized to compute the reaction forces and moments between two consecutive bodies in the kinematic chain.  OpenSim employs a  Newton-Euler formulation to recursively solve for  the reaction forces and moments at each joint from the distal to the proximal joints, as detailed in~\cite{Steele2012}.

In this study, assistive torques provided by the torque actuators are appended to the Newton-Euler equations of bodies and a joint reaction force/torque analysis is performed to study the effect of the assistive devices on the reaction forces and moments of assisted and unassisted joints.

\section*{Multi-criteria design optimization and comparison}

The CMC algorithm resolves the muscle redundancy by applying static optimization to the weighted sum of squares objective function, to solve it in a least-square sense at each instant of time, to track the captured data of the subject with the lowest cost~\cite{Hicks2015}. While studying ideal exoskeletons, the weights assigned for the assistance torques are selected in such a way that, these actuators are promoted to be utilized as much as possible. However, this approach does not take any physical limitations of exoskeletons into account.

Although musculoskeletal simulations of ideal assistive devices provide insights into the maximum assistance that can be provided to the human musculoskeletal system, in real life applications, the exoskeletons are constrained by the power that can be supplied to their actuators and the maximum assistive torque that the actuation modules can provide to the joints of interest. Furthermore, kinematics and inertial distribution of exoskeletons are other physical aspects that need to be taken into account.

In general, there exist  trade-offs among the power consumption, the metabolic cost reduction, and the inertial properties of assistive devices. One the one hand, a high metabolic cost reduction necessitates large torque capacities and large power consumption, and on the other hand, power-efficient and lightweight exoskeleton designs require compact and efficient actuator units that cannot provide high assistance torques.

Taking energy efficiency of exoskeletons into consideration together with their beneficial effects on the metabolic burden of subjects introduces another optimization criterion to be studied while optimizing the performance of devices. Hence, a multi-criteria design optimization problem needs to be addressed to study the trade-off between the energy efficiency and the metabolic cost reduction.

One way to address the  multi-criteria design optimization problem is to use a weighted sum of the cost functions. Assigning predetermined weights to define a single aggregate objective function, scalarization approaches enable the original multi-objective problem to be formulated as a single
criterion optimization problem. The drawback of these approaches is that the preferences between objectives need to be assigned a priori, before having a complete knowledge of the trade-off involved.

\begin{figure}[b!]
	\includegraphics[width=.825\linewidth]{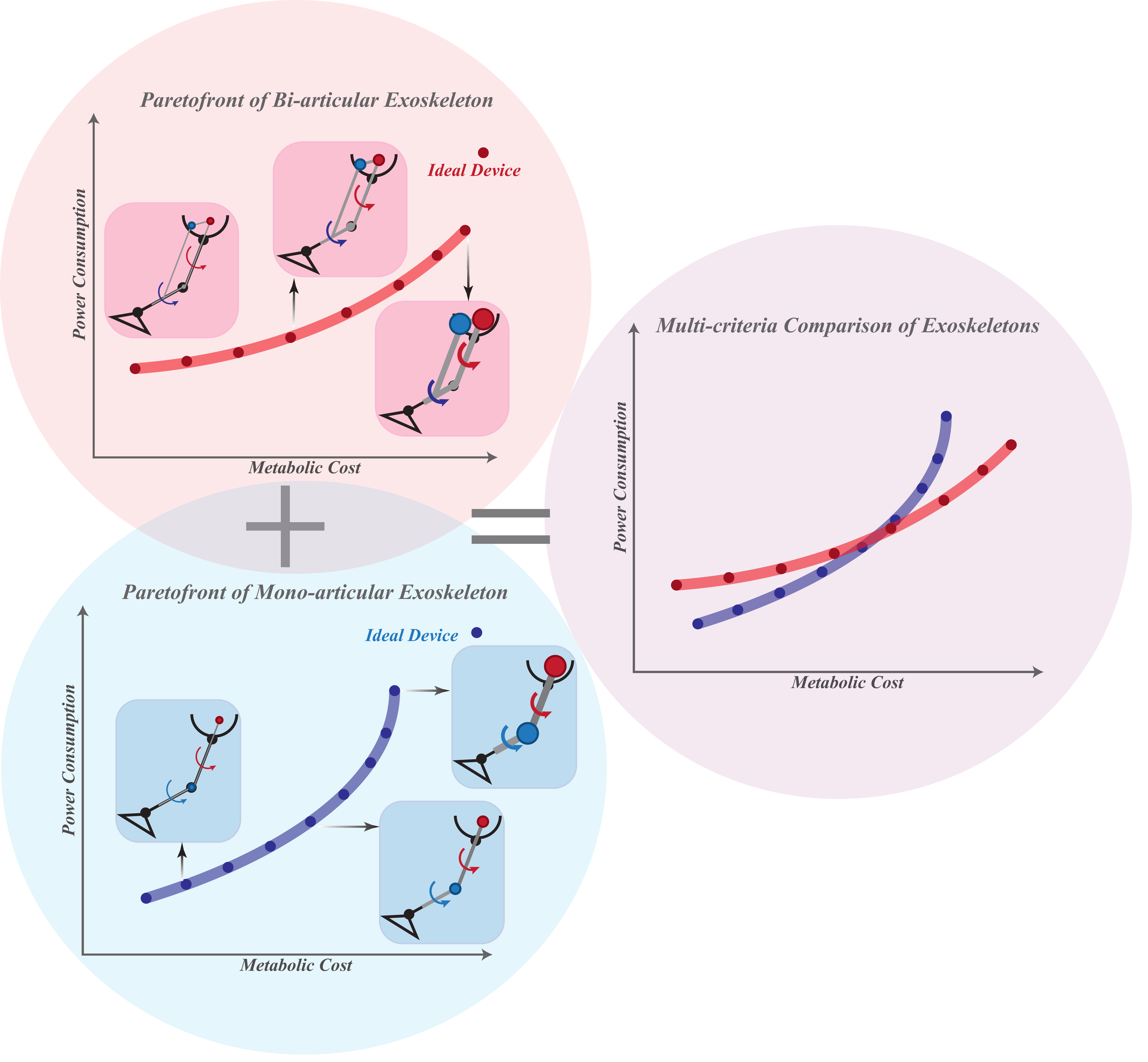}
	\caption{\small{\textbf{Schematic representation of multi-criteria comparison of bi-articular and mono-articular exoskeletons}}}
	\label{Fig_Multicriteria_Comparison}
\end{figure}

We advocate for the use of Pareto methods for musculoskeletal simulation based multi-criteria design optimization of assistive devices. Given a multi-criteria optimization problem, there exist multiple solutions that reflect optimal designs for different design preferences among the selected metrics. All such \emph{non-dominated} solutions constitute the Pareto front~\cite{papalambros_wilde_2000,Marler2010,unalPareto}.  Unlike one-shot scalarization-based optimization methods, Pareto methods fully characterize the trade-off among objectives. Once the Pareto front is computed, the designer can study these solutions to get an insight of the underlying trade-offs and make an informed decision to finalize the design by selecting an \emph{optimal} solution from the Pareto set. Secondary constraints or new design criteria that have not been considered during the original optimization can also be introduced at the design selection stage. Consequently, Pareto methods allow the designer to choose alternative optimal solutions under different conditions.

A Pareto optimization approach is rewarding as it not only leads to musculoskeletal simulation based design of optimal assistive devices, but also enables \emph{fair comparisons}  among various assistive devices, possibly with different underlying kinematic, dynamic, actuation and loading properties~\cite{Unal2008,Kamadan2017,Yusuf2020}.  Given that the Pareto front for each assistive device characterizes the best performance of that exoskeleton for all possible preferences of the designer, fair comparisons become possible by studying the Pareto front of each exoskeleton. Hence, for any given designer preference, the best possible performances of each exoskeleton can be compared to the best possible performances of other exoskeletons, as depicted in Fig~\ref{Fig_Multicriteria_Comparison}, leading to a fair and systematic comparison methodology.

Pareto optimization approaches consist of three main stages: i) the selection and evaluation of performance metrics, ii) the computation of the Pareto front, and iii) the selection of an optimal solution among all non-dominated solutions.

%
%

\subsection*{Performance metrics}

The main goal of this study is to design an energy efficient untethered wearable assistive device that can reduce the metabolic cost of its users. Following the terminology proposed in~\cite{merlet2006parallel},  the performance requirements for assistive devices can be categorized into four groups as the imperative, optimal, primary, and secondary requirements.

Ensuring the safety of physical human-robot interaction and an ergonomic fit are imperative design requirements for exoskeleton-type assistive devices. The safety of physical human-robot interaction can be ensured by guaranteeing coupled stability of low-level interaction controllers~\cite{FatihEmre2020,aydin2018stable,Erdogan2011c,tokatli2015stability,mengilli2021passivity,mengilli2020},
 while ergonomic fit can be achieved through the implementation of proper joint alignment mechanisms~\cite{Ergin2011,Ergin2012,MustafaVolkan2012,BesirVolkan2013,Erdogan2016}. Both of these aspects are out of the scope of this study.

The metabolic cost reduction of its users is an optimal performance requirement that needs to be maximized while designing an assistive exoskeleton. In particular, the assistive devices considered in this study are required to provide assistive torques to decrease the metabolic cost of walking at a self-selected speed, with and without a heavy load. To quantify this metric, the normalized gross whole-body metabolic rate of each subject (for different assistive devices and under two load conditions) are calculated, and the metabolic cost reduction is computed by comparing the metabolic rate of assisted and unassisted subjects. The  metabolic cost calculations are repeated for seven subjects in three trials to obtain the  average metabolic power expenditure. 

The energy efficiency of an untethered assistive device is another optimal performance requirement that needs to be maximized. The power consumption is a crucial metric, as it directly affects the battery life and torque output (consequently, the total weight) of untethered devices. To quantify this metric in the simulated exoskeletons, the absolute power consumed by all actuators is computed, considering that power regeneration mechanisms may not be implemented. Note that the absolute power is defined over the absolute value of instantaneous power and does not consider any power recovery due to negative work done. Additionally, the negative power of the exoskeletons is also computed to analyze the amount of power available for regeneration. This enables an appropriate portion of the negative power to be subtracted from the absolute power consumed  to study the effect of regeneration on power efficiency. Finally, to assess the cost of carrying, the average and standard deviation of the maximum positive power of actuators of each device (normalized by subject mass)  under different loading conditions are calculated, as proposed in~\cite{Dembia2017}.

These two optimal performance requirement of metabolic cost reduction and device power consumption are optimized \emph{simultaneously} to compute a set of non-dominated solutions for each exoskeleton configuration. It should be noted that device power consumption metric indirectly addresses the device weight and inertia, as lower power devices can usually be implemented using smaller actuation and battery modules.

Primary requirements for  untethered exoskeleton-type assistive devices are considered as having light-weight and low inertia. Since static optimization based musculoskeletal simulations utilized in OpenSim do not allow for inertial changes to be studied directly, detrimental effects of the device weight and inertia are considered as part of the primary requirements, such that these effects can be extracted from the relevant human-subject studies in the literature and superposed onto the Pareto-front curves to facilitate the design selection process.

Additional primary requirements for assistive exoskeletons may be considered as changes in muscle activities of the key lower limb  muscles and the reaction forces and moments at the joints under assistance. Similar to the inertial effects, these aspects can be studied for each non-dominated device design computed by the multi-criteria optimization and may help facilitate the design selection process.

Finally, the secondary requirements for the exoskeleton may be considered as ease-of-implementation, robustness, and device cost. The secondary requirements are imposed by the designer based on the specific use scenario and implementation approach; hence, these considerations are kept out of the scope of this study.

\subsection*{Computation of the Pareto front}

Since optimal performance requirements of the metabolic cost reduction and the device power consumption constitute two objectives that conflict with each other, there exists no unique solution that simultaneously optimizes both objectives. Furthermore, the trade-off between these objectives is not entirely apparent; hence, assigning a relative importance to these objectives is not trivial. Instead of a computing a unique solution that optimizes both objectives simultaneously, multi-criteria design optimization aims at computing a set of mathematically equivalent solutions, called non-dominated solutions.  A non-dominated solution is an optimal solution such that  an improvement in one objective requires the degradation of another objective. The set of all non-dominated solutions forms a Pareto front.

There exists a wide range of techniques for multi-objective optimization \cite{Marler2004,Miettinen2008,Jakob2014}. In this study, simultaneous maximization of  metabolic cost reduction and minimization of device power consumption are performed using an $\epsilon$-constraint method, since this method is well-suited for use with musculoskeletal simulations and can solve for both the convex and non-convex regions of the Pareto front.

The $\epsilon$-constraint method computes optimal solutions on the Pareto front by optimizing one of the objectives, while the other objective is systematically expressed as an inequality constraint. In this way, the $\epsilon$-constraint method transforms a multi-objective optimization problem into a series of single-objective problems~\cite{Yacov1971,Laumanns2006,Chircop2013}. Through a systematic variation of the constraint bounds, the set of Pareto optimal solutions are obtained. 

To implement the $\epsilon$-constraint within the OpenSim musculoskeletal simulation framework, we systematically constrained the peak torque of the assistive actuators. Limiting the maximum assistance torques that can be provided during a gait cycle directly affects the redundancy resolution of the CMC algorithm; hence, the muscle recruitment and the metabolic cost reduction caused by the assistance.  Furthermore, given that the user kinematics is kept constant throughout the musculoskeletal simulations,  actuator saturation provides a systematic upper bound on the device power consumption metric. Along these lines, a systematic variation of the maximum torque that assistive actuators can provide over a discrete range results in optimal solutions in the objectives space. After filtering these solutions to select the non-dominated ones, a Pareto front curve is calculated for each configuration of the exoskeleton under each loading condition.

In this study, a symmetry constraint between the left and right leg actuators is imposed and the maximum torque that the actuators can provide to the hip and knee joints are varied between 30~Nm to 70~Nm with a step size of 10~Nm, to study all combinations for both the bi-articular and the mono-articular exoskeletons.

%

\subsection*{Multi-criteria comparisons and selection of an optimal solution among all non-dominated solutions}
\label{mass_inertia_subsection}
Once Pareto front curves are computed for each exoskeleton configuration under different loading conditions, different exoskeletons can be compared to each other  as in Fig~\ref{Fig_Multicriteria_Comparison} and an optimal solution can be selected among all non-dominated solutions. To facilitate the design selection, several metrics can be computed to quantify  the primary and secondary requirements for the non-dominated solutions.

\paragraph*{Effect of the device inertial properties on the energetics of subjects.}

The effect of device inertial properties on the energetics of subjects is studied based on the metabolic model by Browning~\etal~\cite{Browning2007}. The mass and inertia of the proposed bi-articular and mono-articular exoskeletons affect the waist, thigh, and shank segments of the body. Since the bi-articular exoskeleton enables placement of the knee actuation module to the waist instead of the thigh, the inertial properties of the mono-articular and bi-articular exoskeletons differ significantly. 

To quantify the effect of the inertial properties of the exoskeletons on the metabolic rate of assisted subjects, rigid links with identical inertial properties are considered at the hip, thigh and shank of the users. The center of mass (CoM) locations are chosen based on the mean lengths of the thigh and shank segments, under the assumption of uniform distribution of link mass~\cite{Paolo1996}. The inertial properties used for the numerical evaluations are presented in Table~\ref{Table_Exokeletons_Mass_Inerta}.

\begin{table}[b!]
	\renewcommand{\arraystretch}{1.2}
	\begin{adjustwidth}{-1in}{0in}
		\caption{\small{\textbf{Inertial properties of the bi-articular and mono-articular exoskeletons.}}}
		\begin{tabular}{c!{\vline width 0.2pt}c!{\vline width 0.2pt}c!{\vline width 0.2pt}c!{\vline width 0.2pt}c!{\vline width 0.2pt}c!{\vline width 0.2pt}c}
			\toprule
			\multirow{2}{*}{\textbf{Configuration}} & \textbf{Waist}   & \multicolumn{2}{ c!{\vline width 0.2pt} }{\textbf{Thigh}} & \multicolumn{2}{ c!{\vline width 0.2pt} }{\textbf{Shank}} & \textbf{Actuator} \\  \cmidrule{2-7}
			&  mass ($kg$)& mass ($kg$) & CoM ($m$)& mass ($kg$) & CoM ($m$)& inertia ($kg.m^2$) \\ \midrule[0.5pt]
			\textbf{bi-articular} &  4.5 &1& 0.23 &0.9&0.18 + $l_{thigh}$&$5.06\times10^{-4}$\\ \midrule[0.2pt]
			\textbf{mono-articular} & 3 &2.5& 0.30 &0.9&0.18 + $l_{thigh}$&$5.06\times10^{-4}$\\
			\bottomrule
		\end{tabular}
		\label{Table_Exokeletons_Mass_Inerta}
	\end{adjustwidth}
\end{table}

Actuation modules with identical mass are assigned to both exoskeletons. However, both the placement of the actuator units on the body and the transmission ratio assigned to each module to ensure the required maximum torque level result in significantly different inertial properties for each exoskeleton. A  direct drive motor with a peak output torque of 2~Nm is considered and the proper transmission ratio to achieve the required peak torque at each joint is calculated accordingly. The reflected inertia of the actuation unit is then computed using this transmission ratio. The inertia of the moving segments (i.e., thigh and shank segments) are computed about the hip joint in the body frame  by considering the distal mass effect on the inertia, reflected inertia of the actuation module, and the leg inertia provided in~\cite{Browning2007}.




While Browning~\etal~\cite{Browning2007} have characterized the metabolic rate of a subject with loaded segments, for this study, it is more informative to capture the change in the metabolic rate due different reflected inertia levels of various exoskeletons. Along these lines, the change in the metabolic rate due to inertial effects is extracted from~\cite{Browning2007}, by characterizing the change from noload conditions to loaded ones.  Eq~\eqref{Eqn_AddingMass_MetabolicRate}--\eqref{Eqn_AddingInertia_MetabolicRate} present the metabolic model used to analyze the effect of the inertial properties of the exoskeletons on the metabolic rate.

\begin{equation}\label{Eqn_AddingMass_MetabolicRate}
\begin{aligned}
\Delta MC_{Waist} &= 0.045\times m_{Waist}\\
\Delta MC_{Thigh} &= 0.075\times m_{Thigh}\\
\Delta MC_{Shank} &= 0.076\times m_{Shank}
\end{aligned}
\end{equation}

\small
\begin{eqnarray}\label{Eqn_AddingInertia_MetabolicRate}
I_{ratio} &=& \frac{I_{Exo,segment} + I_{un\textit{loaded}\;leg}}{I_{un\textit{loaded}\;leg}} \nonumber\\
\Delta MC_{Thigh} &=& ((-0.74 + (1.81\times I_{Thigh,ratio}))\times MC_{unassisted})-MC_{unassisted}\\
\Delta MC_{Shank} &=& ((0.63749 + (0.40916\times I_{Shank,ratio}))\times MC_{unassisted})-MC_{unassisted}\nonumber
\end{eqnarray} \normalsize

During the multi-criteria comparisons and the selection of an optimal solution among all non-dominated solutions, the detrimental effects of the mass and inertia of the exoskeletons on the metabolic cost of the subjects are reflected on the Pareto solutions according to  Eq~\eqref{Eqn_AddingMass_MetabolicRate}--\eqref{Eqn_AddingInertia_MetabolicRate}. Then, these solutions are filtered to obtain the Pareto front curves with the effect of the exoskeleton inertial properties on the metabolic expenditure.

\paragraph*{Modified augmentation factor.}

To facilitate the design selection among the optimal solutions on the Pareto front curves, the augmentation factor (AF) is utilized as another primary metric to enable a systematic analysis of the performance of devices under the effect of their inertial properties, for both {\it noload} and {\it loaded} conditions. The augmentation factor has been proposed by Mooney~\etal~\cite{Mooney2014_a}  to measure the relationship between the device applied positive power and the change in metabolic power consumption. The augmentation factor  estimates the metabolic power change due to carrying the exoskeleton, balancing the mean positive power resulting in a metabolic improvement and the net dissipated power and device weight causing metabolic detriment.

Although the augmentation factor introduces a general factor for predicting the performance of assistive devices, it does not address the effect of reflected inertia due to the actuation units and  the mass attached to moving segments. On the other hand, Browning~\etal~\cite{Browning2007} note the importance of the inertial effects on the metabolic burden of subjects, advocating for the necessity of including this factor to any performance measurement of exoskeletons. Consequently, we introduce a \emph{modified augmentation factor} (MAF) to extend the augmentation factor with such inertial effects.

In order to include the inertial effects into the augmentation factor, the detrimental effects of the mass and inertia of the exoskeletons on the metabolic cost of the subjects are modeled as in Eq~\eqref{Eqn_AddingMass_MetabolicRate}--\eqref{Eqn_AddingInertia_MetabolicRate} and
the device location factor $\gamma_{i}$ is computed for the inertia applied to each segment, according to Eq~\eqref{Eqn_Inertia_Factor}. It is noteworthy to mention that $\gamma_{i}$ is obtained considering the device inertia in addition to the inertia of the unloaded leg. Furthermore, the mass of each segment is verified to be within the range of weights of the exoskeletons studied by Mooney~\etal~\cite{Mooney2014_a} for calculating the augmentation factor.

\begin{equation}\label{Eqn_Inertia_Factor}
\gamma_{i} = \frac{A_i\times m_{subjects}\times MC_{unloaded}}{I_{unloaded}} \qquad i \in \{1,2,3\}
\end{equation}

In the inertia position factor in Eq~\eqref{Eqn_Inertia_Factor}, $A_i$ are the multipliers of $I_{ratio}$ in Browning metabolic models~\cite{Browning2007} for the foot, thigh and shank segments, while $I_{unloaded}$, $m_{subjects}$, and $MC_{unloaded}$ represent the inertia of a leg without any external load, the mean weight of subjects, and the metabolic rate of subjects walking without any load on their segments, respectively. The modified augmentation factor (MAF) is obtained by adding the effect of inertia on the metabolic detriment part of this factor, as in Eq~\eqref{Eqn_MAF}.

\begin{align}
MAF &= \frac{p^{+}+p^{disp}}{\eta}-\sum_{i=1}^{4}\beta_{i}m_{i}-\sum_{j=1}^{3}\gamma_{j}I_{j}\label{Eqn_MAF}\\
p^{disp} &= \alpha(p^{-}-p^{+}) \qquad \alpha = \left\{\begin{array}{ll}1\quad p^{+}<p^{-}\\0\quad p^{+}\geq p^{-}\end{array}\right.\label{Eqn_disp_power}
\end{align}

In Eq~\eqref{Eqn_MAF}, $p^{+}$, $p^{-}$, and $p^{disp}$ represent the mean positive, negative, and dissipated power calculated through Eq~\eqref{Eqn_disp_power}. The symbol $\beta_i$ in MAF stands for the location factors of the device mass, which are taken as 14.8, 5.6, 5.6, and 3.3~W/kg from the foot to waist, respectively~\cite{Mooney2014_a,Browning2007}; the symbol $\gamma_j$ represent the location factors of the device inertia, which are set to 47.22, 27.78, and 125.07~W/{kg.m$^2$} from the foot to thigh, respectively. Consistent with augmentation factor, MAF uses a muscle-tendon efficiency of $\eta = 0.41$ to convert the mechanical assistive power to metabolic power, as empirically determined  by Sawicki and Ferris~\cite{Sawicki2008} and Malcolm~\etal~\cite{Malcolm2013}. Finally, MAF  is normalized by the weight of each subject.

\paragraph*{Effect of regeneration on the power efficiency.}

The regeneratable power of optimal exoskeletons can be acquired by capturing the negative power profiles of the assistive actuators and obtaining the dissipated power from the negative power profile. This dissipated power is normalized by the mass and the gait duration of each subject during each trial and multiplied by the efficiency factor of the power electronics. The  effect of regeneration on the optimal devices and on their trade-off curves can then be studied by subtracting the regeneratable power from the total power consumption of the devices.

The regeneration performance of optimal devices are studied using various efficiency factors. While the maximum reported efficiency of harvesting dissipated power ranges between 30\% to 37\% for the lower limb assistive devices~\cite{Laschowski2019}, custom power electronics of MIT Cheetah robot~\cite{Seok2015} and the biomechanical power harvester developed by Donelan~\etal~\cite{Donelan2008} report experimentally verified regeneration efficiencies up to~63\%. Along these lines, a maximum efficiency of 65\% is considered in this study.

\paragraph*{Root mean square error and peak to peak difference of trajectories.} The root mean square error (RMSE) and peak-to-peak difference are utilized to establish quantitative comparisons between two trajectories, such as power, assistance torque, joint reaction forces, and joint torque trajectories. RMSE and the peak-to-peak difference can be computed over the whole gait cycle or for certain gait cycle phases, as categorized in~\cite{Perry1992}. Gait phases are computed for each subject and trial, according to their toe-off timing, as presented in~\nameref{S2_Appendix}. RMSE values are reported by their mean and standard deviation over the subjects. Additionally, the peak-to-peak difference for the set of Pareto-front solutions is reported with median and interquartile range (IQR) of their mean values.

\subsection*{Statistical analysis}

The experiments and their simulations were performed on seven subjects walking in two different loading conditions, and each condition is repeated for three trials. To identify the statistically significant factors and to enable multi-comparisons, relevant statistical analyses are used. In particular, N-way repeated measures analysis of variance (ANOVA) is utilized to identify the statistically significant factors and Tukey post-hoc tests are used for multi-comparisons. As relevant, data is partitioned and a series of one-way repeated measures ANOVA tests are also utilized to study individual factors.  SPSS~\cite{spss} is used to perform the statistical analyses and a significance level of 0.05 is used throughout the study.

\section*{Results and discussion}

\subsection*{Validation of simulations}

OpenSim model and simulations are validated according to the comprehensive procedures detailed by Hicks~\etal~\cite{Hicks2015} and utilized by Dembia~\etal~\cite{Dembia2017} to verify simulations of assistive devices.

This study builds upon the verified adjusted models, adjusted kinematics, and processed ground reaction forces made available by~\cite{Dembia2017}. Furthermore, the muscular activations resulting from the simulations of unassisted subjects with these data points have been experimentally validated via electromyography studies as presented in~\cite{Hicks2015,Dembia2017}. 

The \textit{loaded} and \textit{\textit{noload}} joint kinematics and kinetics have also been compared with the results of Huang and Kuo~\cite{Huang2014} and Silder~\etal~\cite{Silder2013} and validated qualitatively. Since our simulations of the unassisted subject for \textit{loaded} and \textit{noload} conditions are the same as those of Dembia~\etal~\cite{Dembia2017}, our results have been verified by the reproduction of their simulation results. 

Another source of error during simulations is the kinematics error, which has been shown to be within the recommended thresholds in~\cite{Dembia2017}. Since the inverse kinematics stage of the simulation has not been reproduced in this study, the markers error are not examined, and we rely on the previously performed verification of this error source. The analysis in~\cite{Dembia2017} on residual errors indicate that the residual forces lie below the threshold recommended by Hicks~\etal~\cite{Hicks2015}; however, the residual moments exceed these thresholds. However, since the joint moments matched with those in~\cite{Hicks2015}, it is claimed that these exceeding residual moments do not affect the interpretations~\cite{Huang2014,Silder2013}.

Another error source in these simulations is the additional moments introduced to compensate for any unmodeled passive structures and muscle weakness. These values have been checked to confirm that they are within their recommended thresholds of less than 5\% of net joint moments, in terms of peak and RMS values~\cite{Dembia2017}.

To ensure that our simulations in both ideal and multi-criteria optimization stages do not deviate from the defined error source thresholds, we have also analyzed the kinematics of all simulations and checked their divergence from the adjusted kinematics resulting from the RRA simulations. Additionally, some simulations of the multi-criteria optimization stage have been selected randomly, and their residual and reserve moments and forces are verified to be within the allowable thresholds.

\subsection*{Ideal exoskeleton optimization results}

In this subsection, we present a simulation-based \emph{single-criterion} optimization of assistive torques of mono-articular and bi-articular exoskeletons to maximize the metabolic cost reduction of subjects walking under different loading conditions, without considering actuator saturations. Note that under such ideal conditions, there exists a linear mapping between the mono-articular and bi-articular exoskeleton torques/velocities; hence, optimal assistance torques for both devices are identical. While the total work done by the \emph{ideal} exoskeleton configurations are also identical, due to significantly different kinematic arrangements, the power profiles; hence, the absolute power of mono-articular and bi-articular can be significantly different.

\subsubsection*{Device performance}

Both the bi-articular and mono-articular \emph{ideal exoskeletons} result in significant metabolic cost reduction of subjects walking under different loading conditions. In particular, the mono-articular and bi-articular configurations of the exoskeleton reduce the gross whole-body metabolic cost of the subject under the \textit{noload} condition by 22.38~$\pm$~4.91\% and 22.47~$\pm$~4.89\%, respectively. Similarly, the bi-articular and mono-articular exoskeletons decrease the metabolic rate of subjects under the \textit{loaded} condition by 20.49~$\pm$~2.87\% and 20.45~$\pm$~2.81\%, respectively.  As expected, due to the linear mapping between their kinematics and lack of any saturation, both ideal exoskeletons achieve the same metabolic cost reduction performance, with a small margin of experimental variation.

\begin{figure*}[h!]
	\centering
	\includegraphics[width=\linewidth]{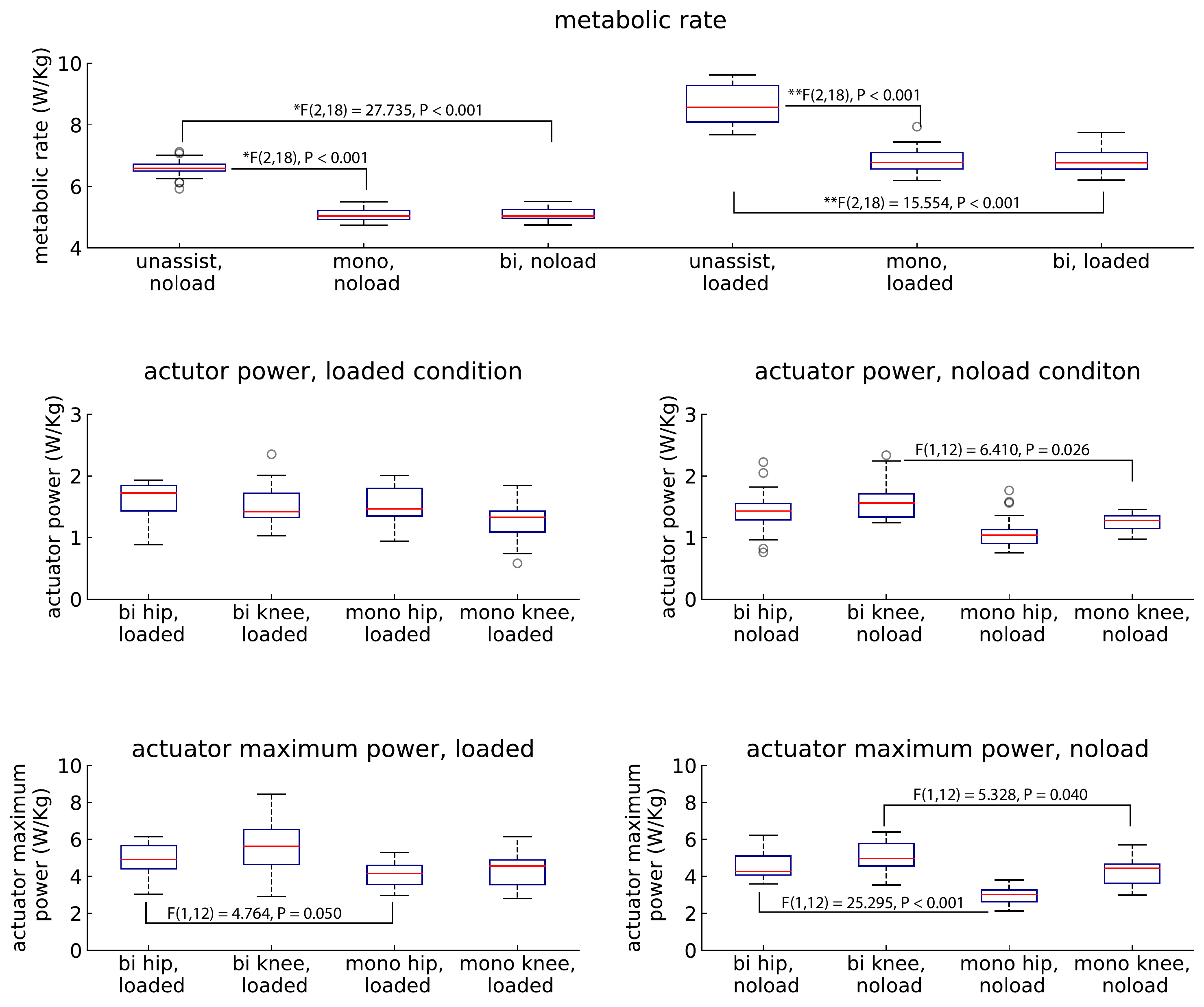}
	\vspace{-3mm}
	\caption{\small{\textbf{The average absolute power consumption of exoskeletons, maximum positive power of exoskeletons and  metabolic rate of subjects.} The average absolute and the positive peak power consumption of assistive actuators at each joint and their effect on whole-body metabolic rate of the subjects while walking under noload and loaded conditions. Superscripts indicate statistically significant differences over 7 subjects with 3 trails, with Tukey Post-hoc test and $P < 0.05$.}}
	\label{Fig_IdealExo_Energy_BoxPlot}
\end{figure*}

The assistance torques provided by both ideal exoskeletons to subjects carrying a heavy load can compensate for the additional metabolic cost due to the heavy load, as presented in the first row of Fig~\ref{Fig_IdealExo_Energy_BoxPlot}. Two-way repeated measures ANOVA (random effect: subjects; fixed effects: loading condition and device configuration) indicates there exists no statistically significant difference between the metabolic cost of the assisted subjects under \textit{loaded} condition and the metabolic cost of unassisted subjects under \textit{noload} condition, meaning that the best the ideal exoskeletons can achieve is to compensate for the metabolic cost of the additional load.

\begin{table}[t]
	\centering\scriptsize
	\renewcommand{\arraystretch}{1.2}
	\begin{adjustwidth}{-1.5in}{0in}
	\caption{\small \bf Average power and maximum positive power consumption of bi-articular and mono-articular ideal devices.}
	\begin{tabular}{c|c|c!{\vrule width 1pt }c|c}
		\toprule[1pt]
		{\bf Loading condition}&\multicolumn{2}{c!{\vrule width 1pt }}{\bf NoLoad Condition} &\multicolumn{2}{c}{\bf Loaded Condition }  \\
		\midrule
		\multirow{2}{*}{\bf Device} & Average absolute power & Maximum positive power & Average absolute power & Maximum positive power \\
								& consumption (W/kg) & (Cost of carrying)  (W/kg) & consumption (W/kg)& (Cost of carrying)  (W/kg)\\
		\midrule
		{\bf Bi-articular} & 3.00~$\pm$~0.32$^{\S}$ & 9.76~$\pm$~1.04$^{\ddagger}$ & 3.11~$\pm$~0.25 & 10.52~$\pm$~1.79$^{*}$  \\
		\midrule
		{\bf Mono-articular} & 2.33~$\pm$~0.29$^{\S}$ & 7.17~$\pm$~0.85$^{\ddagger}$ & 2.75~$\pm$~0.55 & 8.32~$\pm$~1.45$^{*}$ \\
		\bottomrule[1pt]
		\multicolumn{5}{l}{}\\
		\multicolumn{5}{l}{\small{ Note: Identical superscripts indicate a pairwise statistically significant difference with $p<0.05$.}}\\
		\multicolumn{5}{l}{\small{$\pmb{*:}$ $F(1,12) = 6.633$, $P = 0.024$; $$\pmb{\S:}$$ $F(1,12) = 6.410$, $P = 0.026$; $\pmb{\ddagger:}$ $F(1,12) = 29.304$, $P < 0.001$ }}\\
	\end{tabular}
	\label{Table_Avg_Max_PowerConsumption_IdealExo}
\end{adjustwidth}
\end{table} \normalsize \vspace{2mm}

Due to the linear mapping between mono-articular and bi-articular exoskeleton configurations; the optimal assistance torque and the total work done by both of the \emph{ideal} exoskeleton configurations are identical. However, due to different kinematic arrangements, the power profiles; hence, the average absolute power consumption of mono-articular and bi-articular exoskeletons can be different. A repeated measures ANOVA between the bi-articular and mono-articular ideal exoskeleton configurations shows that there exists no statistically significant difference between the average absolute power consumption under no load and loaded conditions for both devices, indicating that the absolute power consumptions of devices are not considerably affected by the loading of the subjects.
The average absolute power consumption of mono-articular and bi-articular exoskeletons  are statistically significantly different from each other during the noload walking condition, as shown in the second row of Fig~\ref{Fig_IdealExo_Energy_BoxPlot}. This difference in absolute power becomes less and loses statistical significance under the loaded walking condition.

The absolute power consumptions of actuators under different loading conditions demonstrate a relatively balanced distribution of absolute power consumption between the knee and hip  actuators for both exoskeleton configurations. The absolute power consumption of the knee actuator of the bi-articular exoskeleton is significantly higher than that of the mono-articular exoskeleton for loaded walking, while this difference is lost under the loaded walking condition. Note that, given the total work done by both ideal exoskeletons are identical, a larger absolute power consumption indicates a larger amount of negative work done, some of which may be captured by regeneration.

Table~\ref{Table_Avg_Max_PowerConsumption_IdealExo} summarizes the average absolute power consumption along with the cost of carrying metric.
A one-way repeated measures ANOVA between the bi-articular and mono-articular ideal exoskeleton configurations shows that the maximum positive power consumption of devices are statistically significantly different under both loading conditions, as shown in the last row of Fig~\ref{Fig_IdealExo_Energy_BoxPlot}. For ideal devices, the costs of carrying for the hip and knee actuators of the mono-articular exoskeleton are significantly lower than that of the bi-articular configuration under the \emph{noload} walking condition, while only the cost of carrying for the hip actuator of the mono-articular exoskeleton is significantly lower during the \emph{loaded}  walking condition.

\begin{figure*}[ht]
	\centering
	\includegraphics[width=\linewidth]{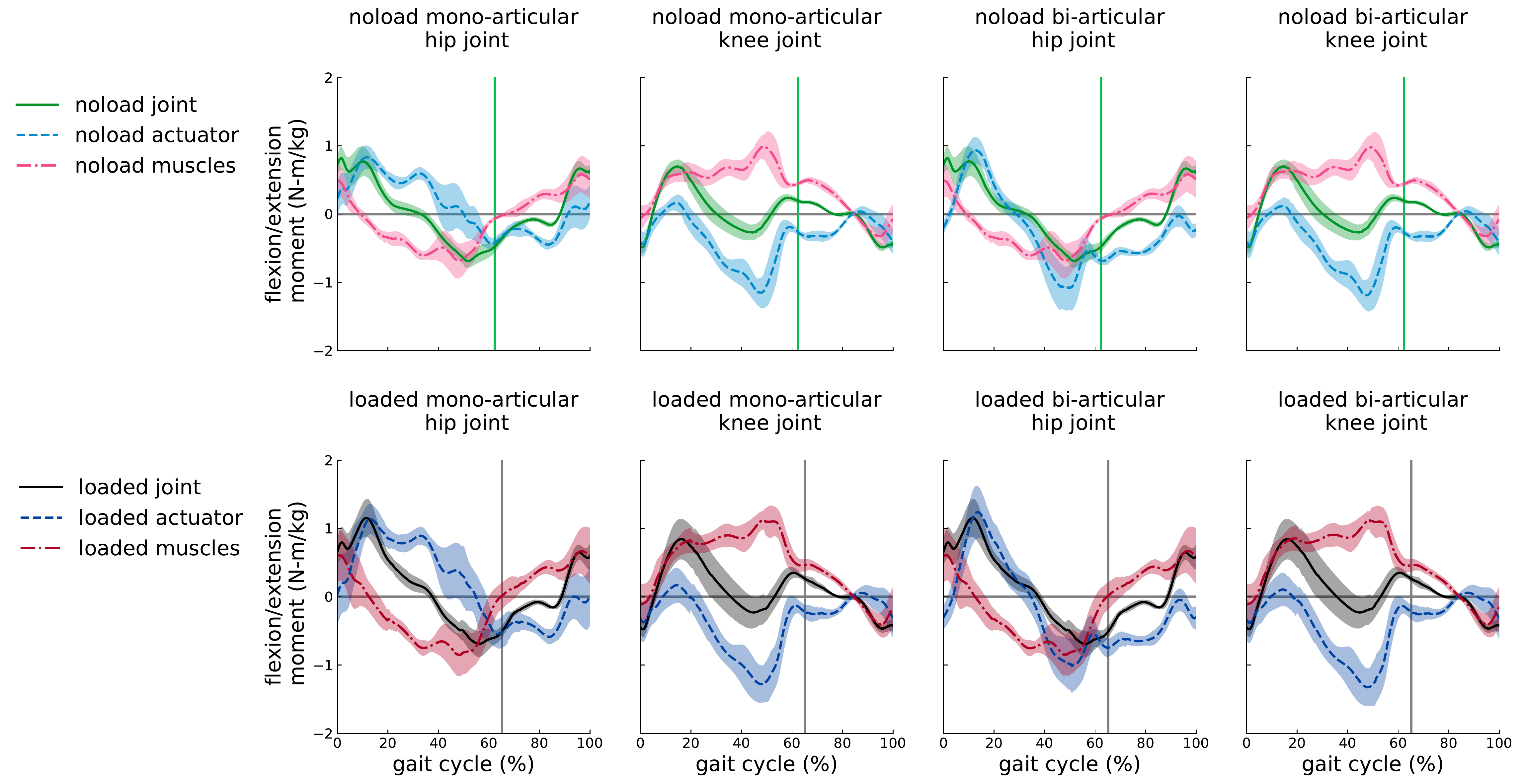}
	\vspace{-3mm}
	\caption{{\small\textbf{Assistance torque profiles and muscle generated moment at each joint under noload and loaded walking conditions.} The assistive torques for subjects walking under \textit{noload} (blue) and \textit{loaded}  (dark blue) conditions, the net joint moments generated by unassisted muscles for walking under \textit{noload} (green) and \textit{loaded} (black) conditions, and the moment generated  by assisted muscles under \textit{noload} (rose red) and \textit{loaded} (dark rose red) conditions are shown for each actuator of the exoskeletons. The curves are averaged over simulations of 7 subjects with 3 trials and normalized by the subject mass; shaded regions around the mean profiles indicate the standard deviations.}}
	\label{Fig_IdealExo_Torque}
\end{figure*} 

\subsubsection*{Assistance torque and power profiles}

The knee actuators of both exoskeletons exhibit identical torque profiles, while the torque profiles of the hip actuators differ significantly due to the kinematic arrangement resulting in different reaction torques, for both \textit{noload} and \textit{loaded} walking conditions, as depicted in Fig~\ref{Fig_IdealExo_Torque}.

Due to the linear mapping between their kinematics, the optimal assistance torque profiles of both ideal exoskeletons configurations are identical, as expected.
These results are also in good agreement with the profiles in~\cite{Dembia2017,Uchida2016_idealexo_running}. The optimal assistance torque profiles do not resemble the net moments of the assisted joints, as already reported in~\cite{Dembia2017,Uchida2016_idealexo_running} for ideal exoskeletons assisting loaded walking and unloaded running, respectively. In particular, the assistive torques provided to both the hip and knee joints exceed the corresponding net joint moments and oppose the muscle generated moments about these joints. The opposition is more emphasized at the knee joint than at the hip joint during the mid-stance to the mid-swing phase, with the highest opposition taking place at the onset of the pre-swing phase. The hip joint has significant actuator and muscle torque opposition during the pre-swing to the terminal swing phases, indicating that different from the knee joint, in which a major portion of antagonism occurs during the stance phase, the hip assistance displays  muscle and actuator torque opposition during the swing phase.

The assistance torque profiles of ideal exoskeletons under \textit{noload} and \textit{loaded} conditions indicate that the loading subjects with a heavy load results in predictable changes in the torque profiles of the assistive devices. The main differences between the assistance torque profiles \textit{noload} and \textit{loaded} conditions are some scaling of the magnitude and a shift in the timing  of the assistive torque profiles. The standard deviation of assistive torques and net joint torque generated by the assisted muscles are considerably greater under the \textit{loaded} walking condition, with this difference becoming evident at the knee joint, where the net joint moment displays a large standard deviation during the stance phase. The high within-subject deviation of torque profiles of subjects carrying a heavy load may motivate subject-specific design and control of the exoskeletons, as proposed in~\cite{Uchida2016_idealexo_running}.

\begin{figure*}[ht]
	\centering
	\includegraphics[width=\linewidth]{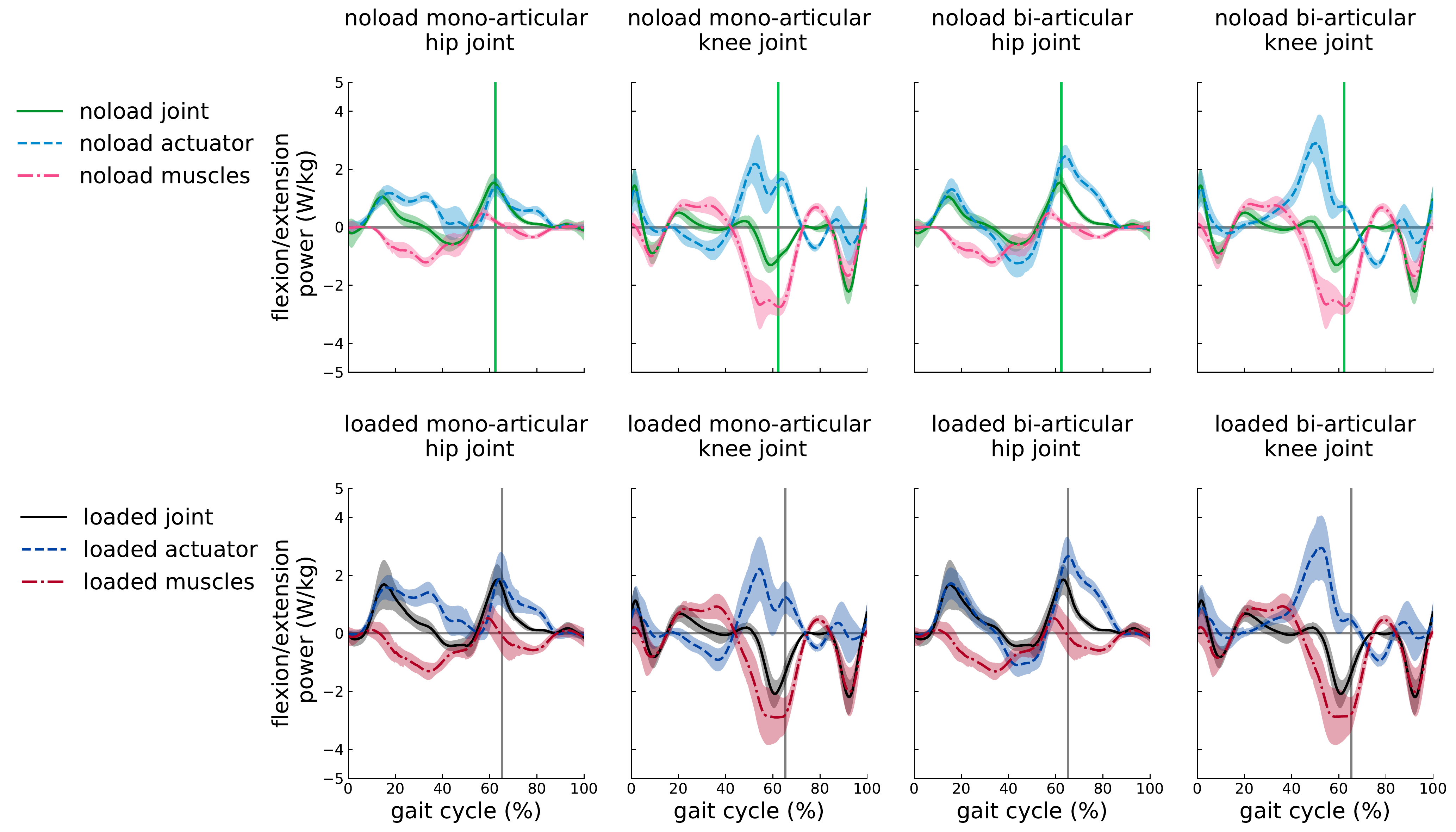}
	\vspace{-3mm}
	\caption{{\small\textbf{Power profiles of exoskeletons and corresponding net power at the joint.} The power profiles of exoskeletons for subjects under \emph{noload} (blue) and \emph{loaded}(dark blue) walking conditions, and the net joint power profiles for \textit{noload} (green) and \textit{loaded} (black) conditions are shown for each actuator of the exoskeletons. The curves are averaged over simulations of 7 subjects with 3 trials and normalized by the subject mass; shaded regions around the mean profiles indicate the standard deviations.}}
	\label{Fig_IdealExo_Power}
\end{figure*}

While both ideal mono-articular and bi-articular exoskeletons provide the same assistance to joints, the distribution of power between the knee and hip actuators is different, as presented in Fig~\ref{Fig_IdealExo_Power}. The power consumptions of the actuators of the bi-articular exoskeleton are different from the mono-articular exoskeleton throughout the gait cycle, except during the loading response and, partially, during the mid-stance phase. The load carried by subjects causes a shift in timing and  amplification of the magnitude of power profiles.

Comparing the power profiles of the mono-articular exoskeleton to those of the bi-articular exoskeleton shows that the bi-articular exoskeleton has higher peak positive power consumption compared to the mono-articular exoskeleton during the terminal stance and the pre-swing phases. This difference between the power profiles of the exoskeletons can explain the observed statistical significance in Table~\ref{Table_Avg_Max_PowerConsumption_IdealExo} between the maximum positive power of the exoskeletons in both loading conditions.

Furthermore, the bi-articular exoskeletons demonstrate a higher potential for energy regeneration than the mono-articular devices. In particular,  the negative mechanical work in the bi-articular devices can be harvested during the initial-swing and mid-swing phases for the knee actuator and the terminal stance phase for the hip actuator, which can be deduced from Fig~\ref{Fig_IdealExo_Power}. Unlike the bi-articular devices, the mono-articular hip actuator performs practically no negative mechanical work, and the regeneratable work of the mono-articular knee actuator takes place during the mid stance to terminal stance and the initial swing phases.

\subsubsection*{Effect of ideal assistance torques on the coordination of muscles}

\begin{figure*}[ht!]
	\centering
	\subfloat[\small{bi-articular muscles}]{\includegraphics[width=1.75in]{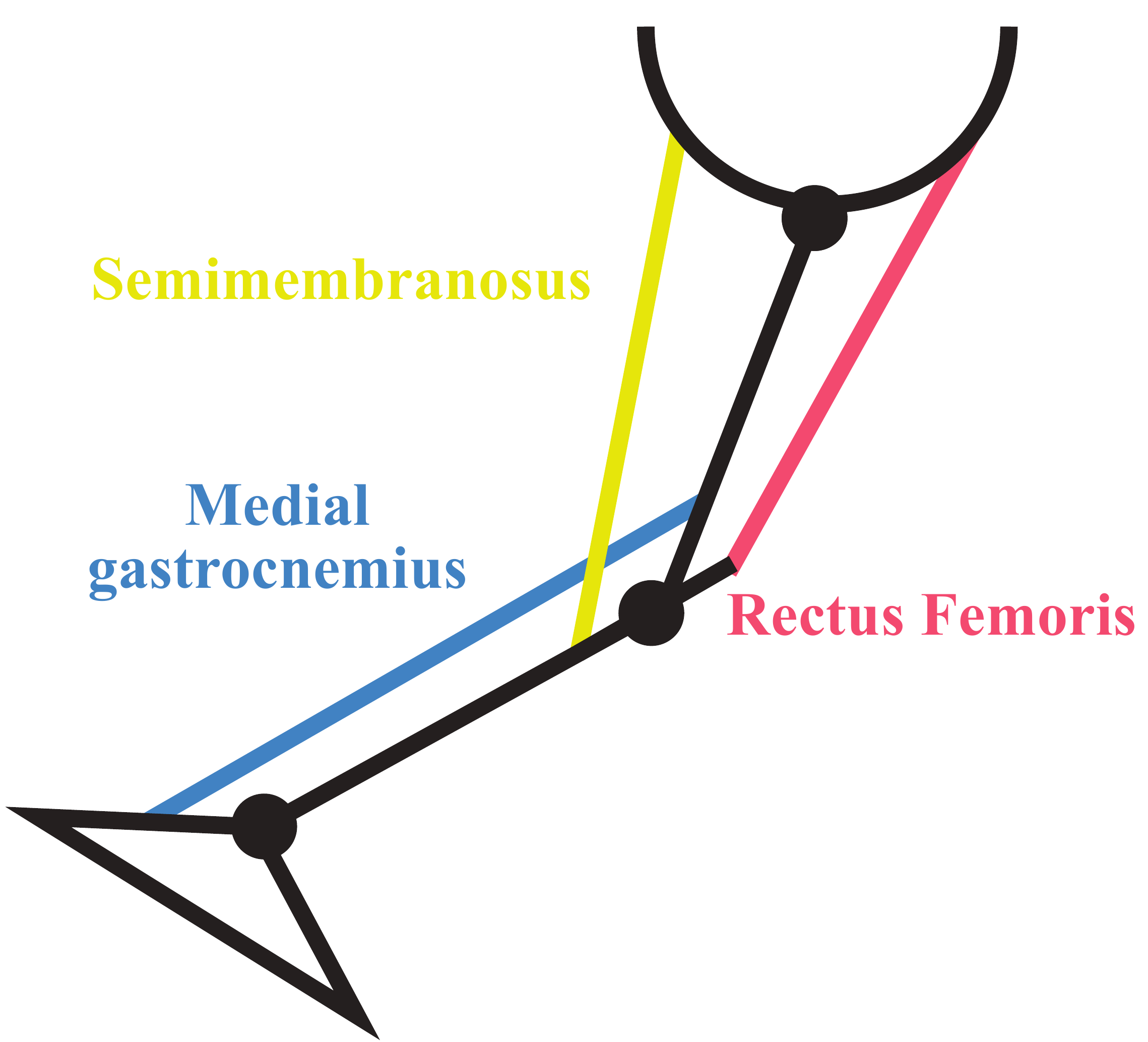}
		\label{Fig_Bi_Muscles}}
	\hfil
	\subfloat[\small{mono-articular muscles}]{\includegraphics[width=1.75in]{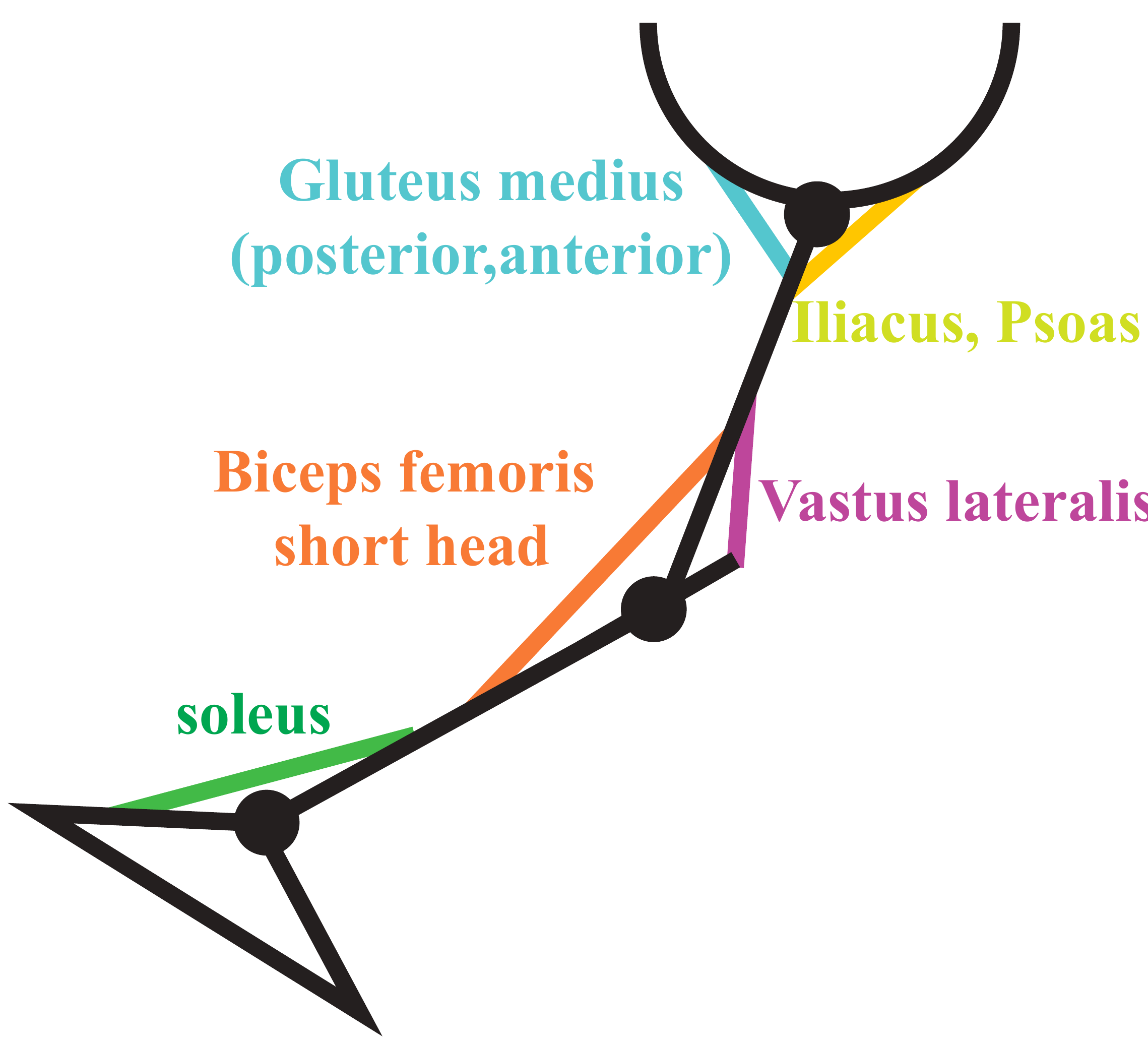}
		\label{Fig_Mono_Muscles}}
	\caption{\small{\textbf{Representative of bi-articular and mono-articular lower extremity muscles}}}
	\label{Fig_Muscles}
\end{figure*}

The muscular activation of the subjects assisted by ideal exoskeletons is highly different from the muscle activation of unassisted subjects. Adding a set of ideal actuators and promoting their use, the static optimizer of the musculoskeletal simulation maximizes the use of assistive torques to track the kinetics and kinematics of the joints. However, appending ideal actuators to provide assistive torques does not necessarily decrease the activity of all muscles, as it may be more economical to increase the activity of specific muscles  to decrease the activity of less cost-effective muscles. Since the metabolic power of muscles is a function of their activity and their fiber properties~\cite{Uchida2016_metabolic_model}, the reduction in the activity of the entire set of recruited muscles results in a gross whole-body metabolic cost reduction.  Fig~\ref{Fig_Muscles} presents a set of representative mono-articular and bi-articular muscles at the lower extremity.

\begin{figure*}[ht]
	\centering
	\includegraphics[width=\linewidth]{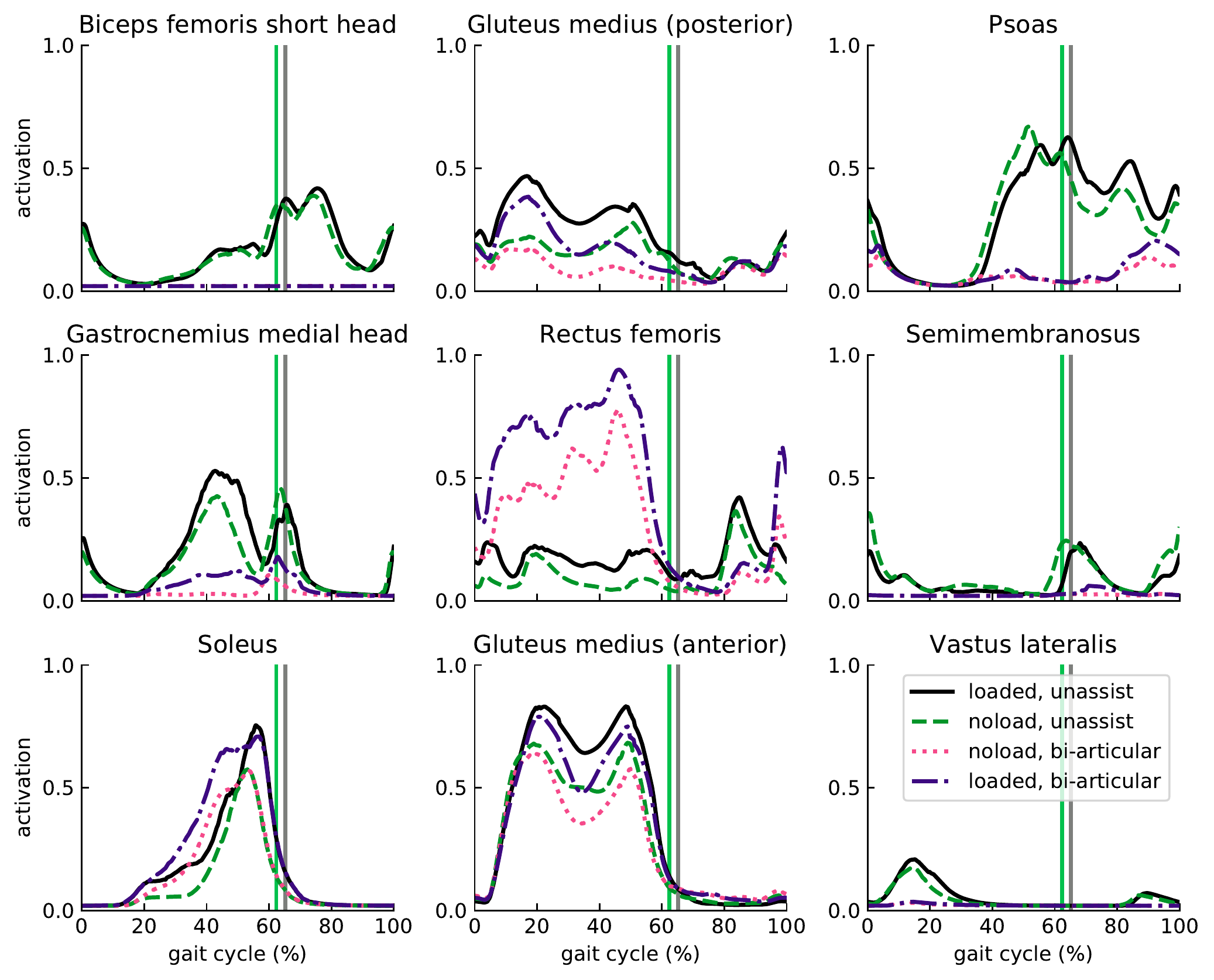}
	\vspace{-3mm}
	\caption{{\small\textbf{Activation of representative lower limb muscles of assisted and unassisted subjects.} The activation of unassisted subjects under \emph{noload} (green) and \emph{loaded} (black) walking conditions, and assisted subjects under \textit{noload} (pink) and \textit{loaded} (dark violet) walking conditions are shown for nine important muscles. The curves are averaged over 7 subjects with 3 trials.}}
	\label{Fig_IdealExo_MusclesActivation}
\end{figure*}

Despite the kinematic difference between the two configurations of the assistive devices, since the ideal assistance torques applied to the joints are practically identical, they result in an identical effect on the muscular activation of the subjects. Note that this result is valid only for ideal devices, assuming that there are no constraints on the actuator torques and the devices lack mass and inertial effects. The muscle activations in Fig~\ref{Fig_IdealExo_MusclesActivation} show the effect of the ideal exoskeletons on the muscle activation of a set of representative muscles.

The assistive torques provided by the ideal exoskeletons affect the activity of muscles of the lower extremity. The decrease in activation of bicep femoris short head, semimembranosus, and vasti muscles is significant. Some portion of the torque generated by these muscles is taken over by other muscles and/or the assistance torques simulated via ideal actuators. The activation of the rectus femoris, which is a large knee extensor and a hip flexor bi-articular muscle, is considerably increased during the stance phase. This increase occurs so that the optimizer can take advantage of the high force-generation capacity of the rectus femoris to exert hip flexion and knee extension moments more economically. In the meanwhile, this high muscular activity of the rectus femoris results in high knee extension and hip flexion moments exceeding the net joint moments, and this effect is counteracted by the assistance torques, which are modeled as
ideal actuators whose use is favored for the application of high torques.

An activation, in which the moment required to execute the hip flexion and the knee extension can be applied by a set of more cost-effective muscles and ideal actuators, results in a substantial reduction in the activity of psoas and iliacus muscles, which serve as two major hip flexor muscles, and the vasti muscles (vastus lateralis, vastus intermedius, and vastus medialis) which constitute the knee extensor muscles. The activity of semimembranous, another bi-articular muscle that contributes to hip extension and knee flexion moments, is also significantly decreased by the assistive torques, practically replacing the whole of its activity. The activity of the medial gastrocnemius, serving as a critical knee flexor and ankle plantar flexor muscle, is also substantially reduced by the assistive torques, yet the muscle remains partially active to supply an ankle plantarflexion moment.

The activity reduction of ankle plantar-flexion moment is compensated by increasing the soleus activity as another primary ankle plantarflexor muscle. The assistive torques affect the activity of the gluteus medius muscles as well, which are not only responsible for a significant fraction of hip abduction moment, but also contribute to hip motion in the sagittal plane, as well as the hip rotation. The anterior and posterior portions of the gluteus medius muscle support the hip extension and flexion and its lateral and medial rotations, in addition to their primary contribution to the hip abduction. The reduction in co-contraction of the hip rotation due to modified muscle coordination of assisted subjects results in a reduction of the muscular activity of the gluteus medius muscle. It is important to note that, while the assistance torques are only provided to the hip and knee joints, these torques also affect the muscular activity of the ankle joint and the muscles responsible for the hip abduction.

The main differences between the muscular activity of the subjects under \emph{noload} and \emph{loaded} walking conditions are the scaling of the magnitude and a shift in the timing of the muscular activations. This result seems to be consisted over the muscle activation profiles. However, the increased activation under the \emph{loaded} walking condition becomes more emphasized for certain muscles, such as the semimembranosus, the muscle activation of which can no longer be entirely replaced by the assistance torques, as in the \emph{noload} case.

\subsubsection*{Effect of ideal assistance torques on the joint reaction forces/moments}

The change in the muscle activations and the assistive torques provided to the subjects also affect the reaction forces/moments at both the assisted and the unassisted joints. The relationship between the muscle activity and the joint reaction forces has been established in the literature~\cite{DeMers2014,VanVeen2019}. Due to the assistance provided, the modified activation of the muscles at the ankle joint reduces the reaction forces/moments of this joint during the swing phase, while slightly increasing the reaction forces/moments at the ankle  during the stance stage, as shown in Figs~B1 and~B2 in~\nameref{S3_Appendix}. The effect of muscle recruitment is evident in the medial-lateral reaction force and the extension-flexion reaction moment at the ankle. Veen~\etal~\cite{VanVeen2019} have shown that an increase in the activation of the rectus femoris and gastrocnemius muscles along with a decrease in the activation of the soleus muscle can reduce the reaction forces of the ankle joint. Although the assistive torques increase the activation of the rectus femoris, the effect of these assistive torques on the gastrocnemius and soleus muscles is not favorable in reducing the reaction force, especially during the stance phase. This coordination of muscles explains the behavior of the reaction moments/forces at the ankle joint.

\begin{figure*}[ht]
	\centering
	\includegraphics[width=\linewidth]{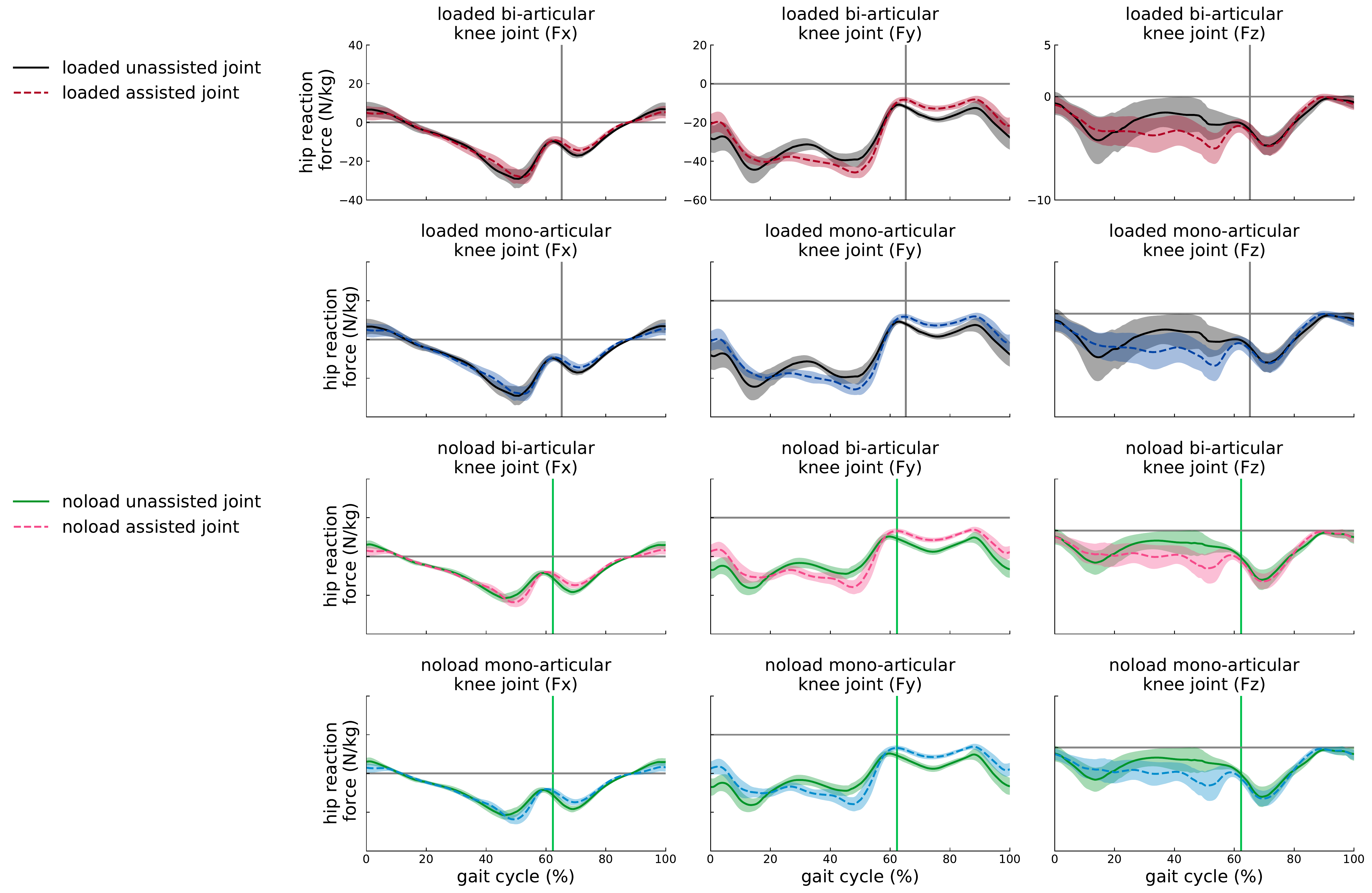}
	\vspace{-3mm}
	\caption{{\small\textbf{Reaction forces of knee joint of assisted and unassisted subjects.} The knee joint reaction forces of unassisted subjects under \emph{noload} (green) and \emph{loaded} (black) walking conditions, and assisted subjects by bi-articular/mono-articular exoskeleton during \textit{noload} (rose red/blue) and \textit{loaded} (dark rose red/dark blue) walking conditions are shown in anterior/posterior, proximal/distal, and medial/lateral directions. The curves are averaged over 7 subjects with 3 trials.}}
	\label{Fig_IdealExo_Knee_JRF}
\end{figure*}

The effect of altered muscle recruitment due to assistive torques on the reaction moments/forces at the patellofemoral and the knee joints is substantial. The reaction forces at the knee joints decrease considerably during the loading response, and initial swing phases by 45.53~($\pm$~24.89\%), 33.64~($\pm$~6.37\%), and 24.63~($\pm$~18.16\%) in anterior/posterior, proximal/distal, and medial/lateral directions during the \emph{noload} walking condition, respectively, as shown in Fig~\ref{Fig_IdealExo_Knee_JRF}. Unlike the \emph{noload} walking condition, the maximum reduction of the peak reaction force in anterior/posterior, proximal/distal, and medial/lateral directions of the knee joint of subjects assisted by the ideal exoskeletons occur during the loading response, terminal swing, and mid-swing phases of \emph{loaded} walking, by reducing the peak reaction forces by 22.93~($\pm$~26.85\%), 29.76~($\pm$~5.20\%), and 12.93~($\pm$~27.83\%), respectively.

\begin{figure*}[ht]
	\centering
	\includegraphics[width=\linewidth]{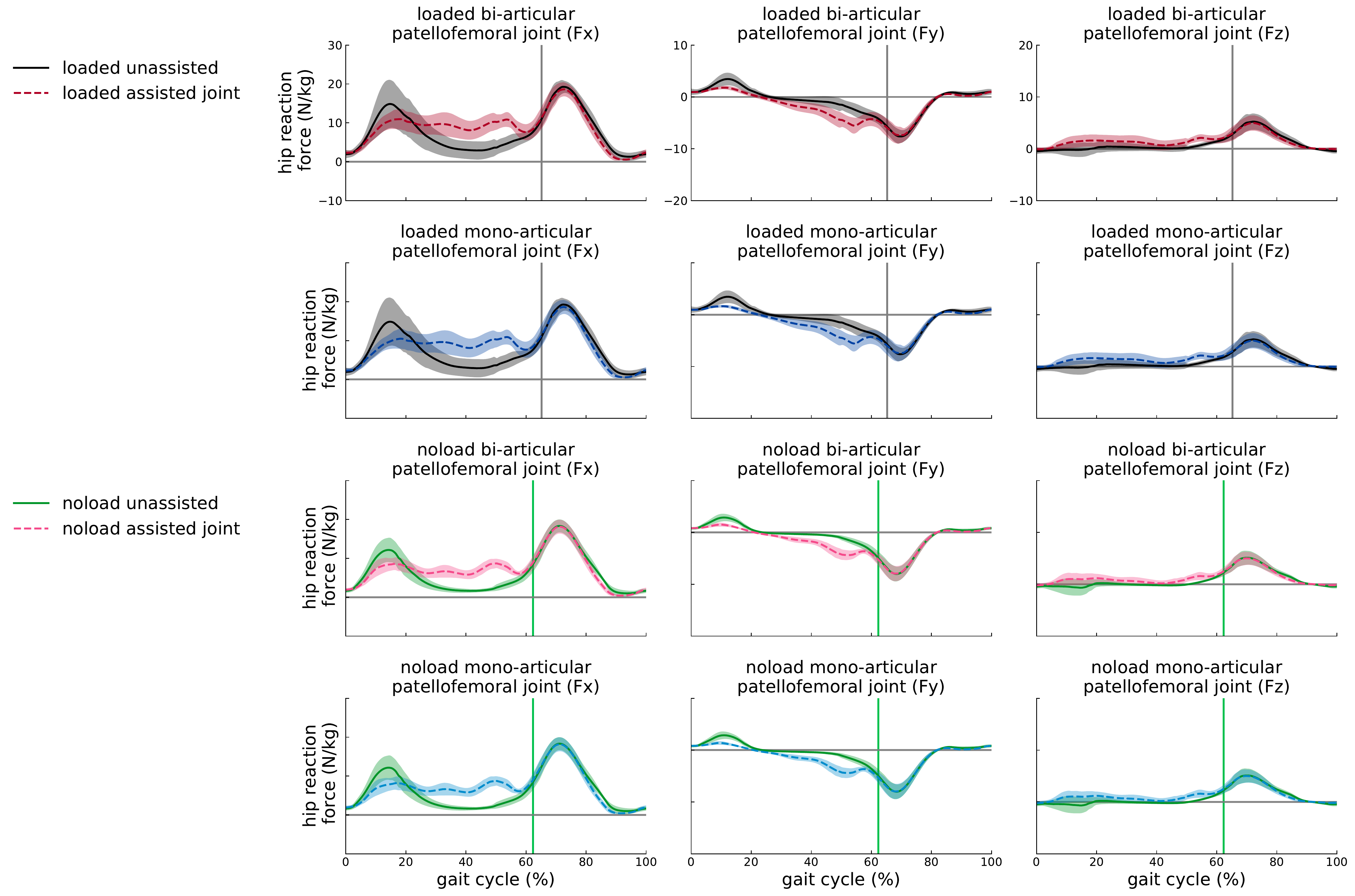}
	\vspace{-3mm}
	\caption{{\small\textbf{Reaction forces of patellofemoral joint of assisted and unassisted subjects.} The patellofemoral joint reaction forces of unassisted subjects under \emph{noload} (green) and \emph{loaded} (black) walking conditions, and assisted subjects by bi-articular/mono-articular exoskeleton during \textit{noload} (rose red/blue) and \textit{loaded} (dark rose red/dark blue) walking conditions are shown in anterior/posterior, proximal/distal, and medial/lateral directions. The curves are averaged over 7 subjects with 3 trials.}}
	\label{Fig_IdealExo_Patellofemoral_JRF}
\end{figure*}

The maximum reduction of the peak reaction force in anterior/posterior, proximal/distal, and medial/lateral directions of the patellofemoral joint of subjects assisted by the ideal exoskeletons occur during the loading response, mid-stance, and terminal swing phases of the \emph{noload} walking condition, by reducing the peak reaction force by 28.35~($\pm$~9.17\%), 48.58~($\pm$~7.17\%), and 30.22~($\pm$~20.10\%), respectively. A similar analysis for the \emph{loaded} walking condition shows that the highest reduction of reaction forces in the patellofemoral joint take place during the loading response, and terminal swing phases by reducing the peak reaction force by 21.35~($\pm$~20.89\%) in the anterior/posterior, 42.98~($\pm$~13.94\%) proximal/distal, and 20.77~($\pm$~51.33\%) medial/lateral directions, as shown qualitatively in Fig~\ref{Fig_IdealExo_Patellofemoral_JRF}.

Although the standard deviation of these reductions in anterior/posterior and medial/lateral directions are large, indicating a requirement of subject-specific assistance strategies, the reduction of the reaction forces in proximal/distal with low within-subject variation along with high reaction forces of the joints in this direction shows promise for the assistance strategy to reduce the load on the joints. The knee and patellofemoral reaction moments are also shown in Figs~B3 and~B4 in~\nameref{S3_Appendix}.

The analysis of the tibiofemoral forces shows that the hamstring muscles significantly impact the reaction forces of the knee during the early stance phase, while the gastrocnemius, rectus femoris, and iliopsoas muscles affect the reaction forces during the late stance phase~\cite{DeMers2014,VanVeen2019}.
The increase in the activation of the soleus and the decrease in the activation of the hamstring muscles (i.e., semimembranosus, semitendinosus, and biceps femoris muscles) reduce the reaction force of the knee in the early stance phase. During the late stance phase, we hypothesize that the substantial promotion and reduction of the rectus femoris and gluteus medius activities, sequentially, become dominant to the reduction of activities of other muscles and result in an increase in the tibiofemoral reaction force. Since the behavior of the other reaction force components in both the patellofemoral and knee joints is practically identical to the tibiofemoral case, it seems logical that the muscle arrangement has similar effects on the other reaction forces.  Although the reaction moments in both joints follow a similar trend with the reaction forces, the effect of devices on the reaction moments is slightly different in that the bi-articular exoskeleton is able to reduce the reaction moments and has lower peaks compared to the mono-articular exoskeleton in terms of the extension-flexion reaction moment. While the effect of the hip muscles on the knee reaction force has already been established in the literature~\cite{DeMers2014,VanVeen2019}, above observations still need to be verified under more isolated conditions, such as while physically assisting a single joint.

Although the reaction forces/moments at the knee joint increase during the late stance phase, the assistive torques reduce the peak reaction forces/moments at the knee joint. Additionally, the effect of the modified muscle recruitment on the reaction forces/moments during the swing phase is significantly lower than their effect during the stance phase, while the tibiofemoral displays considerable reaction force reduction even during the swing phase.

The reaction forces of the hip joint are affected by the activity of a group of muscles, including the gluteus minus, gluteus medius, iliopsoas, and rectus femoris muscles, as detailed in~\cite{VanVeen2019}. The increase in the activity of the rectus femoris incorporation with the iliopsoas and gluteus medius muscle activity reduction decrease the reaction forces at the hip joint. This reduction is significant during the late stance and the early swing phases, and the effect on the subjects under the \emph{noload} walking condition is more dominant compared to the subjects under the \emph{loaded} walking condition, as shown in Fig~B5 in~\nameref{S3_Appendix}.

The peak reaction force reduction in anterior/posterior and proximal/distal directions of the hip joint of subjects assisted by the ideal exoskeletons occurs during the initial swing phase of walking in both loading conditions, reducing the peak reaction forces by 72.60~($\pm$~12.57\%) and 46.47~($\pm$~4.81\%) during the \emph{noload} walking condition, and 61.06~($\pm$~20.68\%) and 45.46~($\pm$~7.51\%) during the \emph{loaded} walking condition along the anterior/posterior and proximal/distal directions. On the other hand, the peak reaction force in medial/lateral direction occurs during the terminal stance phase of walking in both loading conditions by reducing 58.86~($\pm$~8.50\%) and 50.86~($\pm$~12.68\%) during \emph{noload} and \emph{loaded} walking conditions, respectively.

Overall, the effect of ideal assistance torques on joint reaction forces/moments is significant and these modifications in the reaction forces/moments of the assisted subjects can improve the health of joint tissues~\cite{Carter1988}. The large joint loads are identified as an essential factor of onsetting and progressing osteoarthritis~\cite{Baliunas2002,Sharma1998} and joint pain~\cite{Schnitzer1993}, and a reduction in the reaction forces/moments has the potential to prevent or reduce joint pain and osteoarthritis onset.

\subsubsection*{Summary of Ideal Exoskeleton Optimization Results}

The ideal exoskeleton optimization results verify that both the ideal mono-articular and bi-articular devices can reach the same level of metabolic rate reduction, while the total power consumption and the cost of carrying (the peak positive power) for bi-articular and mono-articular devices are significantly different during the \emph{noload} walking condition. The simulations indicate that loading subjects with a heavy backpack results in a predictable change in assistance torque profiles, causing a proportional increase in the magnitude and a time shift. The results also verify that the assistance torques provided by the exoskeletons can indirectly affect the activity of muscles at the ankle joint and the muscles related to the hip abduction. Finally, the simulations indicate that the assistance torques can  significantly decrease the peak reaction forces/moments at the knee, patellofemoral, and hip joints.

\subsection*{Multi-criteria optimization results}

While ideal exoskeleton optimizations in the previous subsection provide many useful insights, different exoskeleton designs cannot be meaningfully compared without introducing some realistic physical limits on the actuator torques and considering the power consumption of these devices. This subsection introduces different levels of peak torque constraints on the actuators of both mono-articular and bi-articular exoskeleton configurations and compares their performance by \emph{simultaneously} optimizing the metabolic cost reduction and the power consumption metrics.

\subsubsection*{Optimal devices performance}

In the previous subsection, the exoskeleton configurations and their effect on the metabolic cost and muscular activation of the subjects are studied under two different load conditions and under the assumption that the exoskeleton actuators have no bounds on the amount of the torque they can supply to the musculoskeletal model. However, this assumption is not a realistic assumption for the real life design and control of assistive devices, because all mechatronics systems, especially the untethered exoskeletons, have some constraints on the amount of torque that their actuation unit can provide and the instantaneous power available from the power unit/battery is limited.

A simulation-based multi-criteria optimization can be posed to analyze the performance of assistive exoskeleton configurations under the actuator saturations to increase real life applicability of the optimization results. Note that, when actuator saturation is introduced, the linear mapping between the different exoskeleton configurations becomes no longer valid and both devices display significantly different assistance characteristics.

\begin{figure*}[ht]
	\centering
	\includegraphics[width=\linewidth]{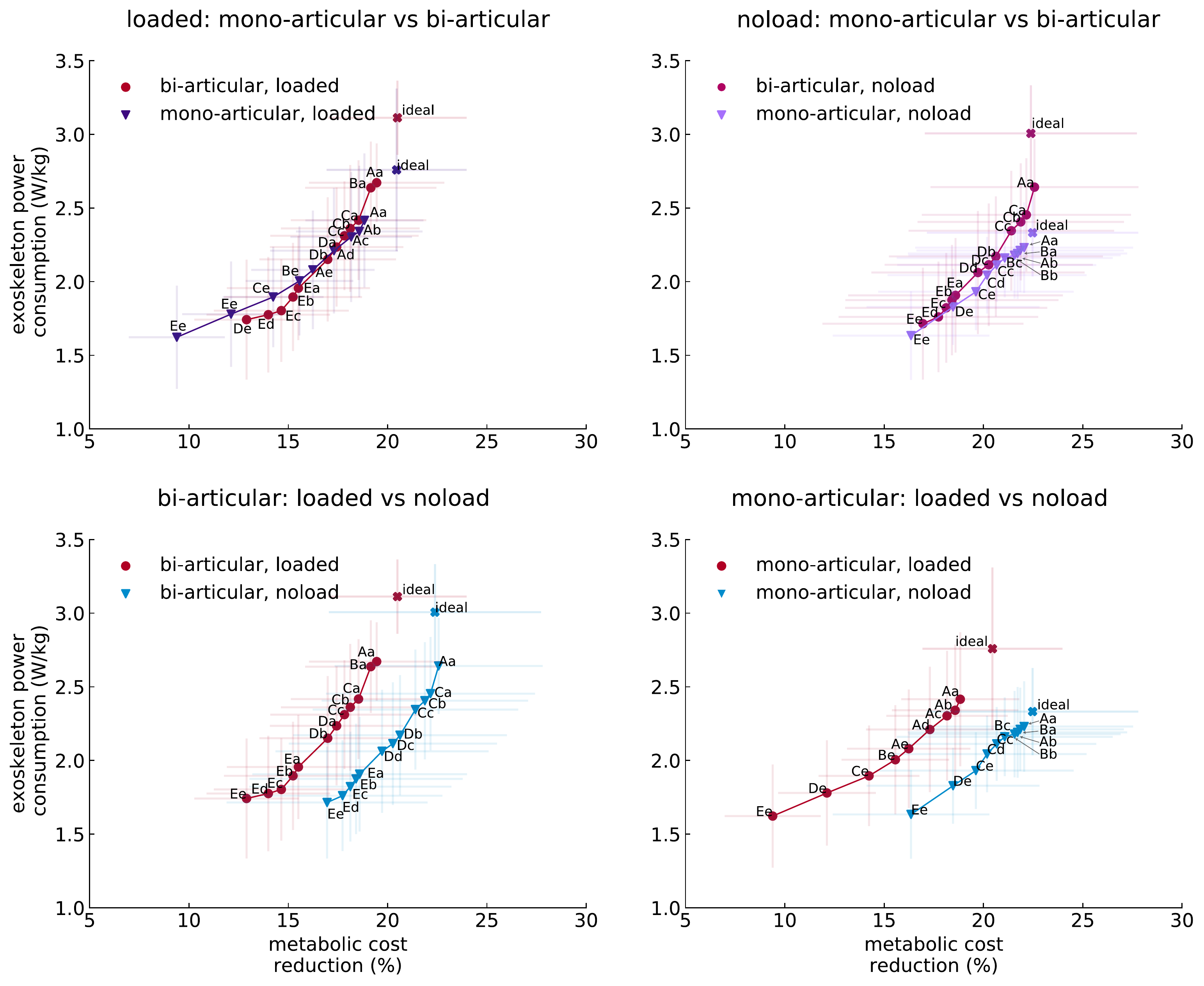}
	\vspace{-4mm}
	\caption{{\small\textbf{Pareto-front curves characterizing the trade-off between metabolic cost reduction and absolute power consumption.} The data points on the Pareto-front curves represent the average and standard deviations over 7 subjects and 3 trials per subjects. The label on each marker denotes results from different peak torque constraints. The hip peak torque constraints are labeled with capital letters from {\it A} to {\it E} to represent the range of torques from 70~Nm to 30~Nm, while the knee peak torque constraints are labeled with lower-case letters from {\it a} to {\it e} to represent the range of torques from 70~Nm to 30~Nm, respectively. }}
	\label{Fig_Main_Paretofronts}
\end{figure*}

To characterize  the trade-off between  the metabolic cost reduction and the device power consumption, a Pareto optimization is performed through the application of the $\epsilon$-constraint method with the OpenSim musculoskeletal simulation framework, by systematically constraining the peak torque of the assistive actuators to introduce an upper bound on the device power consumption. The average Pareto-front curves for the  mono-articular and bi-articular exoskeleton configurations under \emph{noload} and \emph{loaded} walking conditions are presented in Fig~\ref{Fig_Main_Paretofronts}.

One of the immediate results that can be observed from the Pareto-front curves is that the torque-limited exoskeletons with sufficiently large torque capablities at the hip and knee can provide almost the same level of metabolic cost reduction with the ideal devices, while displaying significantly lower power consumption, under both \emph{noload} and \emph{loaded} walking conditions. Table~\ref{Table_Device_Performance_Comparison} presents torque-limited devices that lie on the Pareto-front curve of both exoskeleton configurations which can provide nearly the same level of assistance as the ideal exoskeletons. The power consumption and the cost of carrying of the torque-limited exoskeletons are statistically significantly lower than the ideal devices under both loading conditions, as presented in Table~\ref{Table_Device_MaxPower_Comparison}.

\begin{table}[h!]
	\centering
	\renewcommand{\arraystretch}{1.2}
	\begin{adjustwidth}{-2.3in}{0in} \small
	\caption{\small{\textbf{Power consumption and metabolic cost reduction of ideal and torque-limited exoskeletons.}}}
	\begin{tabular}{c!{\vline width 0.2pt}c!{\vline width 0.2pt}c!{\vline width 0.2pt}c!{\vline width 0.2pt}c!{\vline width 0.2pt}c}
		\toprule
		\multirow{2}{*}{\textbf{Configuration}} & \multirow{2}{*}{\textbf{Device type}} & \multirow{2}{*}{\textbf{Condition}} & \textbf{Hip power}& \textbf{Knee power}  & \textbf{Metabolic cost}\\
		&  &  &\textbf{consumption (W/kg)} & \textbf{consumption (W/kg)} & \textbf{reduction (\%)} \\
		\midrule[0.75pt]
		\multirow{4}{*}{\textbf{bi-articular}} & ideal & \textit{noload} & 1.42~$\pm$~0.32 & 1.58~$\pm$~0.30$^{*}$ & 22.38~$\pm$~4.91 \\
		\cmidrule[0.2pt]{2-6}
		& ideal & \textit{loaded} & 1.58~$\pm$~0.29$^{\dagger}$ & 1.52~$\pm$~0.29$^{\ddagger}$ & 20.49~$\pm$~2.87 \\
		\cmidrule[0.2pt]{2-6}
		& Torque limited "Db"  & \textit{noload} & 1.03~$\pm$~0.31 & 1.03~$\pm$~0.14$^{*}$ & 20.36~$\pm$~5.39 \\
		\cmidrule[0.2pt]{2-6}
		& Torque limited "Db"  & \textit{loaded} & 1.03~$\pm$~0.24$^{\dagger}$ & 1.11~$\pm$~0.24$^{\ddagger}$ & 16.99~$\pm$~3.44 \\
		\midrule[0.75pt]
		\multirow{4}{*}{\textbf{mono-articular}} & ideal & \textit{noload} & 1.09~$\pm$~0.24 & 1.24~$\pm$~0.13$^{\gamma}$ & 22.47~$\pm$~4.89 \\
		\cmidrule[0.2pt]{2-6}
		& ideal & \textit{loaded} & 1.52~$\pm$~0.28$^{\S}$ & 1.24~$\pm$~0.27 & 20.45~$\pm$~2.81 \\
		\cmidrule[0.2pt]{2-6}
		& Torque limited "Ce"  & \textit{noload} & 0.98~$\pm$~0.20 & 1.10~$\pm$~0.15$^{\gamma}$ & 20.67~$\pm$~5.01 \\
		\cmidrule[0.2pt]{2-6}
		& Torque limited "Be" & \textit{loaded} & 1.07~$\pm$~0.17$^{\S}$ & 0.93~$\pm$~0.22 & 15.56~$\pm$~2.71 \\
		\bottomrule
		\multicolumn{6}{l}{\small{Note: Identical superscripts indicate a pairwise statistically significant difference (7 subjects, 3 trails, Tukey post-hoc, $p < 0.05$).}} \\
		\multicolumn{6}{l}{\small{$\pmb{*:}$ $F(1,12) = 6.284$, $P = 0.028$; $$\pmb{\S:}$$ $F(1,12) = 11.889$, $P = 0.005$; $\pmb{\ddagger:}$ $F(1,12) = 4.747$, $P=0.050$; $\pmb{\dagger:}$ $F(1,12) = 12.313 $, $P = 0.004 $.}} \\
		\multicolumn{6}{l}{\small{$\pmb{\gamma:}$ $F(1,12) = 13.591$, $P=0.003$.}}\\
	\end{tabular}%
	\label{Table_Device_Performance_Comparison}
	\end{adjustwidth} \normalsize
\end{table}

%

\begin{table}[h!]
	\centering
\renewcommand{\arraystretch}{1.2}
\begin{adjustwidth}{-2.3in}{0in} \small
	\caption{\small{\textbf{Maximum positive power of ideal and torque-limited exoskeletons.}}}
	\begin{tabular}{c!{\vline width 0.2pt}c!{\vline width 0.2pt}c!{\vline width 0.2pt}c!{\vline width 0.2pt}c!{\vline width 0.2pt}c}
		\toprule
		\multirow{2}{*}{\textbf{Configuration}} & \multirow{2}{*}{\textbf{Device type}} & \multirow{2}{*}{\textbf{Condition}} & \textbf{Hip power}& \textbf{Knee power} & \textbf{Metabolic cost}\\
		&  &  &\textbf{consumption (W/kg)} & \textbf{consumption (W/kg)} & \textbf{reduction (\%)} \\
		\midrule[0.75pt]
		\multirow{4}{*}{\textbf{bi-articular}} & Ideal & noload & 4.65~$\pm$~0.75$^{*}$ & 5.10~$\pm$~0.81$^{**}$ & 22.38~$\pm$~4.91 \\
		\cmidrule[0.2pt]{2-6}
		& Ideal & loaded & 4.90~$\pm$~0.76$^{\dagger}$ & 5.62~$\pm$~1.55$^{\ddagger}$ & 20.49~$\pm$~2.87 \\
		\cmidrule[0.2pt]{2-6}
		& torque limited "Db"  & noload & 3.54~$\pm$~0.87$^{*}$ & 3.07~$\pm$~0.43$^{**}$ & 20.63~$\pm$~5.39 \\
		\cmidrule[0.2pt]{2-6}
		& torque limited "Db"  & loaded & 3.06~$\pm$~0.54$^{\dagger}$ & 3.26~$\pm$~0.68$^{\ddagger}$ & 16.99~$\pm$~3.44 \\
		\midrule[0.75pt]
		\multirow{4}{*}{\textbf{mono-articular}} & Ideal & noload & 2.92~$\pm$~0.44$^{\S}$ & 4.24~$\pm$~0.74 & 22.47~$\pm$~4.89 \\
		\cmidrule[0.2pt]{2-6}
		& Ideal & loaded & 4.08~$\pm$~0.69$^{\gamma}$ & 4.24~$\pm$~0.97$^{\gamma\gamma}$ & 20.45~$\pm$~2.81 \\
		\cmidrule[0.2pt]{2-6}
		& Torque limited "Ce"  & noload & 2.29~$\pm$~0.39$^{\S}$ & 3.97~$\pm$~0.83 & 20.67~$\pm$~5.01 \\
		\cmidrule[0.2pt]{2-6}
		& Torque limited "Be"  & loaded & 2.67 $\pm$ 0.45$^{\gamma}$& 3.19 $\pm$ 0.59$^{\gamma\gamma}$ & 15.65 $\pm$ 2.71 \\
		\bottomrule
		\multicolumn{6}{l}{\small{Note: Identical superscripts indicate a pairwise statistically significant difference (7 subjects, 3 trails, Tukey post-hoc, $p < 0.05$).}} \\
		\multicolumn{6}{l}{\small{$\pmb{*:}$ $F(1,12) = 5.863$, $P = 0.032$; $\pmb{\S:}$ $F(1,12) = 8.361$, $P = 0.014$; $\pmb{\ddagger:}$ $F(1,12) = 13.491$, $P=0.003$; $\pmb{\dagger:}$ $F(1,12) = 20.082 $, $P = 0.001 $.}} \\
		\multicolumn{6}{l}{\small{$\pmb{\gamma:}$ $F(1,12) = 22.015$, $P=0.001$; $\pmb{**:}$ $F(1,12) = 39.441$, $P < 0.001$; $\pmb{\gamma\gamma:}$ $F(1,12) = 5.944$, $P = 0.031$ }} \\
	\end{tabular}%
	\label{Table_Device_MaxPower_Comparison}
\end{adjustwidth} \normalsize
\end{table}

The mono-articular and bi-articular exoskeleton configurations show similar performance in assisting subjects under both \emph{noload} and \emph{loaded} walking conditions, for different torque limits at their hip and knee actuators. On the other hand,  the power consumption of the optimal exoskeleton configurations lying on the Pareto-front curve reveals that the mono-articular exoskeletons have a slightly superior power consumption performance under \textit{noload} walking condition, while this performance is significantly affected by the load carried; the power consumption of the mono-articular exoskeletons becomes practically same with the bi-articular exoskeletons when subjects are under \textit{loaded} walking condition.

\begin{figure*}[h!]
	\centering
	\includegraphics[width=\linewidth]{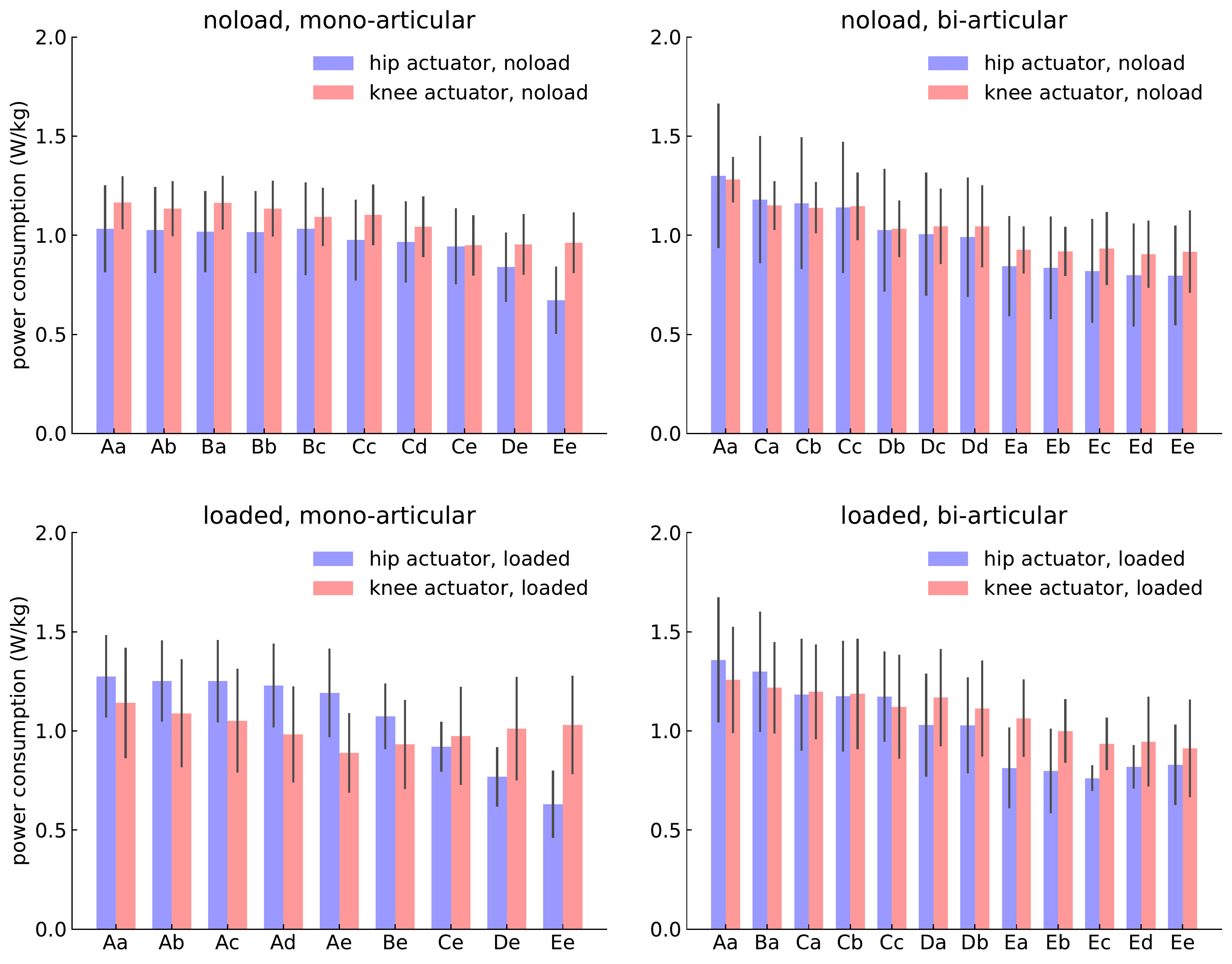}
	\vspace{-3mm}
	\caption{{\small\textbf{Power consumption of actuators of optimal exoskeletons lying on the Pareto-front curves}. The horizontal axes refer to the non-dominated exoskeletons on the Pareto-front curve. The vertical error bars represent the standard deviation of devices over 7 subjects and 3 trials per subject.}}
	\label{Fig_Paretofronts_Actuators_EnergyBarPlot}
\end{figure*}

The power distribution between the hip and knee actuators of the optimal mono-articular exoskeletons presented in Fig~\ref{Fig_Main_Paretofronts} indicates that
the knee actuators are more dominant compared to the hip actuators for all non-dominated device configurations under \emph{noload} walking condition, while loading significantly increases the mechanical work done by  the hip actuators such that hip actuators become dominant under \emph{loaded} walking condition, as presented in Fig~\ref{Fig_Paretofronts_Actuators_EnergyBarPlot}.

In contrast, the power distribution between the hip and knee actuators of the non-dominated bi-articular exoskeletons is not significantly affected by loading of the subjects, as the power consumptions of all actuators increase in a uniform manner. Additionally, the torque limitations at the hip and knee joints of the non-dominated bi-articular exoskeletons stay largely unchanged under both \emph{noload} and \emph{loaded} walking conditions, while the torque limitations of the non-dominated  mono-articular exoskeletons differ significantly based on the load. Insensitivity of non-dominated  bi-articular exoskeleton designs on loading may facilitate the design of assistive devices and the development of generic controllers that can assist subjects under different load conditions.

\subsubsection*{Torque and power profiles of non-dominated exoskeletons}

\begin{figure*}[t!]
	\centering
	\includegraphics[width=\linewidth]{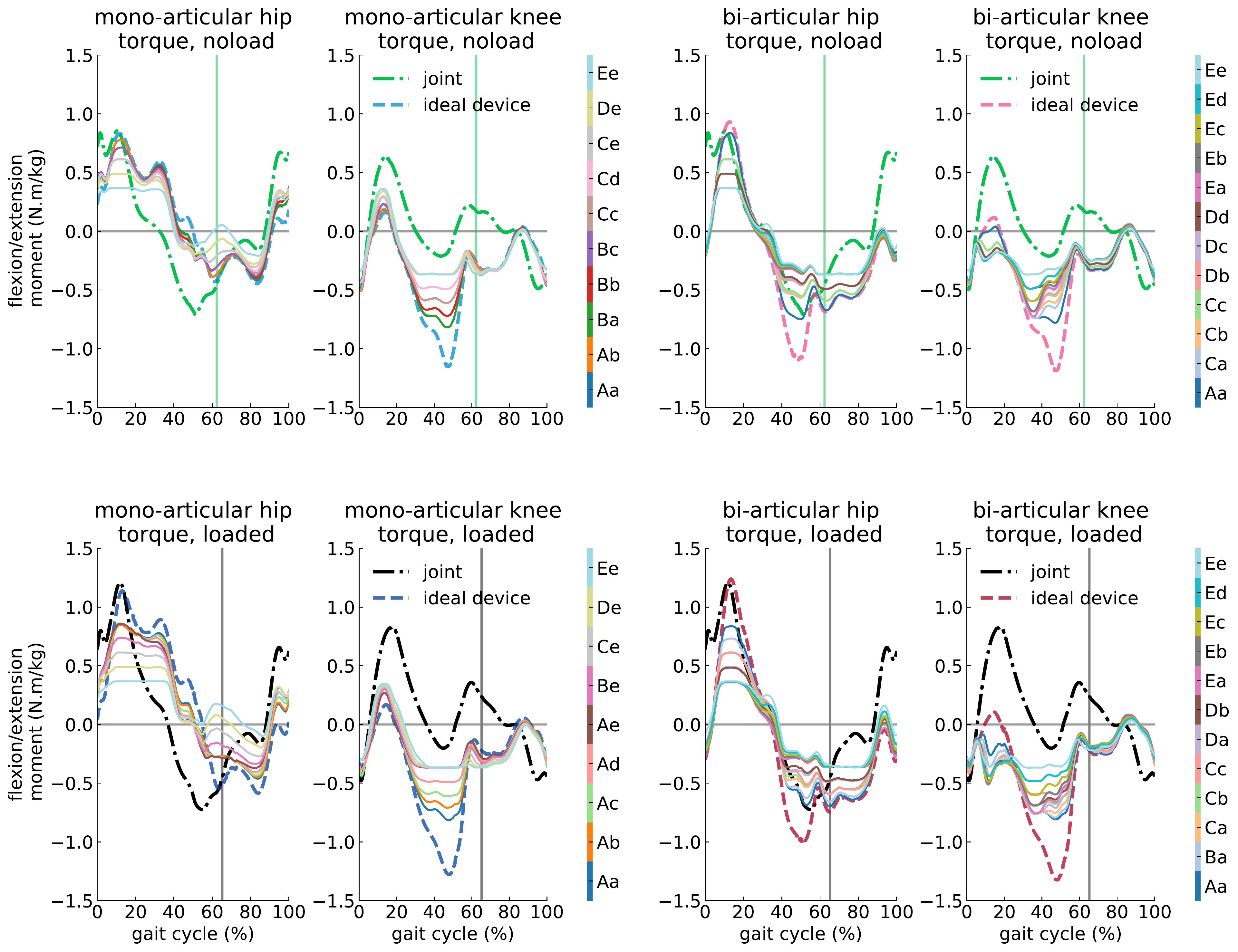}
	\vspace{-4mm}
	\caption{{\small\textbf{Torque profiles of the non-dominated exoskeletons together with the joint moments.} Each line represents the torque profile of a non-dominated exoskeleton as defined in the color bar. The data points represent the average over 7 subjects with 3 trials and are normalized by subject mass. The label on each marker denotes results from different peak torque constraints, as defined in Fig~\ref{Fig_Main_Paretofronts}. }}
	\label{Fig_Paretofronts_Torque_Profiles}
\end{figure*} \vspace{-2mm}

The torque profiles of non-dominated exoskeleton configurations under torque limits differ from the net joint moments of the hip and knee joints. The general torque trajectories of these actuators follow a similar trend with the torque profiles of ideal exoskeletons, as represented in Fig~\ref{Fig_Paretofronts_Torque_Profiles}. The ideal and torque-limited torque profiles of bi-articular hip actuators display a magnitude difference mostly during the mid-stance phase and the terminal-stance to terminal-swing phases. Quantitatively, the ideal and torque-limited profiles during the \emph{noload} walking condition have peak to peak differences of 45.96\%~(IQR,~26.74) in loading response, 46.87\%~(IQR,~26.51) in mid-stance, 60.25\%~(IQR,~14.73) in terminal-stance, and 45.36\%~(IQR,~26.63) during pre-swing phases, respectively. Similarly, the median of differences between the profiles during the \emph{loaded} walking condition are 59.18\%~(IQR,~20.28) in loading response, 58.16\%~(IQR,~20.89) in mid-stance, 46.16\%~(IQR,~24.63) in terminal-stance, and 43.75\%~(IQR,~27.40) during pre-swing phases, respectively.

In contrast to the hip actuator, the knee actuators follow a very similar profile with the ideal knee actuator during the swing phase, while the torque profiles of the knee actuators are highly different from those of the ideal device during the stance phase. For bi-articular exoskeletons, loading  affects only the magnitude of the torque-limited profiles, but assistive torques follow a similar trend with respect to the ideal torque profiles under both \emph{noload} and \emph{loaded} walking conditions. Quantitative comparisons of torque-limited and ideal assistance torque profiles of the knee actuators of the bi-articular exoskeletons show that the main differences occur with 45.87\%~(IQR,~13.16) and 45.49\%~(IQR,~10.72) during the terminal stance phase, 41.31\%~(IQR,~12.54) and  54.46\%~(IQR,~16.42) during the pre-swing phase,  in \emph{loaded} and \emph{noload} walking conditions, respectively.

Assistive torque profiles of the hip and knee actuators of the non-dominated torque-limited mono-articular exoskeletons demonstrate larger differences in comparison to the profiles of the ideal exoskeleton throughout the gait cycle. The hip actuators display a magnitude difference during the load response phase, and almost all of the torque-limited hip actuators saturate from the mid-stance to initial swing phases. The difference between the ideal and torque-limited hip actuators becomes significant during the pre-swing to mid-swing phases, during which the torque trajectories of the torque-limited hip actuators are not only different from the ideal actuator but also exhibit high variations among non-dominated solutions.

The quantitative comparisons of the torque-limited profiles of the hip actuator with those of ideal devices show that the main changes of the profiles during \emph{loaded} walking condition occur from loading response to mid-swing. In particular, the median of peak to peak differences between the torque-limited and ideal profiles of the hip actuator are 27.21\%~(IQR,~20.84) in loading response, 25.77\%~(IQR,~21.01) in mid-stance, 25.34\%~(IQR,~14.78) in terminal-stance, 29.07\%~(IQR,~21.37) during pre-swing, 28.07\%~(IQR,~24.24) during initial swing, and 26.80\%~(IQR,~27.18) during mid-swing phases, respectively. Similarly, the significant peak to peak differences between the hip actuator profiles during the \emph{noload} walking condition occur with 20.01\%~(IQR,~11.96) in loading response,  20.63\%~(IQR,~12.33) in mid-stance, and  27.25\%~(IQR,~17.76) during initial swing phases. It is noteworthy to mention that the high interquartile ranges confirm the high variation of the profiles for the mono-articular exoskeletons.

The torque profiles of the mono-articular knee actuators have a greater resemblance to the ideal knee actuator torque profiles, with the torque limited and ideal actuators having practically identical profiles during the swing phase. The main differences between the knee torque profiles take place with median peak to peak difference of 54.60\%~(IQR,~27.24) and 19.57\%~(IQR,~26.92) during the mid-stance, 70.49\%~(IQR,~19.67) and 46.15\%~(IQR,~28.29) during the terminal-stance phase, and 38.11\%~(IQR,~14.40) and 29.44\%~(IQR,~24.67) during the pre-swing phase, in \emph{loaded} and \emph{noload} walking conditions, respectively.

Some of the torque-limited bi-articular knee actuators demonstrate a change in the direction of assistance torques during the early stance phase in comparison to the ideal exoskeleton torque profiles. The reason for this direction change is that the muscle generated moments exceed the knee joint moment during the early stance phase, and the knee actuators oppose the muscle generated moment to follow the knee moment trajectory, as shown in Fig~C1 in~\nameref{S4_Appendix}, where the torque profiles of the bi-articular ''Aa'' and ''Ea'' exoskeleton configurations are presented under {\it noload} walking condition. The reduction of the torque capacity of the hip actuator due to saturation decrease the activity of the rectus femoris and soleus muscles, and increased the activity of gastrocnemius, iliopsoas, and vasti muscles, respectively, as shown in Fig~C2 in~\nameref{S4_Appendix}. The increase in the activity of the vasti muscles serving as the knee extensor muscles, along with the excessive activity of rectus femoris, cause a higher extension moment on the knee joint during the early stance phase, and this moment is compensated by the assistive knee actuator by changing the direction of its torque trajectory.

Similar to the bi-articular knee actuator, the hip actuators of the mono-articular exoskeletons display large variance during the pre-swing and initial swing phases. The reason for this significant variance is rooted in the muscular activities of assisted subjects and the torque capacity of mono-articular hip actuators. Conducting a comparison between mono-articular ''Ae'' and ''Ee'' exoskeletons under \emph{loaded} walking condition shows that by reducing the torque generation capacity of the hip actuator, the activity of rectus femoris is reduced and consequently, the activity of iliopsoas muscles increase during the pre-swing and the initial swing phases, as shown in Fig~C3 in~\nameref{S4_Appendix}. This modified muscle coordination results in a muscle-generated moment profile exceeding the net joint moment profile during the pre-swing and the initial swing phases, which is counteracted by the hip actuators of the mono-articular exoskeletons, as can be observed in Fig~C4 in~\nameref{S4_Appendix}.

Unlike the non-dominated bi-articular exoskeleton configurations, in which the load condition only affects the magnitude of the assistance torque profiles and there exists a close similarity between the torque trajectories under \emph{noload} and \emph{loaded} walking conditions, the torque profiles of the hip actuator of the non-dominated mono-articular exoskeleton configurations differ considerably under \emph{noload} and \emph{loaded} walking conditions. In particular, the hip torque profiles of the mono-articular exoskeletons under different load conditions exhibit different trajectories and magnitudes during all phases of the gait cycle, and the differences are quite hard to predict during the load response to mid-swing phases.

\begin{figure*}[h!]
	\centering
	\includegraphics[width=\linewidth]{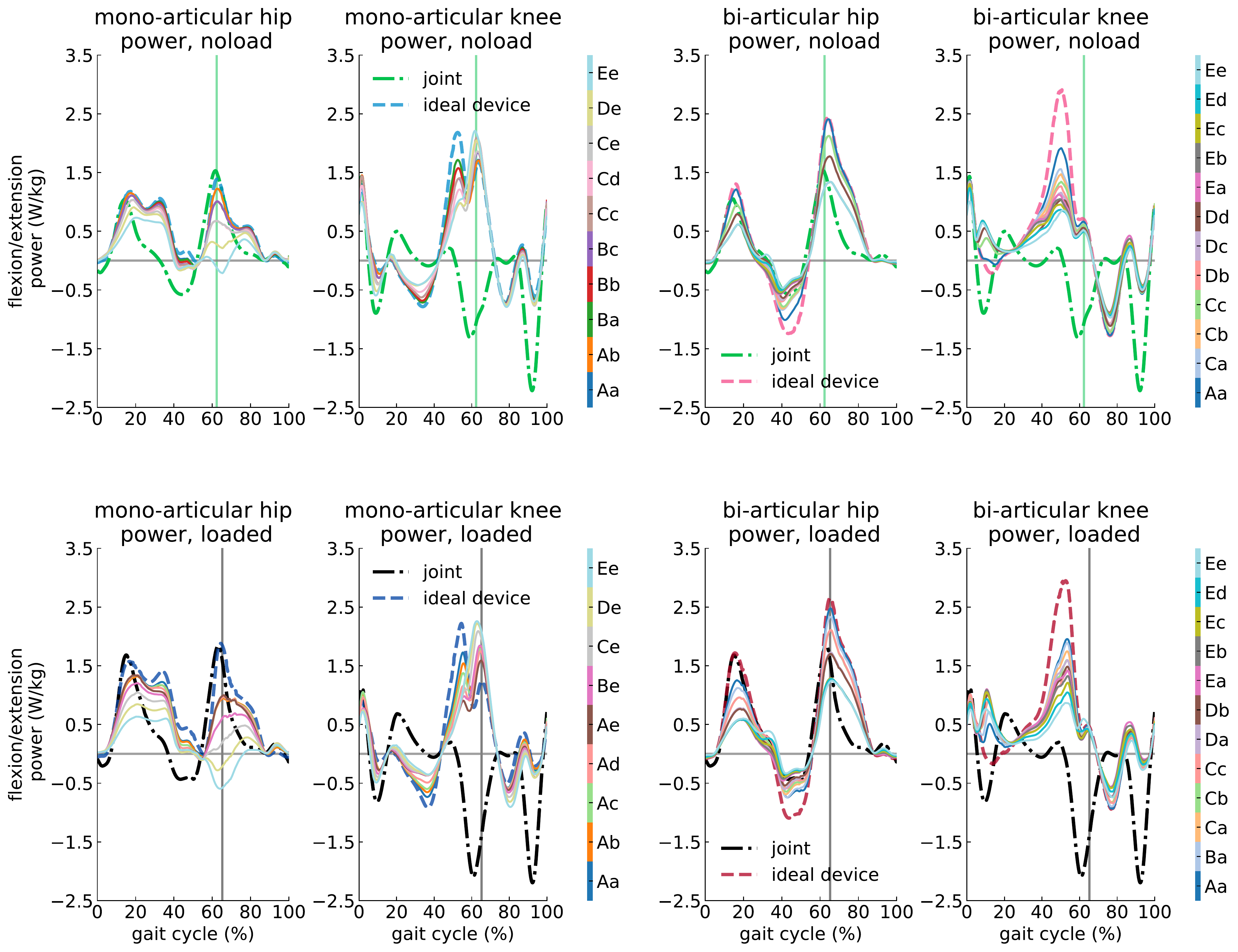}
	\vspace{-4mm}
		\caption{{\small\textbf{Power profiles of the non-dominated exoskeletons together with the joint power.} Each line represents the power profile of a non-dominated exoskeleton as defined in the color bar. The data points represent the average over 7 subjects with 3 trials and are normalized by subject mass. The label on each marker denotes results from different peak torque constraints, as defined in Fig~\ref{Fig_Main_Paretofronts}.} }
	\label{Fig_Paretofronts_Power_Profiles}
\end{figure*}

The power profiles of the non-dominated torque-limited bi-articular and mono-articular exoskeletons are presented in Fig~\ref{Fig_Paretofronts_Power_Profiles}. These power profiles resemble those of the ideal devices; however, the power consumption and maximum positive power are dramatically reduced for torque-limited exoskeletons, even under the highest torque limit.

The hip actuators of the torque-limited bi-articular exoskeletons demonstrate similar power trajectories under both loading conditions, with considerably lower power than ideal exoskeletons. The assistance torque magnitude difference becomes substantial with  median of peak to peak differences of 43.17\%~(IQR,~27.29) and 57.30\%~(IQR,~21.0) in loading response, 36.20\%~(IQR,~25.80) and 52.72\%~(IQR,~22.01) in mid-stance, and 29.17\%~(IQR,~32.56) and 35.42\%~(IQR,~31.19) in initial swing phases, during \emph{loaded} and \emph{noload} walking conditions, respectively.
	
Although the power profiles of the bi-articular knee actuators have a high correlation with the ideal actuators, quantitative comparison of their power profiles demonstrate peak to peak differences with a median of -9.84\%~(IQR,~5.60) and -57.74\%~(IQR,~24.29) during loading response, 58.26\%~(IQR,~13.05) and 49.874\%~(IQR,~11.23) during terminal stance, and 54.98\%~(IQR,~17.38) and 40.66\%~(IQR,~12.05) during initial swing phases, in \emph{loaded} and \emph{noload} walking conditions, respectively.

The power profiles of the non-dominated torque-limited mono-articular hip actuators possess a high variation among different solutions, with most of the non-dominated solutions displaying dissimilar power profiles. Only the non-dominated torque-limited mono-articular hip actuators with the highest peak torque limitations show a close resemblance to the knee torques provided by the ideal exoskeleton. In contrast to the mono-articular hip actuators, the power trajectories of the mono-articular  knee actuators under both loading conditions are similar to those of the ideal exoskeleton, with slight differences due to actuator saturation during the pre-swing to mid-swing phase, when the peak power consumption takes place during the toe-off.

The torque and power profiles of non-dominated mono-articular and bi-articular exoskeletons reveal that, although the non-dominated mono-articular exoskeleton configurations have a lower power consumption compared to non-dominated bi-articular exoskeleton configurations, the variation of the torque and power profiles within the optimal configurations of the mono-articular exoskeletons is significantly higher than that of bi-articular devices. Furthermore, the torque and power profiles of the non-dominated bi-articular exoskeleton configurations are significantly less sensitive to loading conditions. High variations among the non-dominated mono-articular exoskeletons indicate that achieving general design and control policies to assist subjects under different load conditions may be more challenging for mono-articular exoskeleton configurations, as different assistance torque strategies,  actuator and power modules need to be considered for varying load conditions.

We present quantitative comparisons among several non-dominated solutions for mono-articular and bi-articular exoskeletons under both load conditions, in terms of their torque and power profiles at each phase of the gait cycle, together with a comprehensive discussion through~\cite{Alibonab2021_thesis}. These comparisons indicate that even though the mono-articular and bi-articular exoskeleton configurations can achieve the same level of performance on the objectives space, they will have significantly different power and torque profiles for delivering assistance, and these differences affect the muscle coordination of the assisted subjects, causing dominant muscles, such as rectus femoris and psoas, demonstrate different activation patterns. These comparisons also show that the bi-articular exoskeleton configurations have more predictable power and torque profiles under both loading conditions than the mono-articular exoskeleton configurations.

\bigskip
\subsubsection*{Effect of non-dominated exoskeletons on the joint reaction forces/moments}

Actuator saturations do not seem to considerably affect the majority of the joint reaction force/moment profiles, and  these profiles of the subjects assisted by the torque-limited non-dominated exoskeletons mostly resemble the joint reaction force/moment profiles of subjects assisted by the ideal exoskeletons. For instance, the reaction moments/forces at the ankle joint closely follow the profiles of the ideal devices, indicating that the non-dominated  mono-articular and bi-articular exoskeletons have practically the same effect on the muscles contributing to the reaction loads/moments at the ankle joint, as presented in Figs~B6 and~B7 in~\nameref{S3_Appendix}.  Similar to the ankle joint, the reaction forces at the hip joint also closely resemble the trajectories of the subjects assisted by the ideal exoskeletons. Nevertheless, there exists a magnitude difference between the ideal and torque-limited joint reaction force/moment trajectories, especially during the stance phase of a gait cycle, as shown in Fig~B8 with  in~\nameref{S3_Appendix}.

On the other hand, torque-limited exoskeletons have a different effect on the reaction forces/moments at the knee, and patellofemoral joints compared to the ideal exoskeletons. In particular, the reaction forces of the knee and patellofemoral joints increase during the loading response and mid-stance phases in compressive ($F_y$) and medial-lateral ($F_z$) directions for both \emph{loaded} and \emph{noload} walking conditions. Nonetheless,  torque-limited non-dominated bi-articular exoskeletons  demonstrate a better performance than the ideal bi-articular exoskeletons in the late stance phase and do not increase the reaction forces as done by the ideal bi-articular exoskeletons (see Fig~B9--B12 in~\nameref{S3_Appendix}). Along these lines, it is reasonable to deduce that this different behavior is due to the changes in the activations of the rectus femoris, iliopsoas, and hamstring muscles, as discussed in the case studies in~\cite{Alibonab2021_thesis} for various non-dominated bi-articular exoskeleton configurations.

The torque-limited non-dominated mono-articular exoskeletons display better performance on reducing the reaction forces/moments at the knee and patellofemoral joints. The reaction forces/moments of  mono-articular exoskeletons closely resemble those of the ideal mono-articular exoskeleton during the loading response, and the torque-limited mono-articular exoskeletons are able to reduce the peak forces/moments during the late stance phase better than the ideal mono-articular exoskeletons (see Figs~B8--B12 in~\nameref{S3_Appendix}). The torque-limited mono-articular exoskeletons display larger variations among non-dominated  solutions, in terms of the reaction force/moment trajectories, as is the case for the assistance torque profiles of these exoskeletons.

\bigskip
\subsection*{Inclusion of regeneration effects on the Pareto solutions}

\begin{figure*}[t!]
	\centering
	\includegraphics[width=\linewidth]{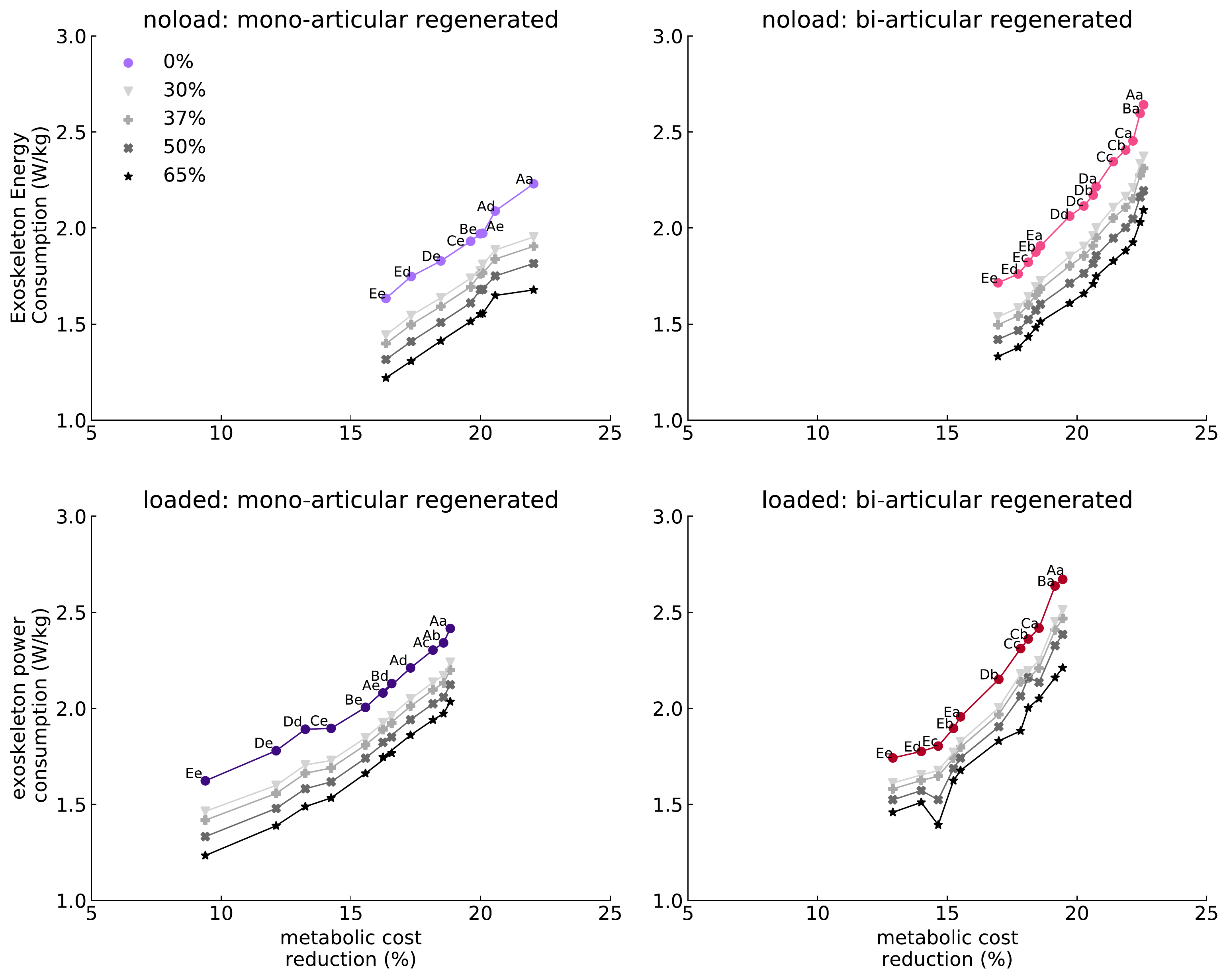}
	\vspace{-3mm}
	\caption{{\small\textbf{Inclusion of regeneration effect with different efficiencies on the Pareto solutions.} The label on each marker denotes different peak torque constraints, as defined in Fig~\ref{Fig_Main_Paretofronts}. The different markers represent different power regeneration efficiencies, ranging from 0 to 65\%. }} 
	\label{Fig_Paretofronts_Regeneration_Efficiency_Comparison}
\end{figure*}

\begin{figure*}[t!]
	\centering
	\includegraphics[width=\linewidth]{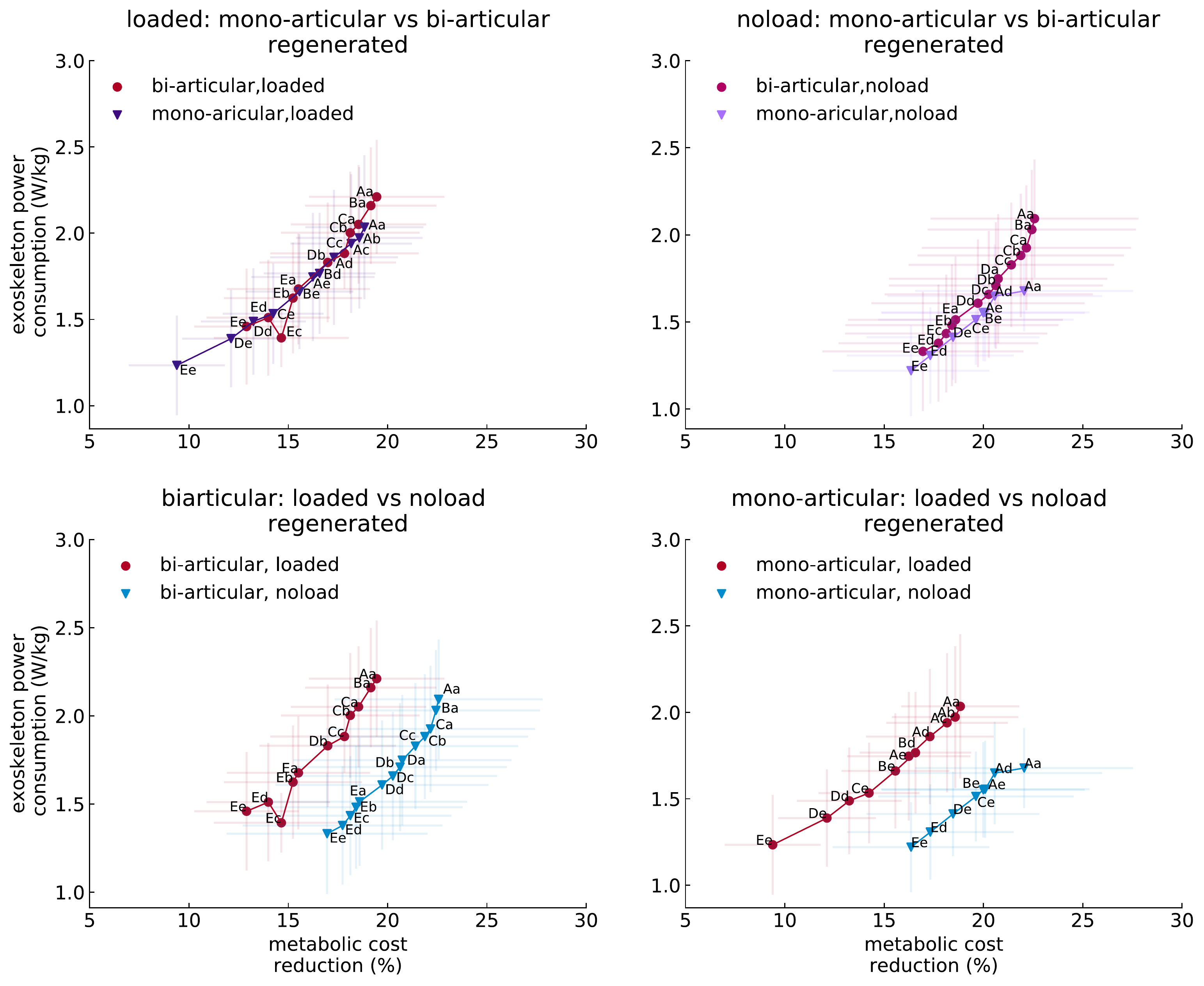}
	\vspace{-4mm}
	\caption{{\small\textbf{Pareto fronts under regeneration with 65\% efficiency.} The label on each marker is denoted to results from different peak torque constraints, as defined in Fig~\ref{Fig_Main_Paretofronts}. The data points on the Pareto-front curves are computed by averaging over 7 subjects and 3 trials.}}
	\label{Fig_Regenerated_Main_Paretofronts}
\end{figure*}

High power requirements of assistive exoskeletons and the finite energy density of the battery packs significantly limit the battery life of untethered mobile exoskeletons~\cite{Laschowski2019}. Young and Ferris~\cite{Young2017} have reported that the maximum battery life of portable exoskeletons is limited to 5~hours, indicating the need for several charging breaks within a day. Harvesting the dissipated power from assistive exoskeletons can help increase power efficiency of mobile exoskeletons, significantly prolonging their device battery life~\cite{Laschowski2019,Riemer2011}. Power regeneration and harvesting have been utilized in the literature for several assistive devices and robotic systems with a reported highest regeneration/power harvesting efficiency of 65\%, as reviewed in~\cite{Laschowski2019,Riemer2011}.

Fig~\ref{Fig_Paretofronts_Regeneration_Efficiency_Comparison}  presents the effect of regeneration on  Pareto-front curves for the  mono-articular and bi-articular exoskeleton configurations under \emph{noload} and \emph{loaded} walking conditions. Energy regeneration shifts all the Pareto-front curves downward along the y-axis, by improving the power consumption of all exoskeletons. As expected, regeneration efficiency has a high impact on the power consumption of the exoskeletons. However, analyzing the performance of both devices throughout the reported efficiency range demonstrates that the power consumption of exoskeletons can be significantly improved, even with a low regeneration efficiency. Quantitative analysis also shows that the inclusion of regeneration can improve the performance of the torque-limited bi-articular exoskeletons with a median from 10.10\%~(IQR,~0.28) to 21.88\%~(IQR,~0.61) depending on the regeneration efficiency during the \emph{noload} walking condition. Similarly, the performance of mono-articular exoskeletons can be improved from 10.19\%~(IQR,~0.94) to 22.08\%~(IQR,~2.04) based on the regeneration efficiency. Furthermore, regeneration of the dissipated energy can improve the performance from 6.90\%~(IQR,~0.29) to 14.49\%~(IQR,~0.62) for bi-articular exoskeletons, and from 7.93\%~(IQR,~0.1.96) to 17.19\%~(IQR,~4.24) for the mono-articular exoskeletons during \emph{loaded} walking condition, respectively.

Fig~\ref{Fig_Regenerated_Main_Paretofronts} presents the results for energy regeneration with 65\% efficiency. As can be observed from this figure, regeneration positively affects the solutions, enables some new non-dominated solutions to find place on the Pareto-front curve for both mono-articular and bi-articular exoskeletons.

While the design of assistive exoskeletons with regeneration capabilities may be challenging, the results shown in Fig~\ref{Fig_Paretofronts_Regeneration_Efficiency_Comparison} indicate that energy efficiency can be significantly improved, even with a relatively weak performance of a regeneration mechanism. An increase in power efficiency may enable use of smaller battery units, reducing the weight of the exoskeleton  and further improving the metabolic power consumption as discussed in the next subsection.

Comparisons between the exoskeleton configurations and the load conditions indicate that the knee actuators of the torque-limited mono-articular exoskeletons possess large regeneration potential, indicating an inefficient power profile of the knee actuator of the mono-articular exoskeletons. Unlike the mono-articular exoskeletons for which the regenerable power of the knee actuators are significantly higher than that of the hip actuators, both actuators of the bi-articular exoskeleton display similar regeneration potential.

The regenerable power of bi-articular exoskeletons significantly decrease by loading of the subjects. Although the low regenerable power potential reduces the positive effect of the regeneration on the device power consumption, this effect indicates that the power profiles of the bi-articular exoskeletons are more efficient in delivering positive power to the musculoskeletal system under \emph{loaded} walking condition, compared to the mono-articular exoskeletons. The inclusion of regeneration also changes the slope of the Pareto-front curves of bi-articular exoskeletons and enables bi-articular exoskeletons with high torque saturation limits to become  non-dominated solutions that lie on Pareto-front curves. On the other hand, the performance of the mono-articular exoskeletons are not significantly affected around high peak torque regions, as shown in Fig~\ref{Fig_Regenerated_Main_Paretofronts}.

\subsection*{Inclusion of inertial effects on the Pareto solutions}

One of the main challenges in the design of mobile exoskeletons is keeping their mass and inertia low, as inertial effects detrimentally impact the metabolic rate of assisted subjects. The effect of inertial properties on the metabolic cost of walking has been studied in the literature, and strong evidence has been provided to show that the metabolic power consumption increases considerably by added mass and inertia~\cite{Browning2007,Royer2005,Soule1969}.

The exoskeletons considered in this study have significantly different inertial properties from each other, due to their varying kinematic designs and peak torque capabilities. These differences result in significant inertial effects on the metabolic power consumption of subjects that need to be captured. The current control algorithm of OpenSim is unable to simulate any inertial variations in the musculoskeletal model that has not been captured during experimental data collection~\cite{Hicks2015} and data collection with all possible inertial distributions is not feasible. Along these lines, in this study, the detrimental effects due to exoskeleton inertial properties are captured by using the model proposed by Browning et al.\cite{Browning2007} and these effects are superposed on the metabolic cost of subjects to update the Pareto solutions.

Browning~\etal~\cite{Browning2007} propose a linear model for the effect of adding mass/inertia on each segment of the lower limb, by experimentally capturing the metabolic power expenditure of the subjects with the added mass/inertia. In particular, subjects walked at 1.25~m/s without carrying any load on their torso, which closely resembles to protocol used to capture motion data from the subjects under \textit{noload} walking condition, as detailed in~\cite{Dembia2017}. A good agreement between the experimental protocols of~\cite{Browning2007} and~\cite{Dembia2017} enables the utilization of the metabolic model developed in~\cite{Browning2007} to study the effect of mass/inertia addition of assistive exoskeletons, by superposing these effects on the simulation based metabolic rate of subjects under \emph{noload} walking condition.

\begin{figure*}[ht]
	\centering
	\includegraphics[width=\linewidth]{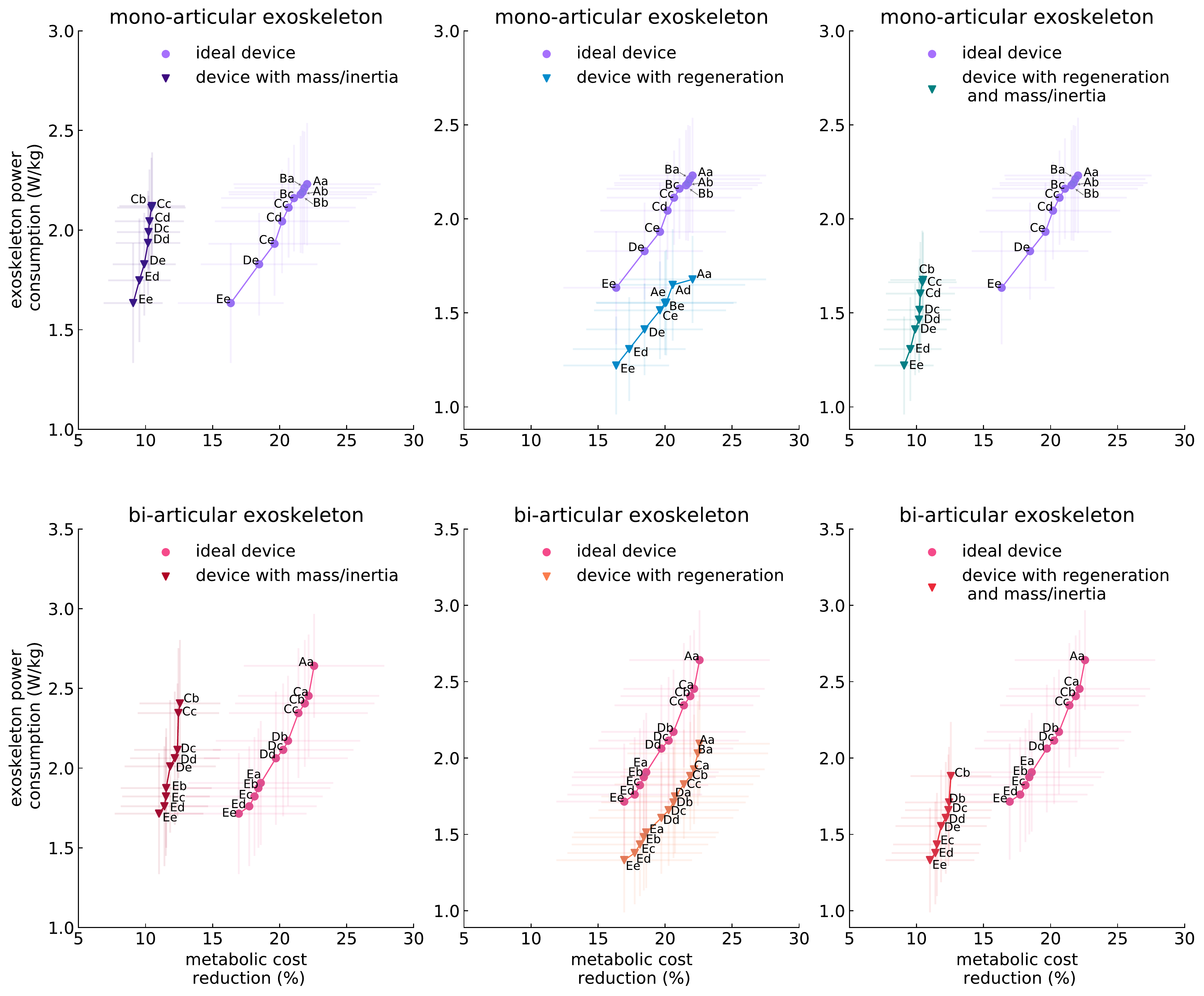}
	\vspace{-3mm}
	\caption{{\small\textbf{Inclusion of inertial and/or regeneration effects on the Pareto solutions.} The label on each marker denotes different peak torque constraints, as defined in Fig~\ref{Fig_Main_Paretofronts}. The data points on the Pareto-front curves are computed by averaging over 7 subjects and 3 trials, under \emph{noload} waking condition.}}
	\label{Fig_Paretofronts_Mass_Regeneration_Effect_Comparison}
\end{figure*}


The columns of Fig~\ref{Fig_Paretofronts_Mass_Regeneration_Effect_Comparison} present the Pareto-front curves for mono-articular and bi-articular exoskeletons, with no inertial or regeneration effects, with no inertial and  regeneration of 65\% efficiency, and with inertial effects and regeneration of 65\% efficiency, respectively. It can be observed from Fig~\ref{Fig_Paretofronts_Mass_Regeneration_Effect_Comparison} that the effect of inertial properties on the both exoskeleton configurations are significant. When inertial effects are considered, the slope of Pareto-front curves becomes very steep, indicating that exoskeleton power consumption becomes highly sensitive to changes in the peak torques capabilities and  the metabolic cost reduction achieved by using high peak torques becomes hard to justify as this gain comes with a very high  cost of increased power consumption of the exoskeleton.

\begin{figure*}[ht]
	\centering
	\includegraphics[width=\linewidth]{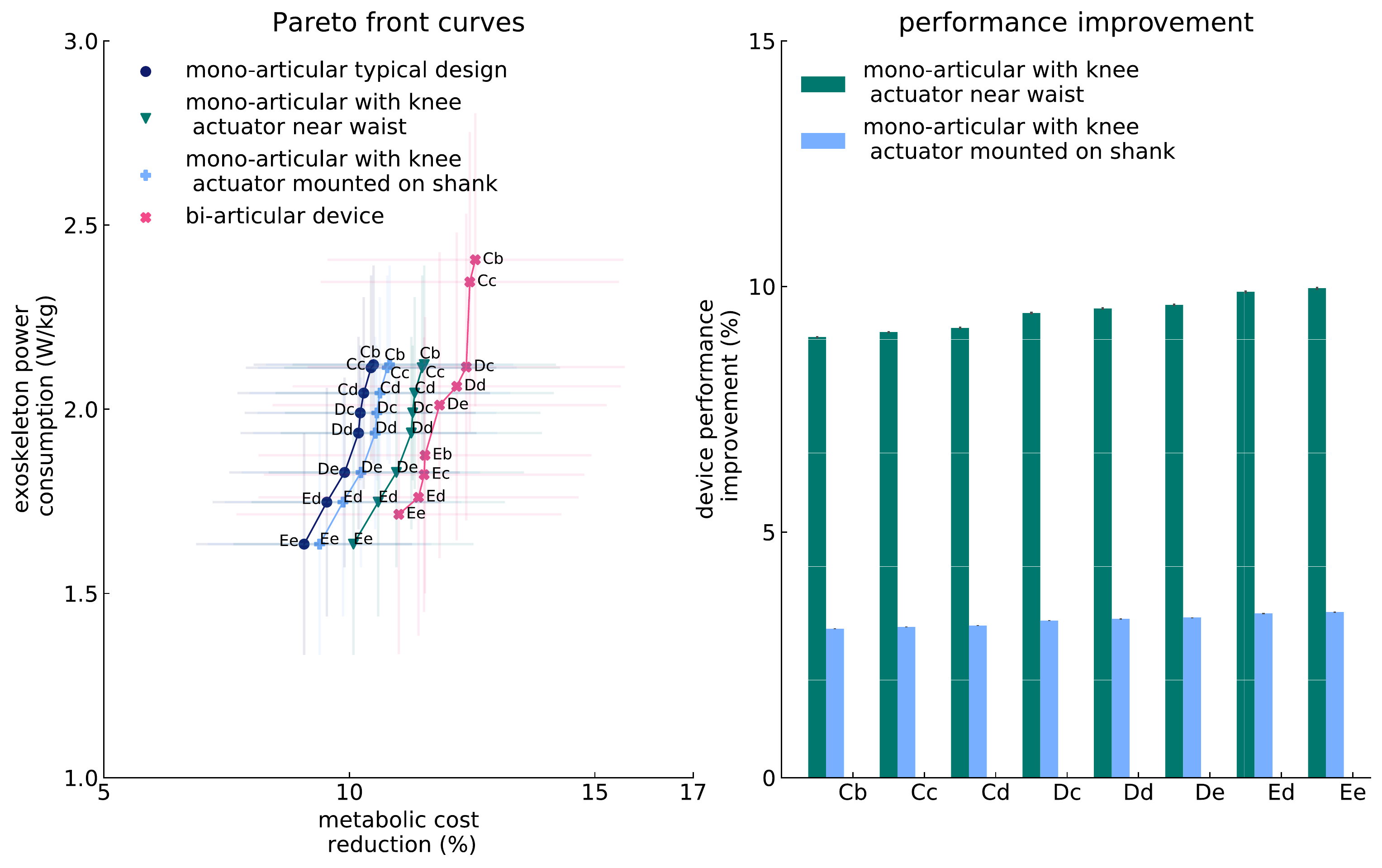}
	\vspace{-3mm}
	\caption{{\small\textbf{Comparison of Pareto-fronts of the mono-articular exoskeletons with various knee actuator placements to bi-articular exoskeletons.} Comparison of Pareto-front curves of various exoskeletons: a mono-articular exoskeleton with knee actuation unit placed at the knee, a mono-articular exoskeleton with the knee actuation unit placed on the upper-leg, a mono-articular exoskeleton with the knee actuation unit placed on the shank and a bi-articular exoskeleton. The data points on the Pareto-front curves are computed by averaging over 7 subjects and 3 trials, under \emph{noload} waking condition.}}
	\label{Fig_monoarticularExoskeleton_DifferentConfigs}
\end{figure*}

Since the leg inertia can be kept significantly lower for the bi-articular exoskeleton configurations compared to the mono-articular exoskeletons, bi-articular exoskeletons display a better performance when the inertial properties are considered. In particular, the performance of the mono-articular exoskeletons is degraded by median of 50.72\%~(IQR,~4.14) for non-dominated devices, while the performance of the bi-articular ones is degraded by median of 38.67\%~(IQR,~4.88) for non-dominated devices that lie on the Pareto front curve. For both the mono-articular and the bi-articular exoskeleton configurations, actuators with 70~Nm peak torque capabilities are not within the set of the non-dominated solutions, as high reflected inertia of these actuators detrimentally affect their performance, resulting in them becoming dominated solutions. The non-dominated solutions of both mono-articular and bi-articular exoskeletons utilize highest peak torques of 60~Nm at the knee, and 50~Nm at the hip joint, respectively.

The results indicate the importance of utilization of actuators with high-torque density to reduce power consumption of subjects by keeping the reflected inertia of actuation modules low. Furthermore, such actuators with low transmission ratios are likely to have better actuation and re-generation efficiency, positively affecting the exoskeleton power consumption.

\begin{figure*}[ht]
	\centering
	\includegraphics[width=\linewidth]{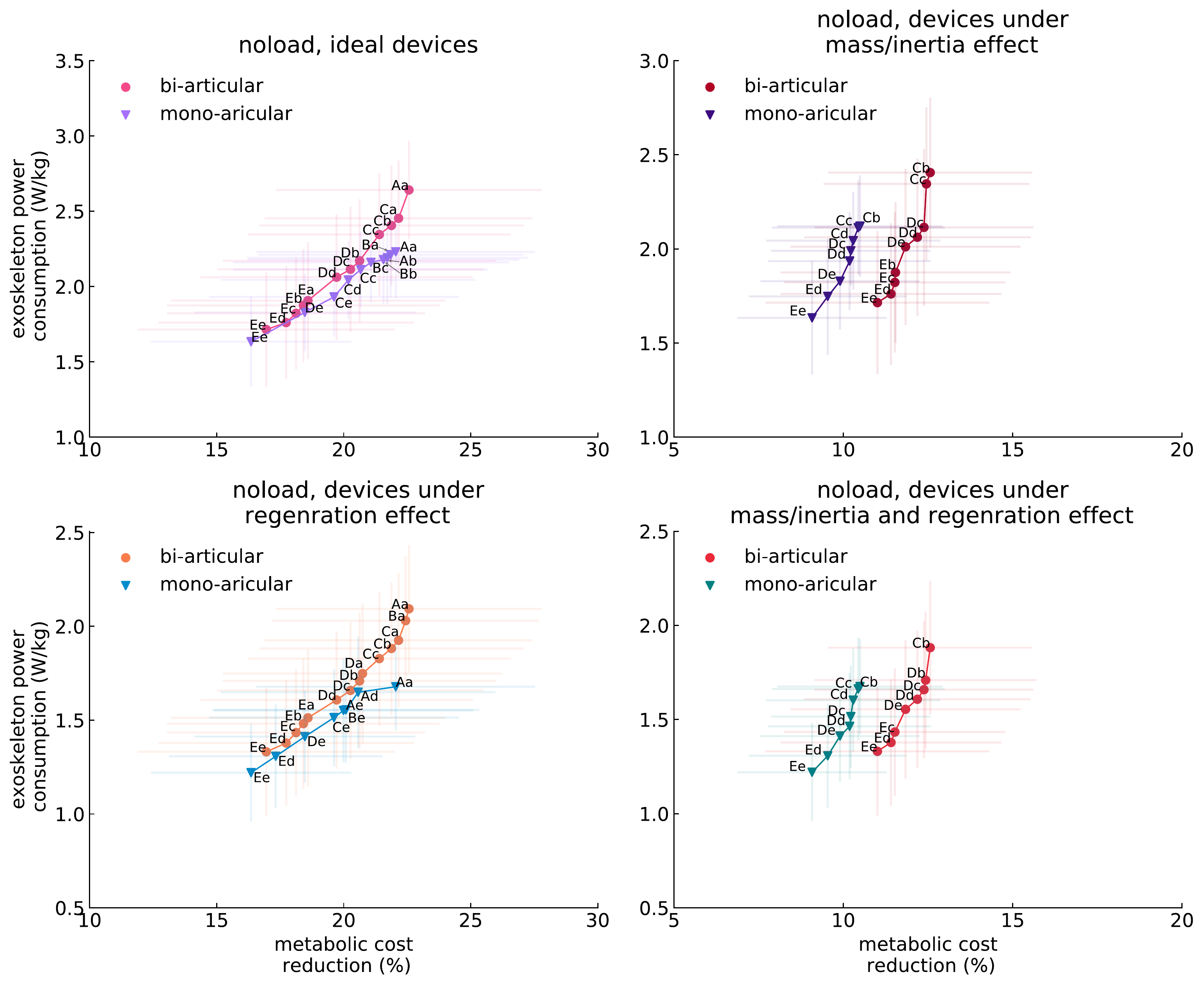}
	\vspace{-3mm}
	\caption{{\small\textbf{Comparison of non-dominated solutions.} The mono-articular and bi-articular exoskeleton configurations are compared under (a) ideal conditions, (b) considering the mass/inertia effects on metabolic power consumption, (c) considering power regeneration effects on power consumption, and (d) under both mass/inertia and regeneration effects. The data points on the Pareto-front curves are computed by averaging over 7 subjects and 3 trials, under \emph{noload} waking condition.}}
	\label{Fig_Paretofronts_Mass_Regeneration_Effect}
\end{figure*}

The results also strongly motivate placing the actuator units near the proximal joints and assisting the joints of interest distally through power transmission mechanism, as implemented by a parallelogram for the knee joint of bi-articular exoskeleton configuration considered in this study. Note that evolution has already implemented such an actuation arrangement in human musculoskeletal system through bi-articular muscles that enables to keep large portion of the muscle mass near the trunk and transfer power to the distal joints through ligaments, to reduce the reflected inertia of human leg. 




As discussed above, bi-articular exoskeleton configurations are more advantageous in terms of their inertial distribution; however, bi-articular actuation may be challenging to implement. As an alternative, mono-articular exoskeletons with improved inertial properties can be implemented by placing the knee actuator units not at the knee joint, but more proximal to the hip. Fig~\ref{Fig_monoarticularExoskeleton_DifferentConfigs} presents Pareto-front curves for two different placements of the knee actuation units, in addition to typical implementations of mono-articular and bi-articular exoskeletons. In particular, in one of the test configurations, the knee actuation unit is placed on the upper part of the thigh, while in the other configuration, the knee actuation unit is placed on the upper part of the shank.

As can be observed from Fig~\ref{Fig_monoarticularExoskeleton_DifferentConfigs}, both alternative placements of the knee actuator unit improve the performance of the mono-articular exoskeletons under inertial effects, but these configurations are still dominated by the bi-articular exoskeleton configuration. Among the alternative knee actuator unit placements, locating the actuator on the thigh proximal to the hip performs better by improving the performance of the mono-articular devices by a median of 9.50\%~(IQR,~0.47), than locating the actuator on the shank which can improve the performance of the mono-articular devices only by a median of 3.21\%~(IQR,~0.15). Yet, locating the actuator unit on the upper shank results in superior performance than placing it at the knee joint, as metabolic cost of walking is less sensitive to inertia addition to the shank than inertia addition to thigh~\cite{Browning2007}.

\subsection*{Selection of optimal exoskeletons among the non-dominated solutions}

Once the Pareto-front curves are computed for each exoskeleton configuration under different loading conditions and inertial and regeneration affects are superposed as depicted in Fig~\ref{Fig_Paretofronts_Mass_Regeneration_Effect}, different exoskeletons can be rigorously compared to each other and an optimal solution may be selected among all non-dominated solutions.

A comparison of non-dominated  mono-articular and bi-articular exoskeleton configurations indicate that due to their favorable inertial properties, bi-articular exoskeleton configurations dominate mono-articular exoskeleton configuration in terms of metabolic cost reduction. In particular, the best performing mono-articular exoskeleton (Cb)also has the highest power consumption (2.12~$\pm$~0.26~W/kg) and can achieve a metabolic cost reduction of 11.52~$\pm$~2.68~\%, which is practically identical with the metabolic cost reduction of the bi-articular exoskeleton (Ee) which has the lowest energy consumption (1.71~$\pm$~0.37~W/kg). Along these lines, bi-articular exoskeletons are a better choice in terms of metabolic cost reduction and power consumption metrics.

\begin{figure*}[ht]
	\centering
	\includegraphics[width=\linewidth]{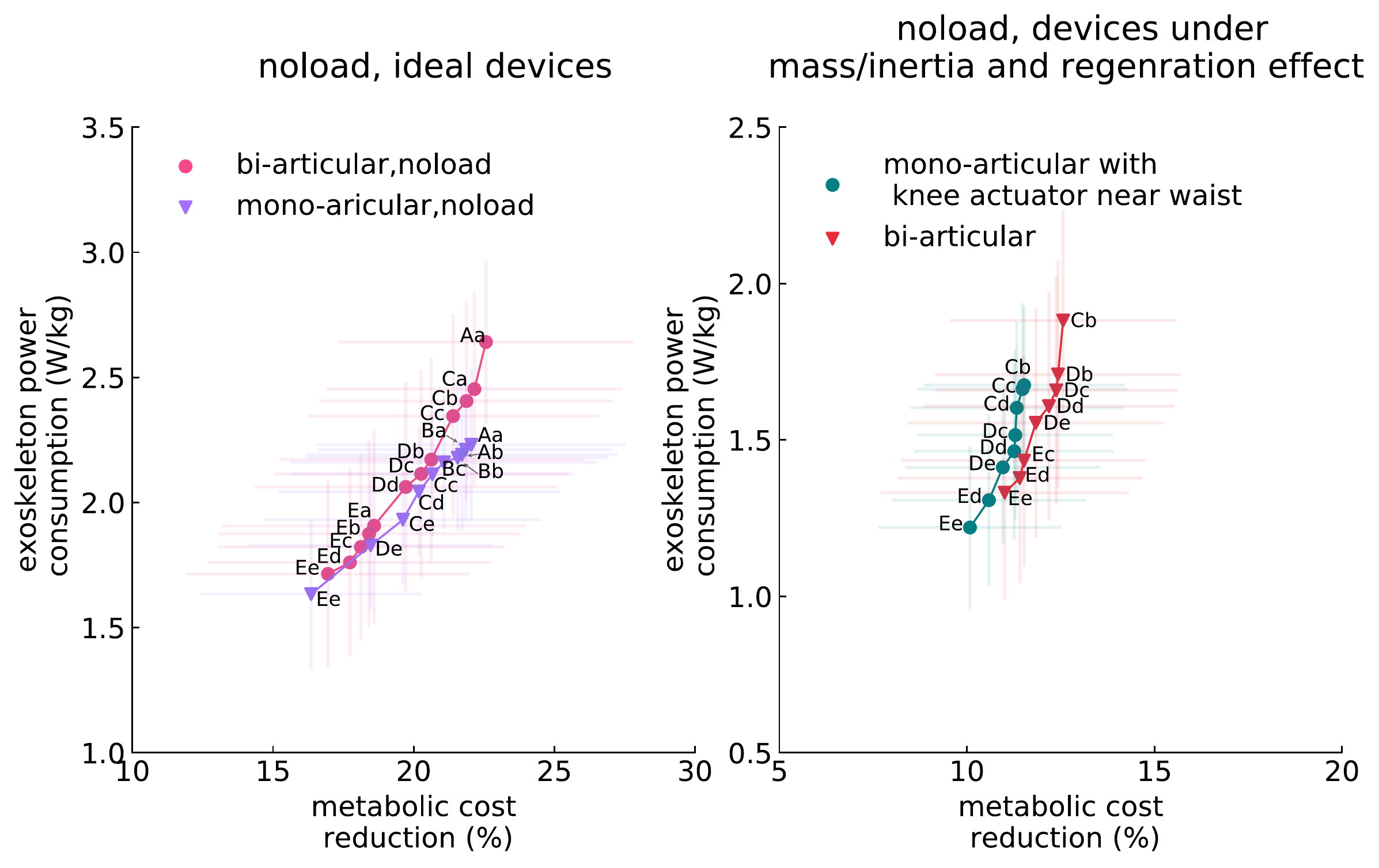}
	\caption{{\small\textbf{Comparison of Pareto-fronts of the mono-articular exoskeletons with knee actuator near waist to bi-articular exoskeletons.} Pareto-front curves of the mono-articular and bi-articular exoskeletons during \emph{noload} walking condition under ideal conditions and comparison of the Pareto-front curves of the mono-articular exoskeleton with knee actuator near the waist and bi-articular exoskeletons during \emph{noload} walking condition under the effect of mass/inertia and regeneration with 65\% efficiency.}} \vspace{-\baselineskip}
	\label{Fig_MonoNearWaistVsBi}
\end{figure*}

However, the implementation bi-articular exoskeletons may be more challenging than the implementation mono-articular exoskeletons, and inertial properties of mono-articular exoskeletons may be improved by more favorable placement of the knee actuator unit, as shown in Fig~\ref{Fig_monoarticularExoskeleton_DifferentConfigs}. Along these lines, Fig~\ref{Fig_MonoNearWaistVsBi} presents the Pareto-front curves of mono-articular exoskeletons for which the knee actuator unit is placed on the thigh near the hip and compare it with the Pareto-front curve of bi-articular exoskeletons during \emph{noload} walking condition.

The comparison of non-dominated solutions in Fig~\ref{Fig_MonoNearWaistVsBi} indicates that after the modified placement of the knee actuation unit, there exists a range of metabolic cost reductions that can be achieved by both kinematic arrangements. Note that the bi-articular exoskeleton configurations can still achieve high metabolic cost reductions that are not possible via mono-articular exoskeleton configurations at a cost of increased power consumption, while the mono-articular exoskeleton configurations can achieve lower power consumption a the cost of lower metabolic reduction.

All non-dominated solutions lying on the Pareto-front curves are optimal. At this point, a design selection can be made by the designer for the intended application, possibly considering additional primary requirements, such as changes in muscle activities of the key lower limb muscles and the reaction forces/moments at the joints under assistance, and the secondary requirements such as ease-of-implementation, robustness, and device cost.

The non-dominated mono-articular and bi-articular exoskeleton configurations with 11.524~$\pm$~2.684~\% and 11.52~$\pm$~3.26~\% metabolic cost reduction, respectively, are selected as sample designs. According to this selection, the mono-articular exoskeleton~Cb has actuators with 50~Nm peak torque at the hip and 60~Nm peak torque at the knee, while the bi-articular exoskeleton~Ec has actuators with 30~Nm peak torque at the hip and 50~Nm peak torque at the knee to achieve the same metabolic cost reduction performance. According to this selection, the mono-articular exoskeleton~Cb has a power consumption of~1.675~$\pm$~0.253~W/kg, while the bi-articular exoskeleton~Ec has 14.6\% lower power consumption of~1.43~$\pm$~0.337~W/kg, under 65\% regeneration effect.

Considering the mass/inertia effect on the efficiency of the devices by computing modified augmentation factor (MAF) shows that bi-articular Ec exoskeleton with 1.01~$\pm$~0.70~W/kg has 91.27\% better performance than mono-articular Cb exoskeleton with 0.5296~$\pm$~0.887~W/kg MAF value. Considering the regeneration effect, the mono-articular device can reach to 0.5636~$\pm$~0.78~W/kg MAF value, while the MAF value of the bi-articular device remains similar, indicating that dissipated power of Ec bi-articular exoskeleton is negligible; hence, it effectively delivers positive power to the subjects.

\begin{figure*}[ht]
	\centering
	\includegraphics[width=\linewidth]{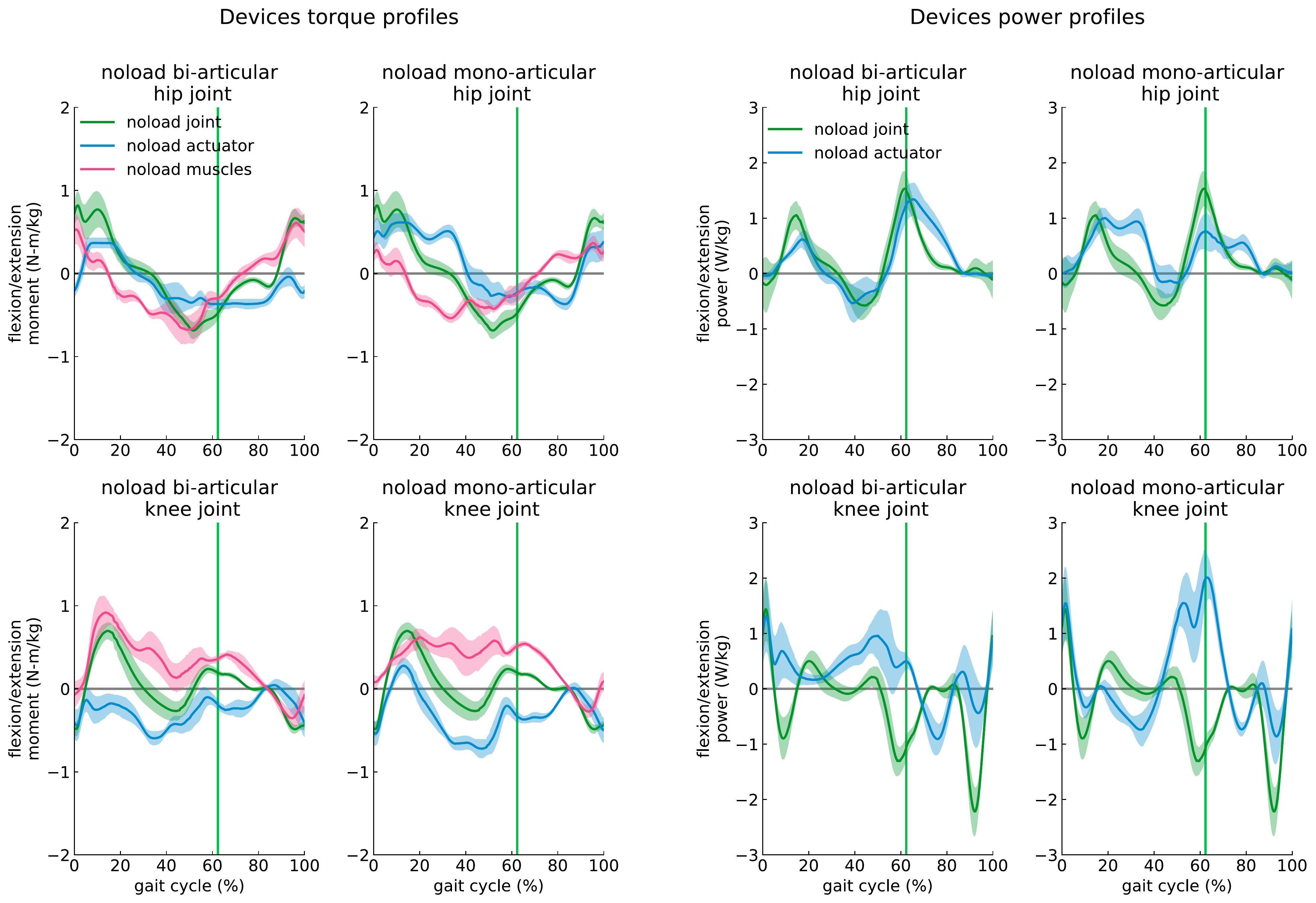}
	\caption{{\small\textbf{Torque and power profiles of mono-articular Cb and bi-articular Ec exoskeletons.} The torque and power profiles of assistive devices for subjects walking without an additional load (blue), and net joint power and torque profile for \textit{noload} (green) condition are shown for each actuator of the devices. The torque profile of moment generated by muscles (rose pink) is shown for each joint for both devices. The curves are averaged over 7 subjects with 3 trials and normalized by subject mass; shaded regions around the mean profile indicate standard deviation of the profile.}}
	\label{Fig_SpecificWeight_CbEc_Profiles} \vspace{-.5\baselineskip}
\end{figure*}
\begin{figure*}[ht]
	\centering
	\includegraphics[width=\linewidth]{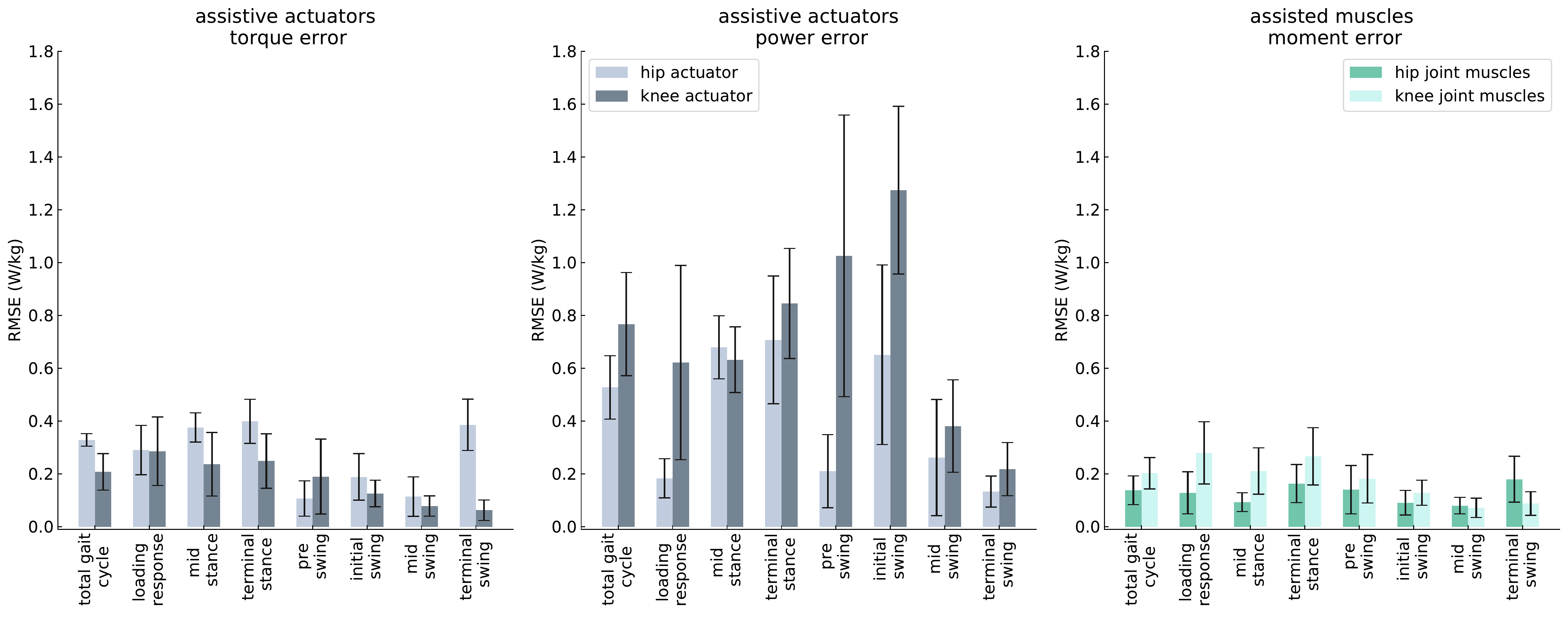}
	\caption{\small{\textbf{Torque, power, and muscles generated moment profiles of assistive devices. } The root mean square error between actuators of bi-articular and mono-articular devices and the muscles generated moment of subjects assisted by these devices. The RMSE was calculated during a total gait cycle (A), loading response (B), mid stance (C), terminal stance (D), pre swing (E), initial swing (F), mid swing (G), and terminal swing (H) phases.}}
	\label{Fig_SpecificWeight_CbEc_RMSE}
\end{figure*}

The difference between the torque profiles of the hip actuators of the selected devices is extensive as the profiles are significantly different from each other during most of the gait phases, as shown qualitatively in Fig~\ref{Fig_SpecificWeight_CbEc_Profiles} and quantitatively in Fig~\ref{Fig_SpecificWeight_CbEc_RMSE}. Although the torque profiles of the knee actuators have significant differences during the stance phase, especially during the loading response phase, these profiles become similar during the swing phase, which can be confirmed through computing the root mean square error between these two profiles shown in Fig~\ref{Fig_SpecificWeight_CbEc_RMSE}. The dissimilarity of the profiles becomes notable through power profiles. As can be quantitatively seen in Fig~\ref{Fig_SpecificWeight_CbEc_RMSE} both the hip and the knee actuators have different profiles during both stance and swing phases, with the exception of the late swing phase for the knee actuators, as shown in Fig~\ref{Fig_SpecificWeight_CbEc_Profiles}.
	
Further analysis of the bi-articular exoskeleton configuration~Ec under \emph{noload} and \emph{loaded} walking conditions, using the MAF metric shows that the performance of this bi-articular exoskeleton is 38.61\% better with a MAF = 1.40~$\pm$~0.80 W/kg under \emph{loaded} walking condition, in comparison to the \emph{noload} walking condition, which results in a MAF value of 1.01~$\pm$~0.70 W/kg. This improvement in MAF shows that the positive power under the \emph{loaded} walking condition is delivered to the subjects more effectively. Since the Cb mono-articular exoskeleton becomes a dominated solution under \emph{loaded} walking condition,  conducting a comparison would not lead to a fair comparison, as this device lose its optimality under different loading conditions.


To study the effect of the loading condition on a mono-articular exoskeleton which remains optimal in both walking conditions, we computed the MAF value for the mono-articular Ee exoskeleton under \emph{noload} and \emph{loaded} walking conditions as -0.20~$\pm$~0.43~W/kg and -0.15~$\pm$~0.56~W/kg, respectively. These values show that although the performance of the mono-articular exoskeletons, similar to those of the bi-articular devices, improve by loading subjects, the mono-articular devices in the low torque region are not able to improve the metabolic power consumption of subjects with respect to the metabolic expenditure of subjects in the no assistance condition. 
	
As shown in this subsection, obtaining a positive effect of the mono-articular devices on assisted subjects under mass/inertia effects is challenging. Studying the designs selected from the Pareto front curves confirms our claims about the overall performance of the mono-articular and the bi-articular exoskeletons and also shows that designing a mono-articular device requires a careful selection of actuators and their transmission ratio to compensate for the negative effect of the device inertia on the metabolic power consumption of subjects.

\subsection*{Study limitations}

Simulation-based studies of assistive exoskeletons have some limitations that need to be considered for proper interpretation of the results. One of the main limitations is the kinematics and ground reaction forces for the assisted subjects is assumed to stay unchanged from the experimentally captured data. Experimental studies have reported that exoskeletons may impose minor~\cite{Collins2015,Panizzolo2016} to significant~\cite{Quinlivan2017,Koller2015} changes on kinematics of assisted subjects. However, CMC algorithm OpenSim used for musculoskeletal simulations cannot capture such changes, and it is assumed  during simulations that unassisted and assisted subjects have the same kinematics, ground reaction force, and joint moment. It has been reported in the literature that metabolic cost may not  be substantially affected by kinematics changes~\cite{Vanderpool2008}.

There exists recent studies to address this limitation of musculoskeletal simulations. In particular, employment of dynamic optimization methods for performing musculoskeletal simulations seems promising, as these methods can capture the changes in the kinematics and dynamics of the assisted subjects. Yet, since altered kinematics can have several side effect, such as increasing joint loads, the kinematic adaptation may not always be desirable~\cite{Dembia2017}.

The assistive exoskeletons considered for musculoskeletal simulations are assumed to be massless, as changes in dynamical properties of subjects is not supported by OpenSim. To remedy this situation, the detrimental effects of mass and reflected inertia of exoskeletons are captured by a metabolic model~\cite{Browning2007} and superposed on the metabolic cost computations provided by the musculoskeletal simulations. Note that while this model is expected to closely approximate the metabolic cost of subjects walking with the exoskeleton, the effect of inertial properties of exoskeletons on the muscle activation profiles and on the power consumption of devices are not captured in this method.

Other limitations of this study are inherited from certain restrictions of the muscle model employed. One such limitation is due to extortionate passive force generated by the muscles~\cite{Hicks2015,Dembia2017}, which can result in extortionate muscular activities.  Extortionate muscular activities have been observed during comparison of musculoskeletal simulations with experimental muscular activities in a closely related work~\cite{Dembia2017}.  Another critical issue in Hill-type muscles model used for musculoskeletal simulations is that  muscle fatigue, an important factor in muscle recruitment strategies, is not taken into account. For instance, rectus femoris muscle, which is highly vulnerable to fatigue due to its fiber properties~\cite{Johnson1973}, experiences extreme activations in all of the assistance scenarios; this may cause subject to experience muscle fatigue on real life applications~\cite{Newham1983}. Tendon modeling, constant force enhancement, short-range muscle stiffness, and training effects are other limiting factors that are not captured by muscle models~\cite{Hicks2015}. The effect of all these factors need to be taken into account during the interpretation of the results presented in this musculoskeletal simulation based study.

In addition to these general limitations of musculoskeletal simulations, the dataset and musculoskeletal models used in this study, have been reported to include some inconsistencies with the experimentally collected data. For instance, musculoskeletal simulations predicted excessive passive force at the knee joint due to excessive activity of the knee extensor muscles. Another inconsistency is reported between the experimentally measured and simulation-based estimated metabolic cost of carrying while walking with a heavy load, during which the simulation-based metabolic cost computations underestimated the increase in metabolic cost of carrying a load. This underestimation indicates that the metabolic savings of the assisted subjects may also be underestimated by the musculoskeletal simulations. We refer readers to~\cite{Dembia2017} for a comprehensive discussion of the limitations of the dataset and musculoskeletal models,  to~\cite{Hicks2015} for a comprehensive discussion of all aspects of the OpenSim simulations, as well as recommendations for accurate interpretation of musculoskeletal simulations.

The physical attachment interface between the exoskeleton and the limb and misalignments between the  axes of actuation of exoskeletons and human joints are other important factors that may significantly affect performance of exoskeletons~\cite{Cenciarini2011}. These interface and alignment aspects are not  captured in this study.

Finally, Pareto optimizations in this study are conducted based on a large number of musculoskeletal simulations. Along these lines, the torque saturation limits for the exoskeletons are discretized to span the range from 70~Nm to 30~Nm with a step size of 10~Nm, to keep the runtime of optimizations in a feasible range. This discretization effect needs to be considered during the interpretation of the Pareto-front curves.

In general, it is not reasonable to expect a close quantitative match between the musculoskeletal simulation results and data collected from human-in-the-loop evaluation of assistive exoskeletons, due to a large number of inherent limitations acknowledged above. However, the musculoskeletal simulations  still hold high promise in providing effective design guidelines, as they still effectively capture the trends and allows for the effect of various design decisions to be studied systematically. Musculoskeletal simulation based optimizations can effectively guide determination of most promising exoskeleton configurations and assistive torque profiles for human-in-the-loop testing.

\section*{Conclusions and Future Work}

This study has introduced a systematic simulation-based design approach to conduct a rigorous and fair comparison among different configurations of exoskeletons. The proposed design approach has been used to simultaneously optimize and compare the metabolic cost reduction of assisted subjects  and the power consumption of bi-articular and mono-articular exoskeletons.

In this study, we have studied a bi-articular exoskeleton configuration to assist hip and knee joints.  This exoskeleton configuration is motivated by human bi-articulation, which is known to improve human bipedal locomotion efficiency. The presence of bi-articular muscles in the human musculoskeletal system advances locomotion performance by enabling power transformation from proximal to distal joints and power regeneration between adjacent joints, facilitating joint movement coupling, resulting in effective distribution of muscle weight to reduce the leg inertia. The mono-articular exoskeleton, which assists each joint directly by mounting an actuator to the joint of interest, is motivated by its simplicity in design and wide-spread use.

The proposed multi-criteria optimization and comparison method subsumes single objective optimization solutions. One of these cases is for the optimization of the metabolic rate reduction of devices without considering their power consumption; these types of devices have been studied under the \emph{ideal exoskeleton} terminology in the literature~\cite{Uchida2016_idealexo_running,Dembia2017}. We have conducted simulations using ideal exoskeletons and showed that both ideal mono-articular and bi-articular exoskeletons can reach the same level of metabolic rate reduction and total power consumption. We have also showed that the assistance torques can considerably reduce the metabolic rate and the peak reaction forces/moments at the knee, patellofemoral, and hip joints. In addition to the direct effect of ideal exoskeletons on the muscular activities of the hip and knee joints, we have confirmed that these devices can also indirectly affect the activity of muscles at the ankle joint and the muscles responsible for hip abduction. Furthermore, these simulations have shown that loading subjects with a heavy load predictively changes the assistant profiles by magnitude and a time shift.

Although studying the ideal exoskeletons provides useful insights about these devices, it is necessary to analyze and compare exoskeletons under more realistic conditions that are applicable in real-life applications. Consequently, we have simultaneously optimized the performance of exoskeletons in terms of metabolic cost reduction and device power consumption using a multi-criteria optimization method. The multi-criteria optimization also enables fair comparisons of different exoskeleton configurations. 

Through the multi-criteria optimization of exoskeletons, we have shown that introducing sufficiently large actuator torque limits to both exoskeleton configurations does not have a large impact on the provided assistance, while it causes a considerable reduction on the power consumption. Additionally, we have shown that both exoskeleton configurations can reach similar performance levels, but for different peak torques assignments. In particular, we have shown that larger peak torque limits are required for mono-articular exoskeletons compared to bi-articular exoskeletons. Despite the similar assistance levels of both exoskeleton configurations, mono-articular exoskeletons demonstrate better performance in reducing the peak reaction forces/moments. By analyzing the Pareto front curves of both exoskeleton configurations under different loading conditions, we have shown that the power consumption of bi-articular exoskeletons is less affected by loading of subjects than mono-articular exoskeletons. Lastly, analyzing the power and torque profiles of non-dominated exoskeleton configurations indicate that the effect of loading  on the profiles of bi-articular exoskeletons is more uniform and predictable than the effect on the profiles of mono-articular exoskeletons.

\bigskip
We have also studied the effect of regeneration on the power consumption of exoskeletons. We have shown that regeneration can significantly improve the power consumption of exoskeletons from 6.54~$\pm$~2.60\% to 25.76~$\pm$~4.34\%, depending on the efficiency of regeneration, the kinematic configuration of the exoskeleton, and the actuator torque limitations.  Additionally, the analysis of actuators of each exoskeleton configuration has revealed that the knee actuators of mono-articular exoskeletons possess more generation potential, while both actuators of the bi-articular exoskeletons display large regeneration capacity.

To study the determent effects of  mass/inertia of exoskeletons on the metabolic cost of walking, we have superimposed the metabolic effect of device inertial properties to the Pareto-front curves. In particular, we have introduced a modification of augmentation factor and adaptation of the model developed by Browning~\etal~\cite{Browning2007} to estimate the effect of adding inertia and mass on metabolic rate of subjects. Our results indicate that non-dominated mono-articular exoskeletons loose their efficiency by 42.51~$\pm$~0.17\%--55.51~$\pm$~0.11\%, while non-dominated bi-articular devices are affected by 35.12 ~$\pm$~ 0.21\%--49.67~$\pm$~0.21\% by detrimental inertial affects. By considering the inertial effects on the metabolic rate, we have shown that Pareto optimal solutions of the bi-articular exoskeleton configuration are not significantly changed, whereas a different set of Pareto solutions needs to be considered for the mono-articular exoskeletons under inertial effects.


Our ongoing works include establishing experimental setups for mono-articular and bi-articular exoskeleton configurations, and validating the simulation based multi-criteria optimization results via a set of human subject experiments. 

Our future works include extending musculoskeletal simulations used in this study to  simulations that utilize dynamic optimization~\cite{Geijtenbeek2019,Dembia2019_Moco}, such that changes in kinematic and effects of inertial properties can also be captured while optimizing the mono-articular and the bi-articular exoskeleton configurations.

\section*{Acknowledgments}

This work has been partially supported  by T\"{U}B\.{I}TAK under Grant 216M200 and by Sabanc{\i} University.
\bigskip
\section*{Supporting documents}

\paragraph*{S1 Appendix.}
\label{S1_Appendix}
{\bf Kinematic models of mono-articular and bi-articular exoskeleton configurations.}

\paragraph*{S2 Fig.}
\label{S2_Appendix}
{\bf Gait phases of subjects under \textit{noload} and \textit{loaded} walking conditions.}

\paragraph*{S3 Appendix.}
\label{S3_Appendix}
{\bf Joint reaction forces/moments of the assisted subjects under \textit{noload} and \textit{loaded} walking conditions.}

\paragraph*{S4 Appendix.}
\label{S4_Appendix}
{\bf Torque profiles and muscles activities of selected non-dominated exoskeletons.}

\newpage
\nolinenumbers
\section*{S1 Appendix. -- Kinematic models of mono-articular and bi-articular exoskeleton configurations} 
\nolinenumbers
\renewcommand{\thefigure}{A\arabic{figure}}
\setcounter{figure}{0}
Both mono-articular and bi-articular exoskeletons are designed to assist the hip and knee joints. The bi-articular exoskeleton configurations are inspired from the bi-articular muscles and the aim to place majority of the actuator mass proximal to the hip joint, while delivering the required power to the knee joint though power transmission mechanisms. In this study, a parallelogram mechanism is considered to deliver power to the knee joint. Fig~\ref{Fig_Exos_Kinematics_Model_Appendix}(a) presents a kinematic model of the bi-articular exoskeleton with parallelogram mechanism, while Fig~\ref{Fig_Exos_Kinematics_Model_Appendix}(b) depicts the kinematic model of a typical mono-articular exoskeleton.

\begin{figure*}[h!]
	\centering
	\subfloat[\small{Bi-articular exoskeleton}]{\includegraphics[width=2in]{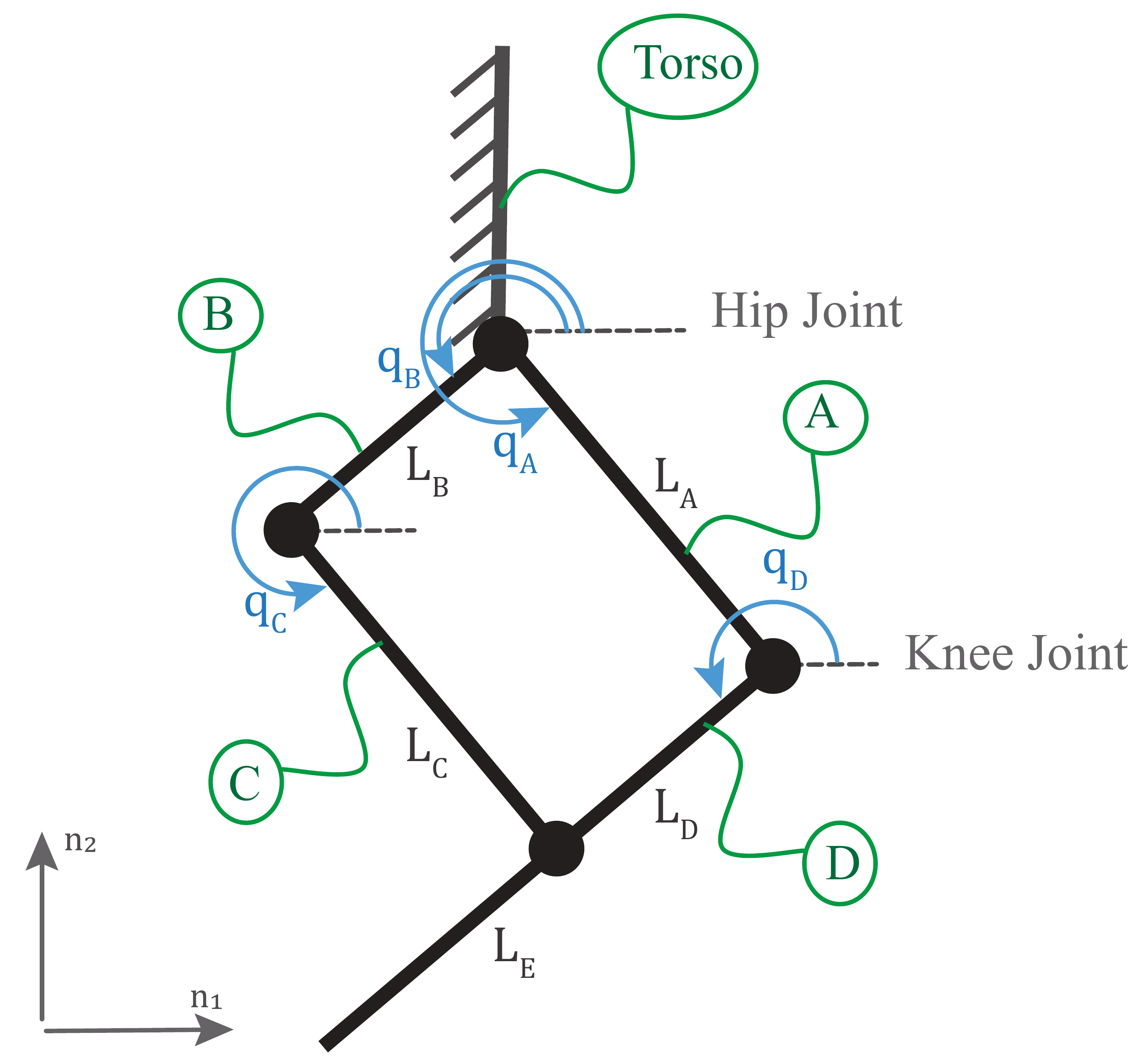}
		\label{Fig_Biarticular_Exo_Mechanism}}
	\hfil
	\subfloat[\small{Mono-articular exoskeleton}]{\includegraphics[width=2in]{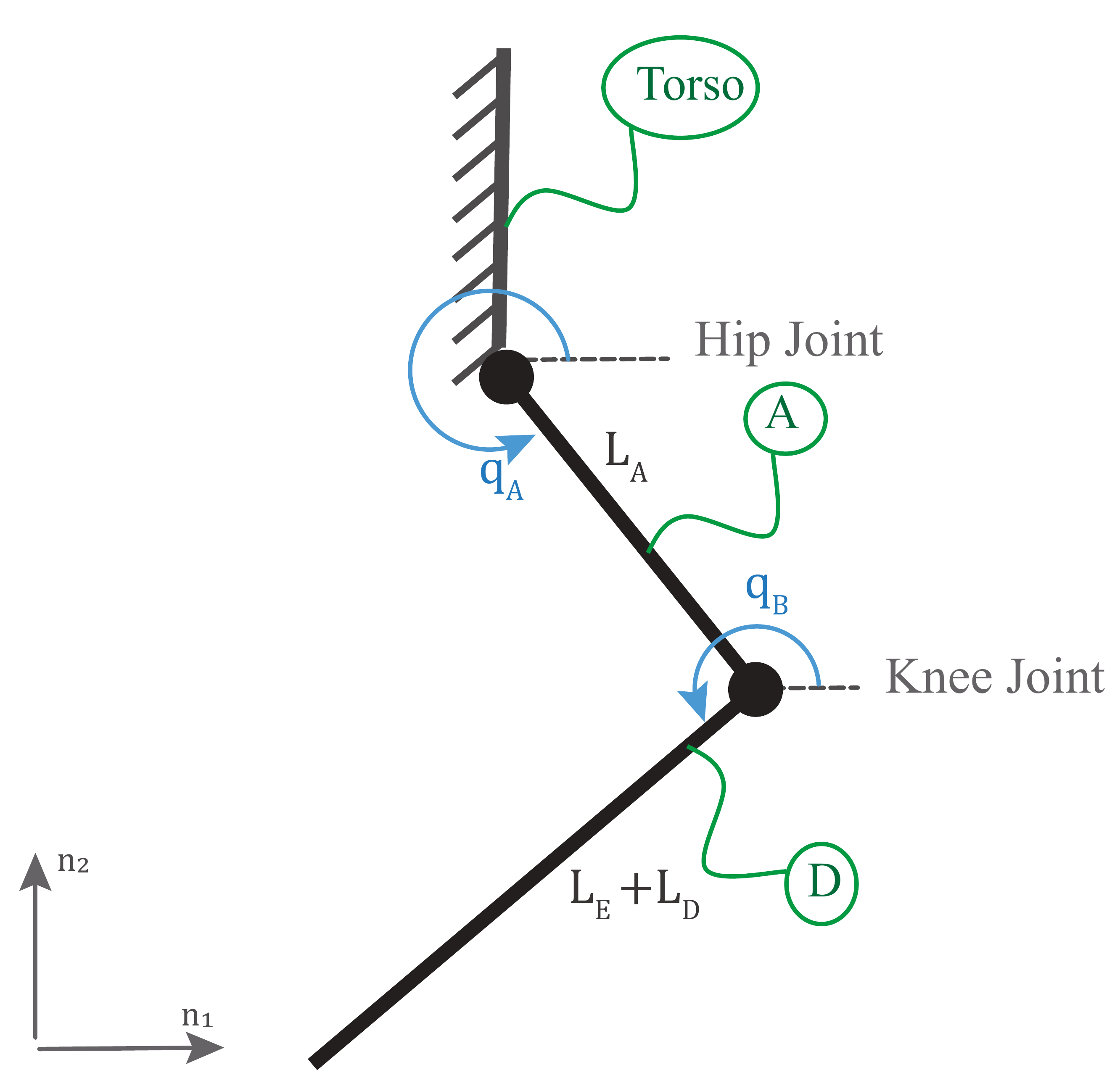}
		\label{Fig_Monoarticular_Exo_Mechanism}}
	\vspace{1mm}
	\caption{\small{\textbf{Kinematic models of mono-articular and bi-articular exoskeleton configurations.} The mono-articular exoskeleton is modeled as a two link serial manipulator, while a parallelogram is used for the bi-articular exoskeleton.}}
	\label{Fig_Exos_Kinematics_Model_Appendix}
\end{figure*}

The configuration-level forward kinematics of the bi-articular exoskeleton can be expressed as
\begin{align}\label{Eqn_Bi_Kin}
x_{bi} &= l_{A}\cos(q_{A}) + (l_{E} + l_ {D})\cos(q_{B})\\
y_{bi} &= l_{A}\sin(q_{A}) + (l_{E} + l_ {D})\sin(q_{B}),
\end{align}
\noindent while the configuration-level forward kinematics of the mono-articular exoskeleton reads as
\begin{align}\label{Eqn_Mono_Kin}
x_{mono} &= l_ {A}\cos(q_{A}) + (l_{E} + l_{D})\cos(q_{A} - q_{B})\\
y_{mono} &= l_ {A}\sin(q_{A}) + (l_{E} + l_{D})\sin(q_{A} - q_{B}).
\end{align}

Taking the time derivative of these equations, the kinematic Jacobians of bi-articular and mono-articular exoskeletons can be expressed as

\begin{gather}\label{Eqn_Bi_Jacobian}
J_{Bi} =
\begin{bmatrix}
-l_{A}\sin{q_{A}}  & -(l_ {E} + l_ {D})\sin (q_ {B})\\
l_ {A}\cos (q_{A}) &  (l_ {E} + l_ {D})\cos (q_ {B})
\end{bmatrix}
\end{gather}

\begin{equation}\label{Eqn_Mono_Jacobian}
\begin{aligned}
J_{Mono}&=
\left[\begin{matrix}
-l_{A}\sin{q_{A}}- (l_ {E} + l_ {D})\sin (q_ {A} - q_ {B})+ (l_ {E} + l_ {D})\sin (q_ {A} - q_ {B})\\
l_{A}\cos{q_{A}} + (l_{E} + l_{D})\cos (q_{A} - q_ {B})- (l_ {E} + l_ {D})\cos (q_ {A} - q_ {B})
\end{matrix}\right]
\end{aligned}
\end{equation}

\newpage
\section*{S2 Fig. -- Gait phases of subjects under \textit{noload} and \textit{loaded} walking conditions} 
\renewcommand{\figurename}{S2 Fig}
\renewcommand{\thefigure}{\arabic{figure}}
\setcounter{figure}{0}
\bigskip
\bigskip
\bigskip
\bigskip
\bigskip
\nolinenumbers
\begin{figure*}[ht]
	\centering
	\includegraphics[width=\linewidth]{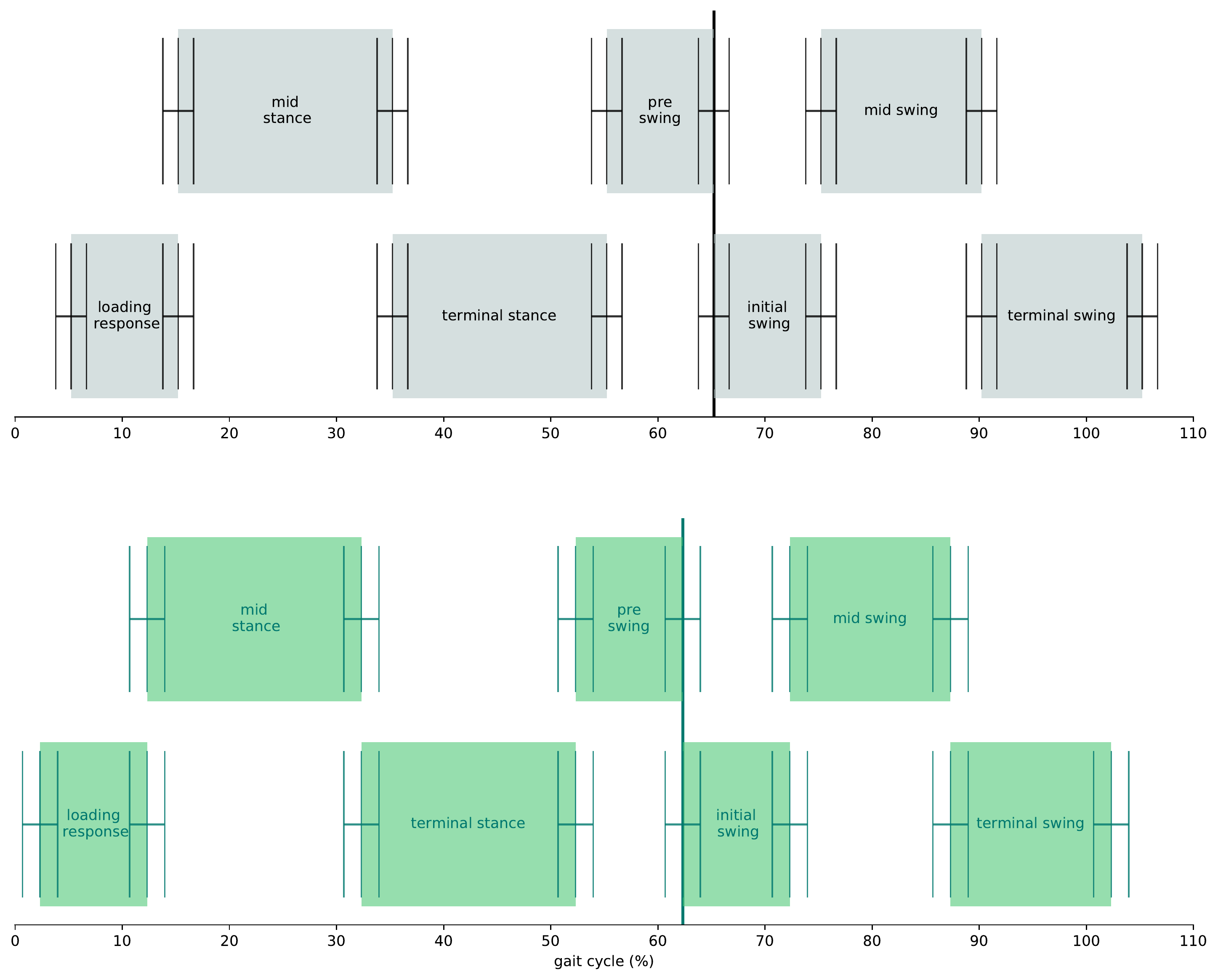}
	\vspace{-3mm}
	\caption{{\small\textbf{Gait phases of subjects } The upper row presents the phases of gait under \textit{noload} walking condition, while the lower row presents the phases of gait under  \textit{loaded} walking condition.}}
	\label{Fig_S2}
\end{figure*}

\newpage
\section*{S3 Appendix. -- Joint reaction forces/moments of the assisted subjects under \textit{noload} and \textit{loaded} walking conditions} 
\renewcommand{\figurename}{Fig}
\renewcommand{\thefigure}{B\arabic{figure}}
\setcounter{figure}{0}
\bigskip
\bigskip
\bigskip
\bigskip
\nolinenumbers
\begin{figure*}[h!]
	\centering \includegraphics[width=\linewidth]{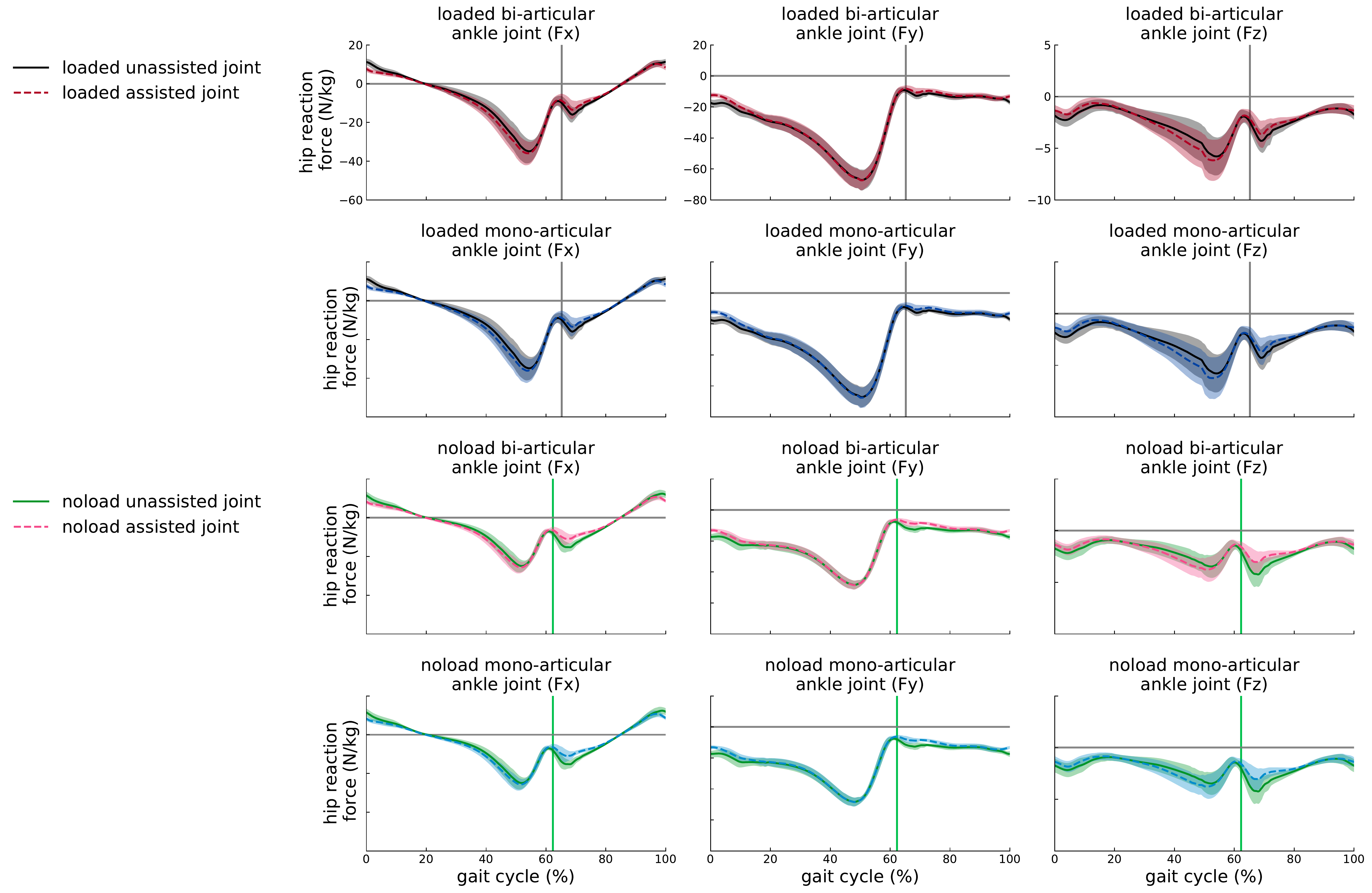}
	\vspace{1mm}
	\caption{\small{\textbf{Ideal devices' effect on joint reaction forces of the ankle joint.}} The reaction forces of the ankle joint in anterior-posterior ($F_x$), compressive ($F_y$), and medial-lateral ($F_z$) directions. The blue and red shades represent the reaction forces of subjects assisted by ideal monoarticular and biarticular exoskeletons, respectively. The black and green profiles represent the reaction forces of unassisted subjects in {\it loaded} and {\it noload} conditions, respectively. The curves are averaged over 7 subjects with 3 trials and normalized by subject mass; shaded regions around the mean profile indicate standard deviation of the profile.}
	\label{Fig_Ankle_ReactionForces}
\end{figure*}
\begin{figure*}[h!]
	\centering
	\includegraphics[width=\linewidth]{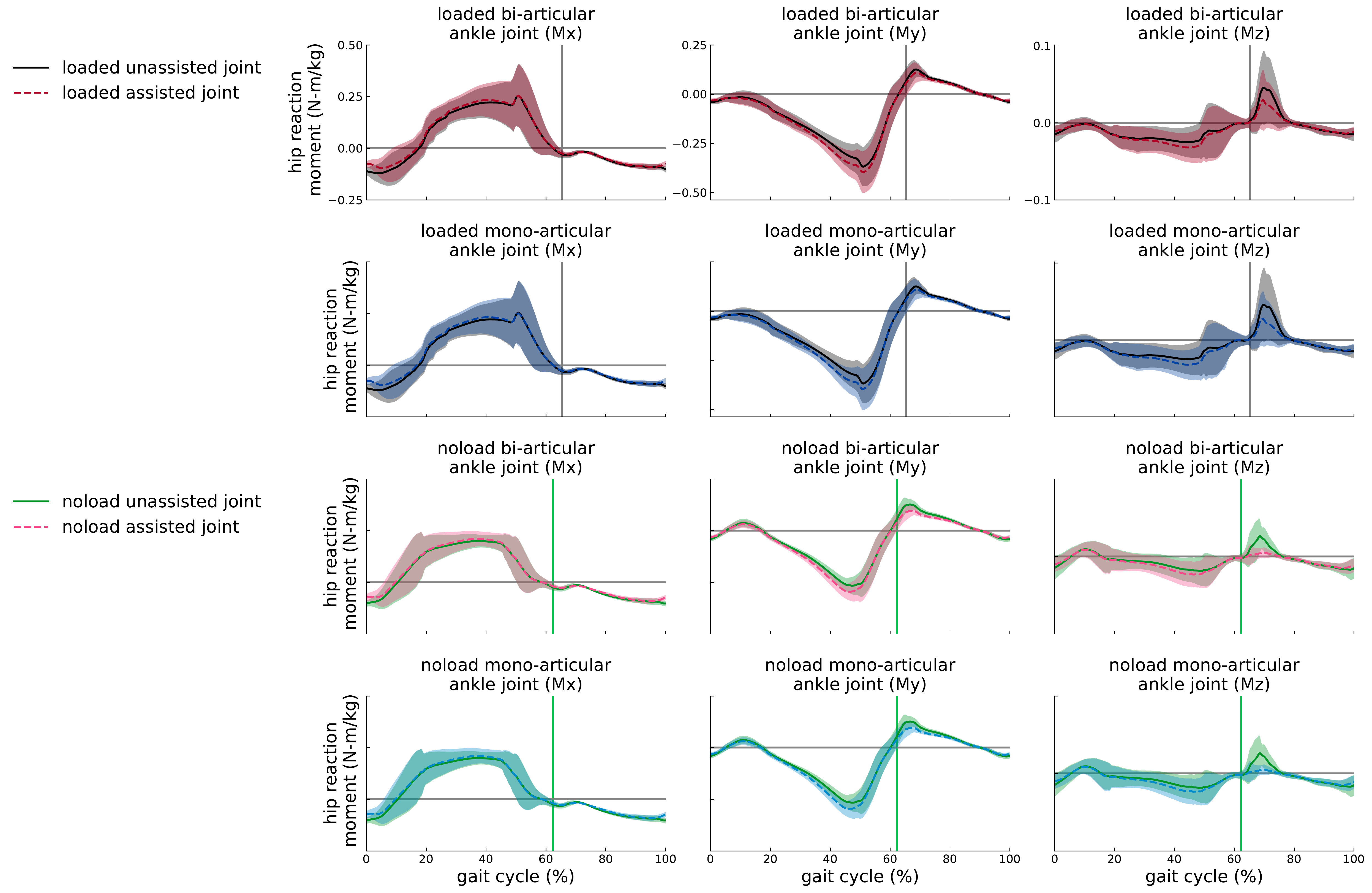}
	\vspace{1mm}
	\caption{\small{\textbf{Ideal devices effect on joint reaction moments of the ankle joint.}} The reaction moments of the ankle joint in adduction-abduction ($M_x$), internal-external rotation ($M_y$), and medial-lateral ($M_z$) directions. The blue and red shades represent the reaction moments of subjects assisted by ideal monoarticular and biarticular exoskeletons, respectively. The black and green profiles represent the reaction forces of unassisted subjects in {\it loaded} and {\it noload} conditions, respectively. The curves are averaged over 7 subjects with 3 trials and normalized by subject mass; shaded regions around the mean profile indicate standard deviation of the profile.}
	\label{Fig_Ankle_ReactionMomentss}
\end{figure*}
\begin{figure*}[h!]
	\centering
	\includegraphics[width=\linewidth]{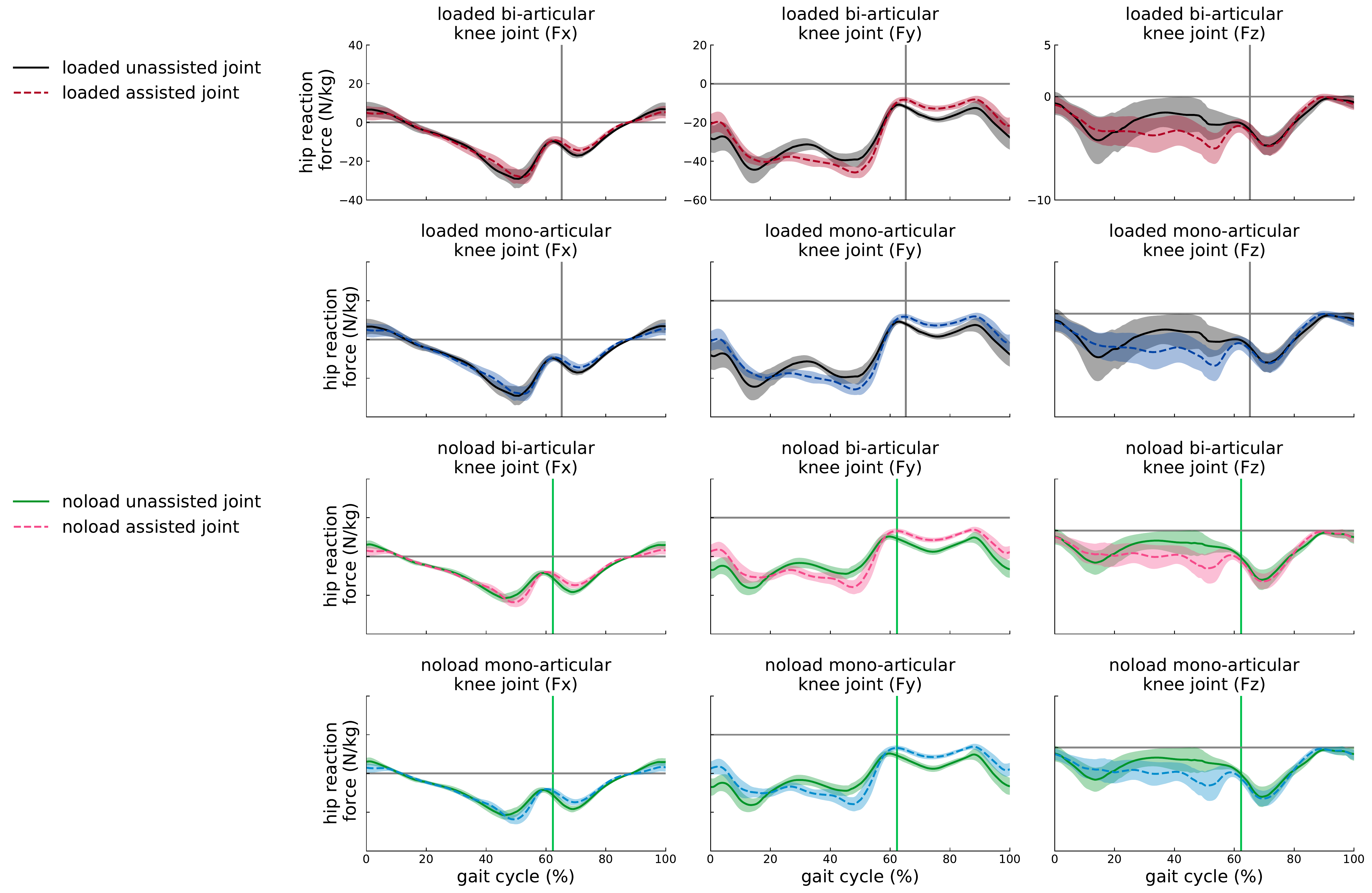}
	\vspace{1mm}
	\caption{\small{\textbf{Ideal devices effect on joint reaction forces of the knee joint.}} The reaction forces of the knee joint in anterior-posterior ($F_x$), compressive ($F_y$, i.e., tibiofemoral force), and medial-lateral ($F_z$) directions. The blue and red shades represent the reaction forces of subjects assisted by ideal monoarticular and biarticular exoskeletons, respectively. The black and green profiles represent the reaction forces of unassisted subjects in {\it loaded} and {\it noload} conditions, respectively. The curves are averaged over 7 subjects with 3 trials and normalized by subject mass; shaded regions around the mean profile indicate standard deviation of the profile.}
	\label{Fig_Knee_ReactionForces}
\end{figure*}
\begin{figure*}[h!]
	\centering
	\includegraphics[width=\linewidth]{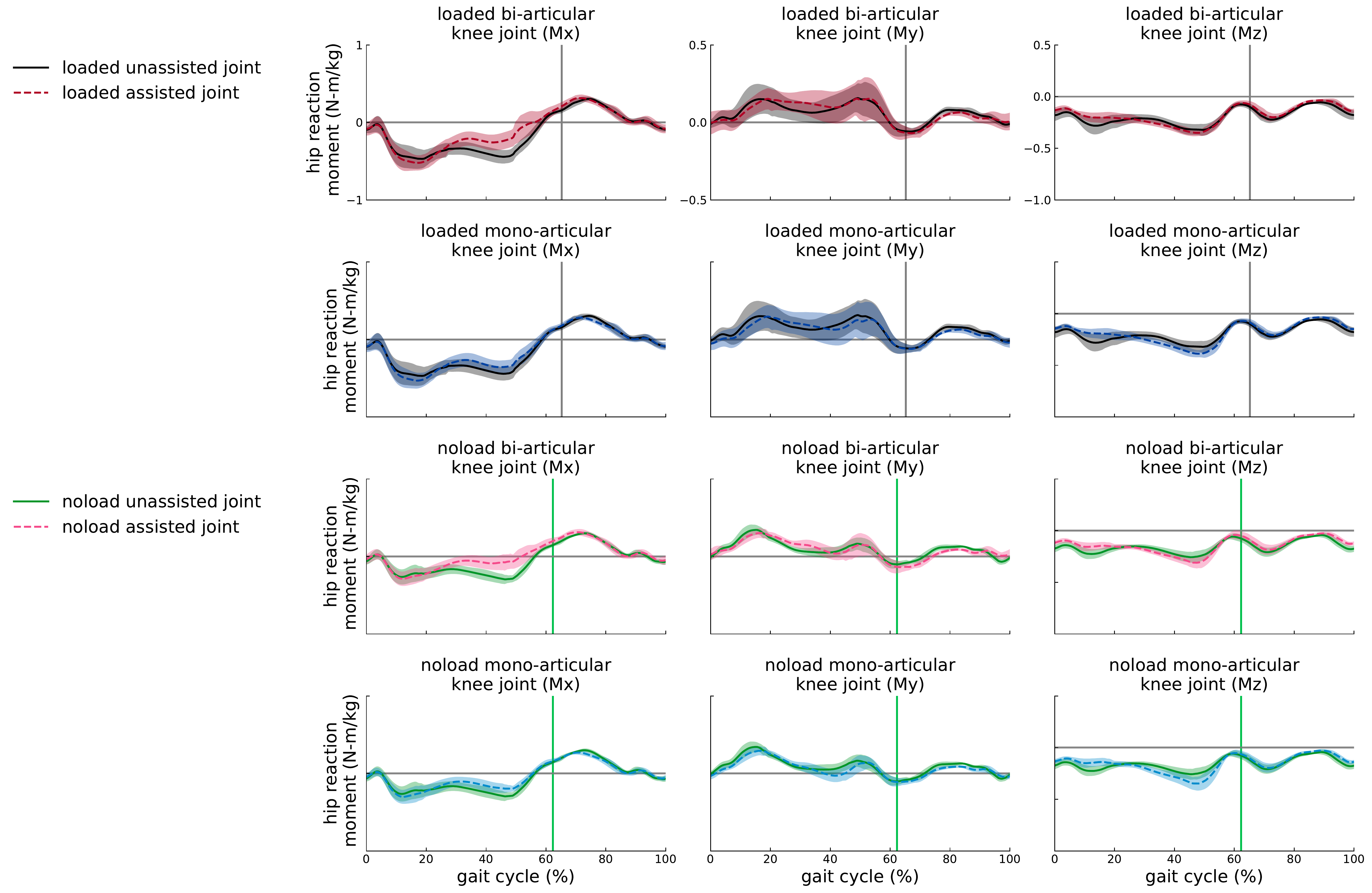}
	\vspace{1mm}
	\caption{\small{\textbf{Ideal devices effect on joint reaction moments of the knee joint.}} The reaction moments of the knee joint in adduction-abduction ($M_x$), internal-external rotation ($M_y$), and medial-lateral ($M_z$) directions. The blue and red shades represent the reaction moments of subjects assisted by ideal monoarticular and biarticular exoskeletons, respectively. The black and green profiles represent the reaction forces of unassisted subjects in {\it loaded} and {\it noload} conditions, respectively. The curves are averaged over 7 subjects with 3 trials and normalized by subject mass; shaded regions around the mean profile indicate standard deviation of the profile.}
	\label{Fig_Knee_ReactionMomentss}
\end{figure*}
\begin{figure*}[h!]
	\centering
	\includegraphics[width=\linewidth]{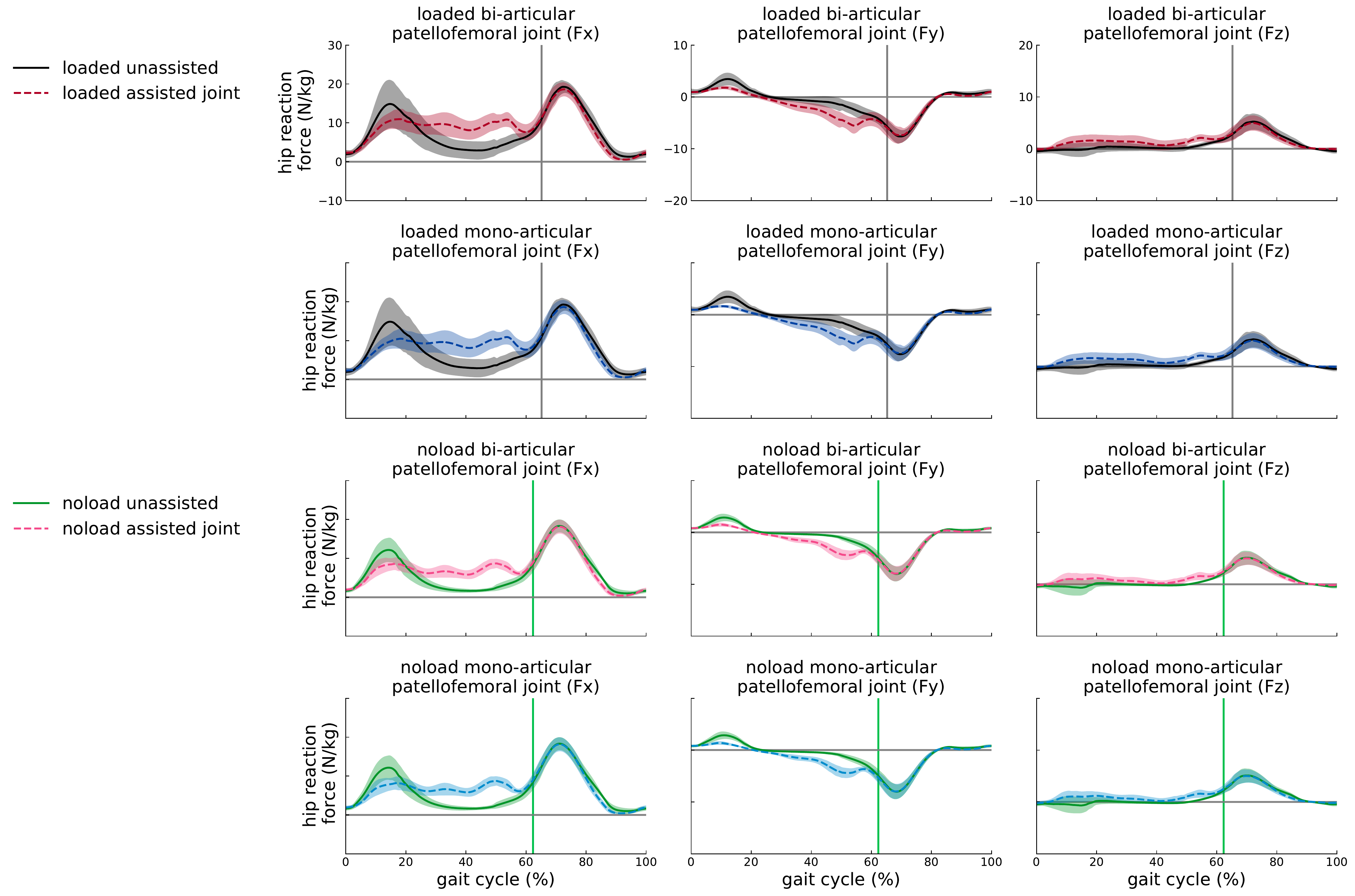}
	\vspace{1mm}
	\caption{\small{\textbf{Ideal devices effect on joint reaction forces of the patellofemoral joint.}} The reaction forces of the patellofemoral joint in anterior-posterior ($F_x$), compressive ($F_y$), and medial-lateral ($F_z$) directions. The blue and red shades represent the reaction forces of subjects assisted by ideal monoarticular and biarticular exoskeletons, respectively. The black and green profiles represent the reaction forces of unassisted subjects in {\it loaded} and {\it noload} conditions, respectively. The curves are averaged over 7 subjects with 3 trials and normalized by subject mass; shaded regions around the mean profile indicate standard deviation of the profile.}
	\label{Fig_Patellofemoral_ReactionForces}
\end{figure*}
\begin{figure*}[h!]
	\centering
	\includegraphics[width=\linewidth]{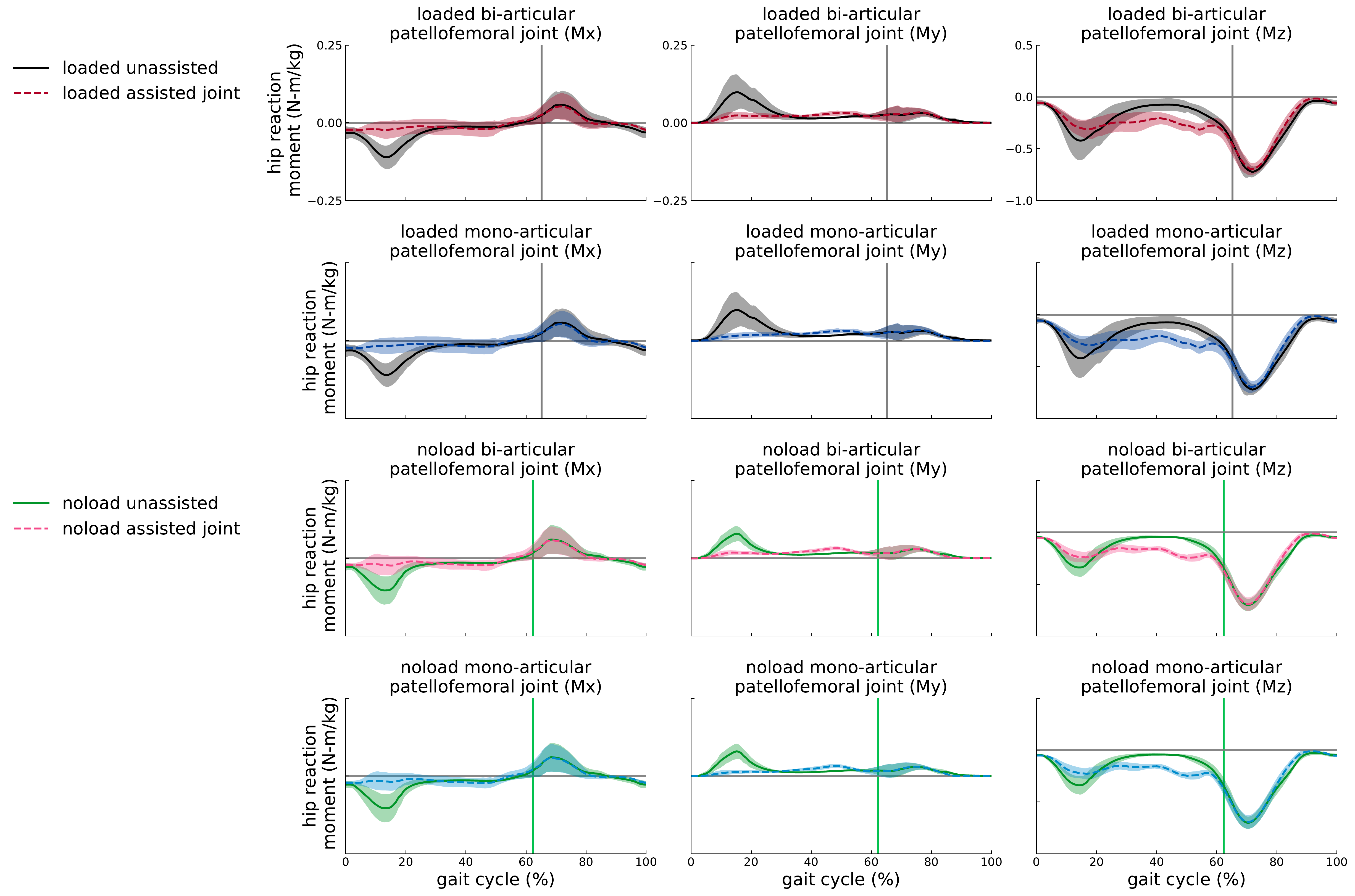}
	\vspace{1mm}
	\caption{\small{\textbf{Ideal devices effect on joint reaction moments of the patellofemoral joint.}} The reaction moments of the patellofemoral joint in adduction-abduction ($M_x$), internal-external rotation ($M_y$), and medial-lateral ($M_z$) directions. The blue and red shades represent the reaction moments of subjects assisted by ideal monoarticular and biarticular exoskeletons, respectively. The black and green profiles represent the reaction forces of unassisted subjects in {\it loaded} and {\it noload} conditions, respectively. The curves are averaged over 7 subjects with 3 trials and normalized by subject mass; shaded regions around the mean profile indicate standard deviation of the profile.}
	\label{Fig_Patellofemoral_ReactionMoments}
\end{figure*}
\begin{figure*}[h!]
	\centering
	\includegraphics[width=\linewidth]{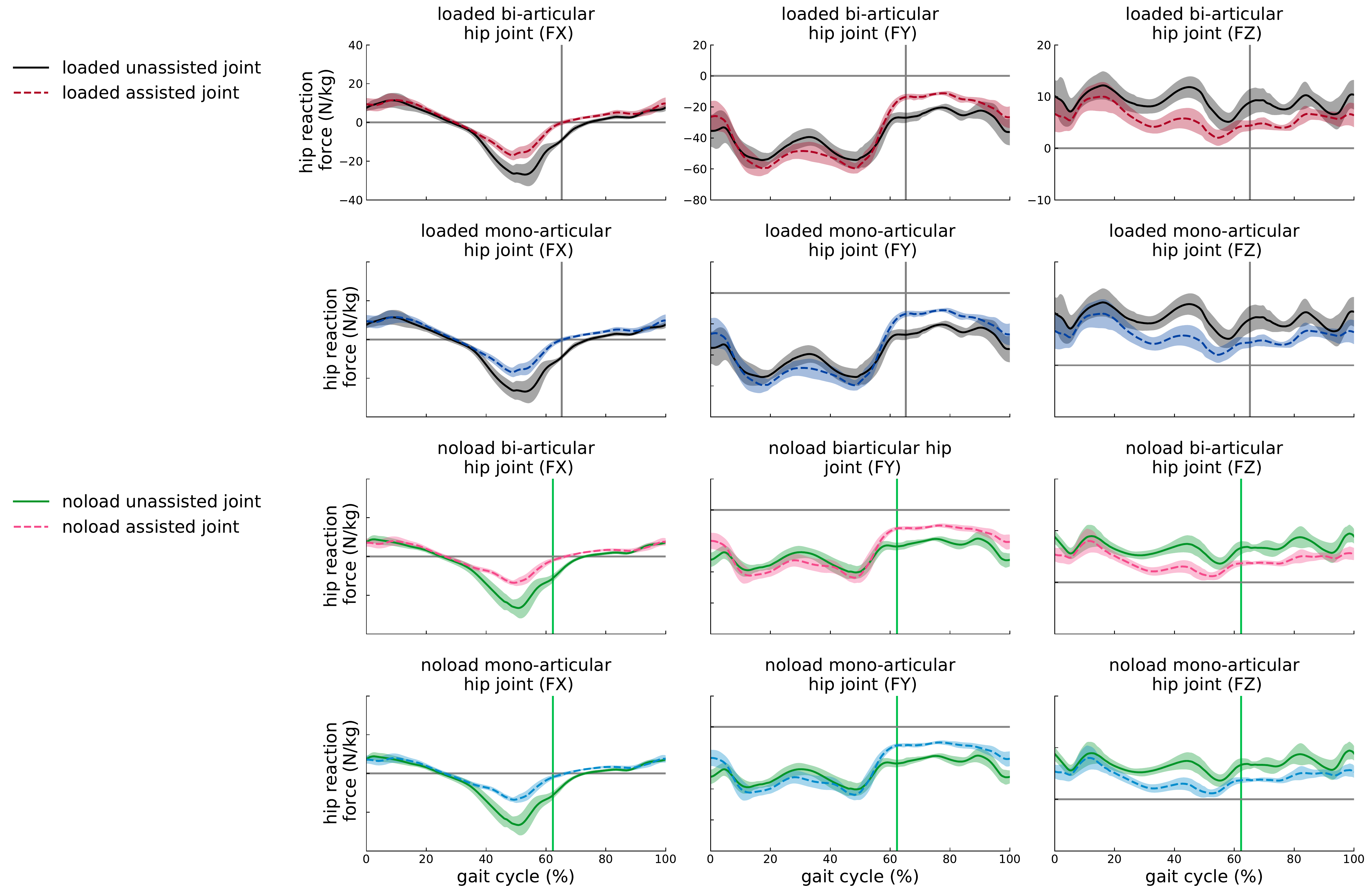}
	\vspace{1mm}
	\caption{\small{\textbf{Ideal devices effect on joint reaction forces of the hip joint.}} The reaction forces of the hip joint in anterior-posterior ($F_x$), compressive ($F_y$), and medial-lateral ($F_z$) directions. The blue and red shades represent the reaction forces of subjects assisted by ideal monoarticular and biarticular exoskeletons, respectively. The black and green profiles represent the reaction forces of unassisted subjects in {\it loaded} and {\it noload} conditions, respectively. The curves are averaged over 7 subjects with 3 trials and normalized by subject mass; shaded regions around the mean profile indicate standard deviation of the profile.}
	\label{Fig_Hip_ReactionForces}
\end{figure*}

\begin{figure*}[h!]
	\centering
	\includegraphics[width=\linewidth]{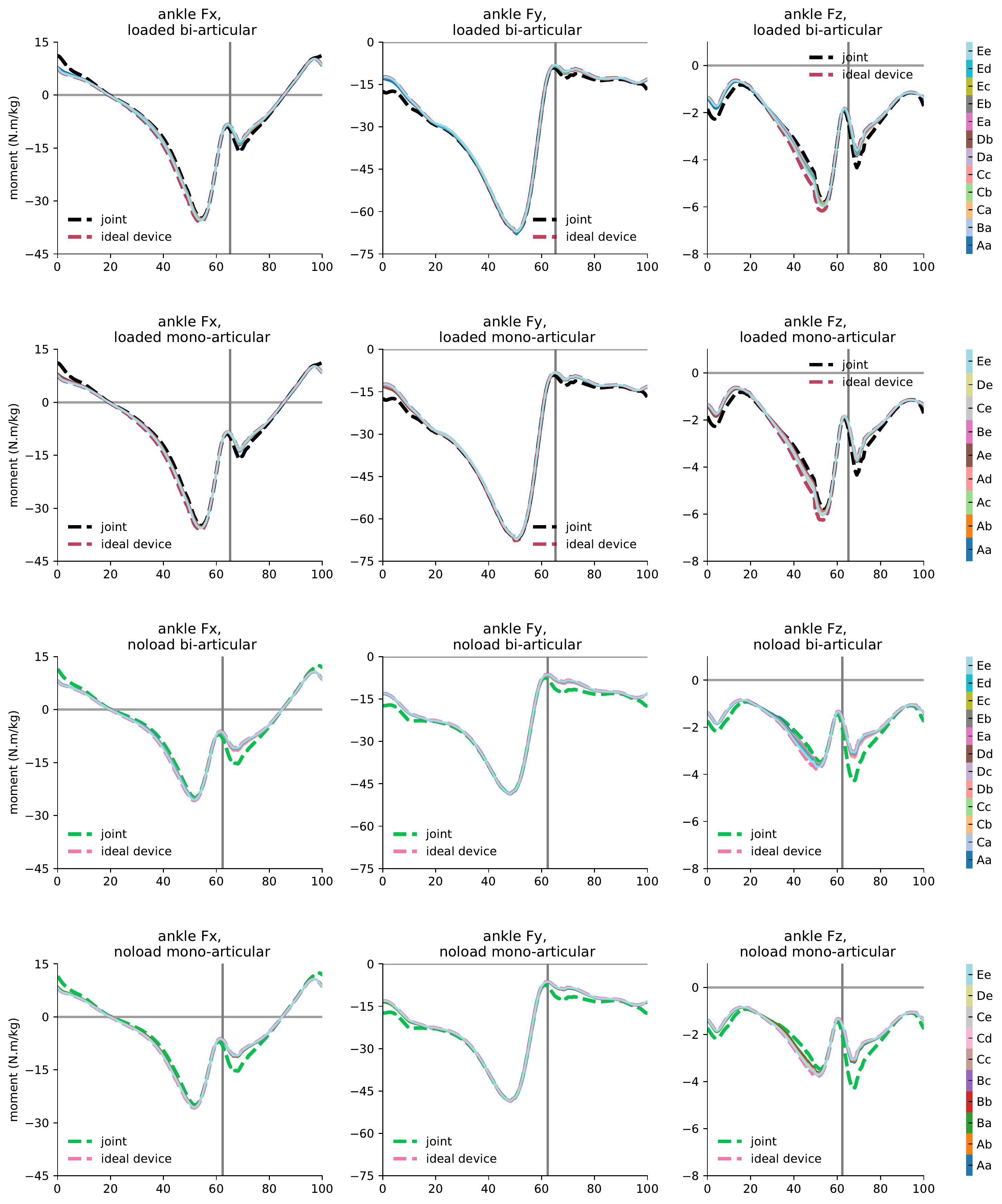}
	\vspace{1mm}
	\caption{\small{\textbf{Optimal devices effect on joint reaction forces of the ankle joint.}} The reaction forces of the ankle joint in anterior-posterior ($F_x$), compressive ($F_y$), and medial-lateral ($F_z$) directions. The color bars represent the reaction forces of subjects assisted by constrained optimal exoskeletons. The black and green profiles represent the reaction forces of unassisted subjects in {\it loaded} and {\it noload} conditions, respectively. The curves are averaged over 7 subjects with 3 trials and normalized by subject mass.}
	\label{Fig_Ankle_ReactionForces_Paretofront}
\end{figure*}
\begin{figure*}[h!]
	\centering
	\includegraphics[width=\linewidth]{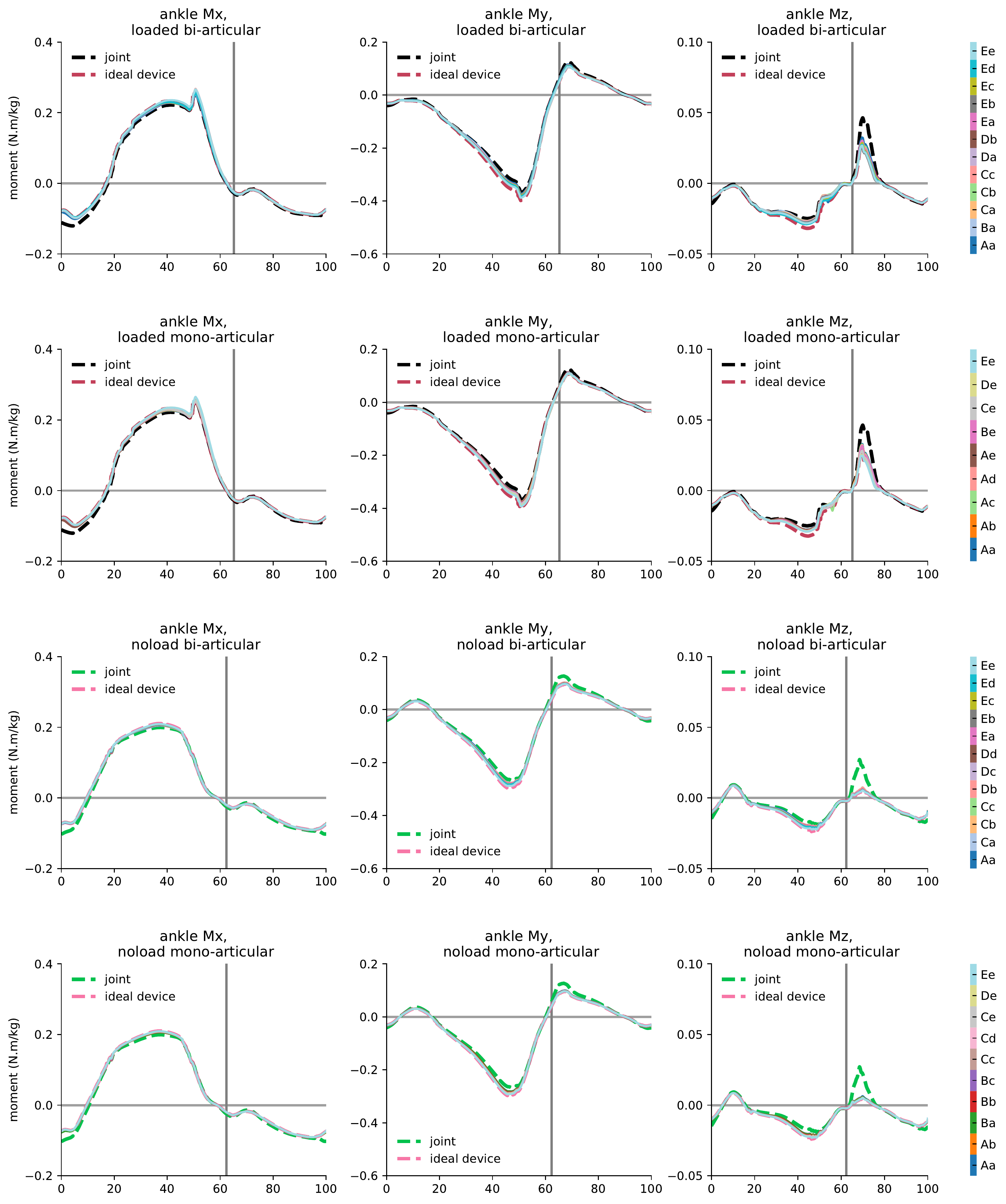}
	\vspace{1mm}
	\caption{\small{\textbf{Optimal devices effect on joint reaction moments of the ankle joint.}} The reaction moments of the ankle joint in adduction-abduction ($M_x$), internal-external rotation ($M_y$), and medial-lateral ($M_z$) directions. The color bars represent the reaction moments of subjects assisted by constrained optimal exoskeletons. The black and green profiles represent the reaction forces of unassisted subjects in {\it loaded} and {\it noload} conditions, respectively. The curves are averaged over 7 subjects with 3 trials and normalized by subject mass.}
	\label{Fig_Ankle_ReactionMoments_Paretofront}
\end{figure*}
\begin{figure*}[h!]
	\centering
	\includegraphics[width=\linewidth]{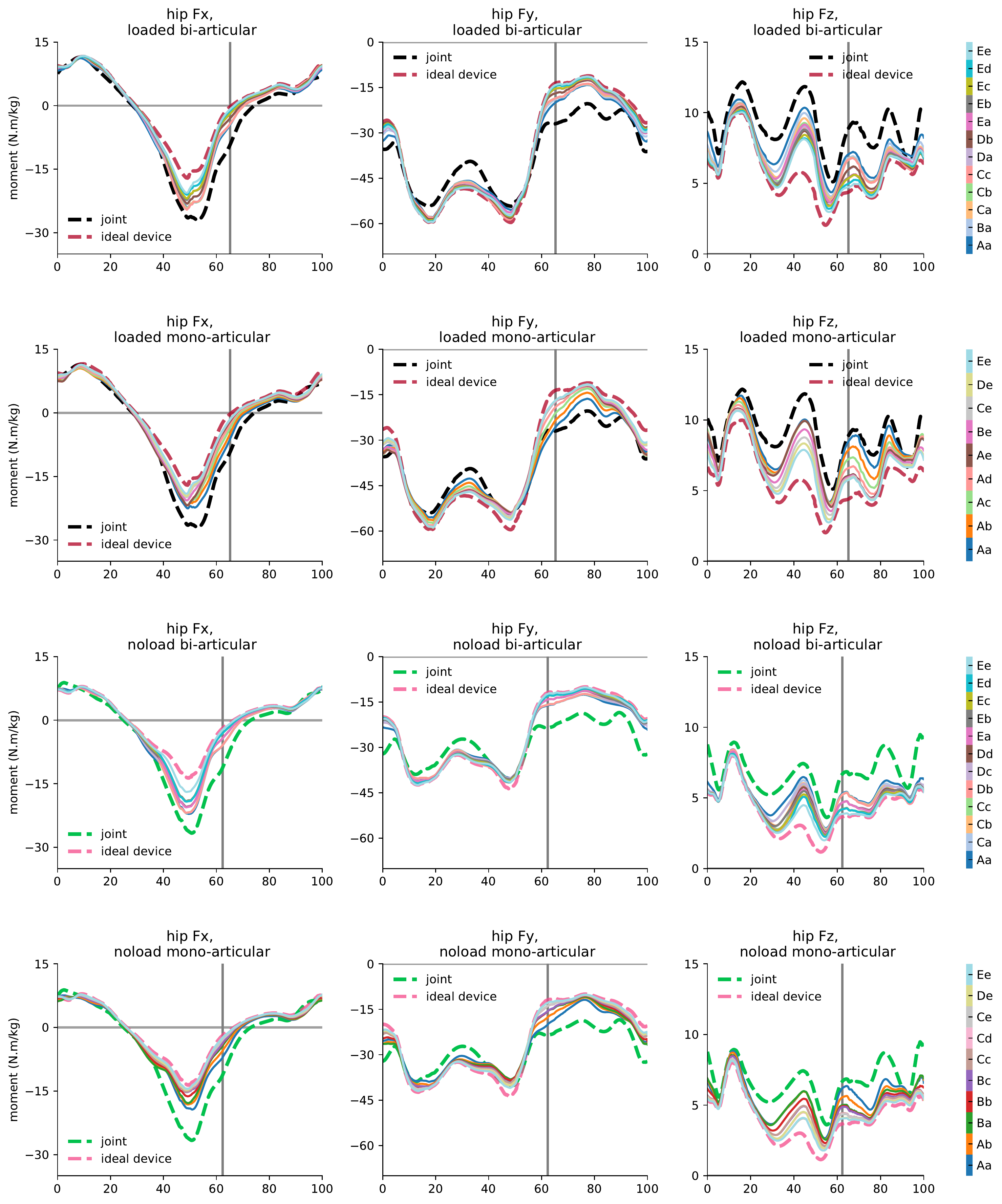}
	\vspace{1mm}
	\caption{\small{\textbf{Optimal devices effect on joint reaction forces of the hip joint.}} The reaction forces of the hip joint in anterior-posterior ($F_x$), compressive ($F_y$), and medial-lateral ($F_z$) directions. The color bars represent the reaction forces of subjects assisted by constrained optimal exoskeletons. The black and green profiles represent the reaction forces of unassisted subjects in {\it loaded} and {\it noload} conditions, respectively. The curves are averaged over 7 subjects with 3 trials and normalized by subject mass.}
	\label{Fig_Hip_ReactionForces_Paretofront}
\end{figure*}
\begin{figure*}[h!]
	\centering
	\includegraphics[width=\linewidth]{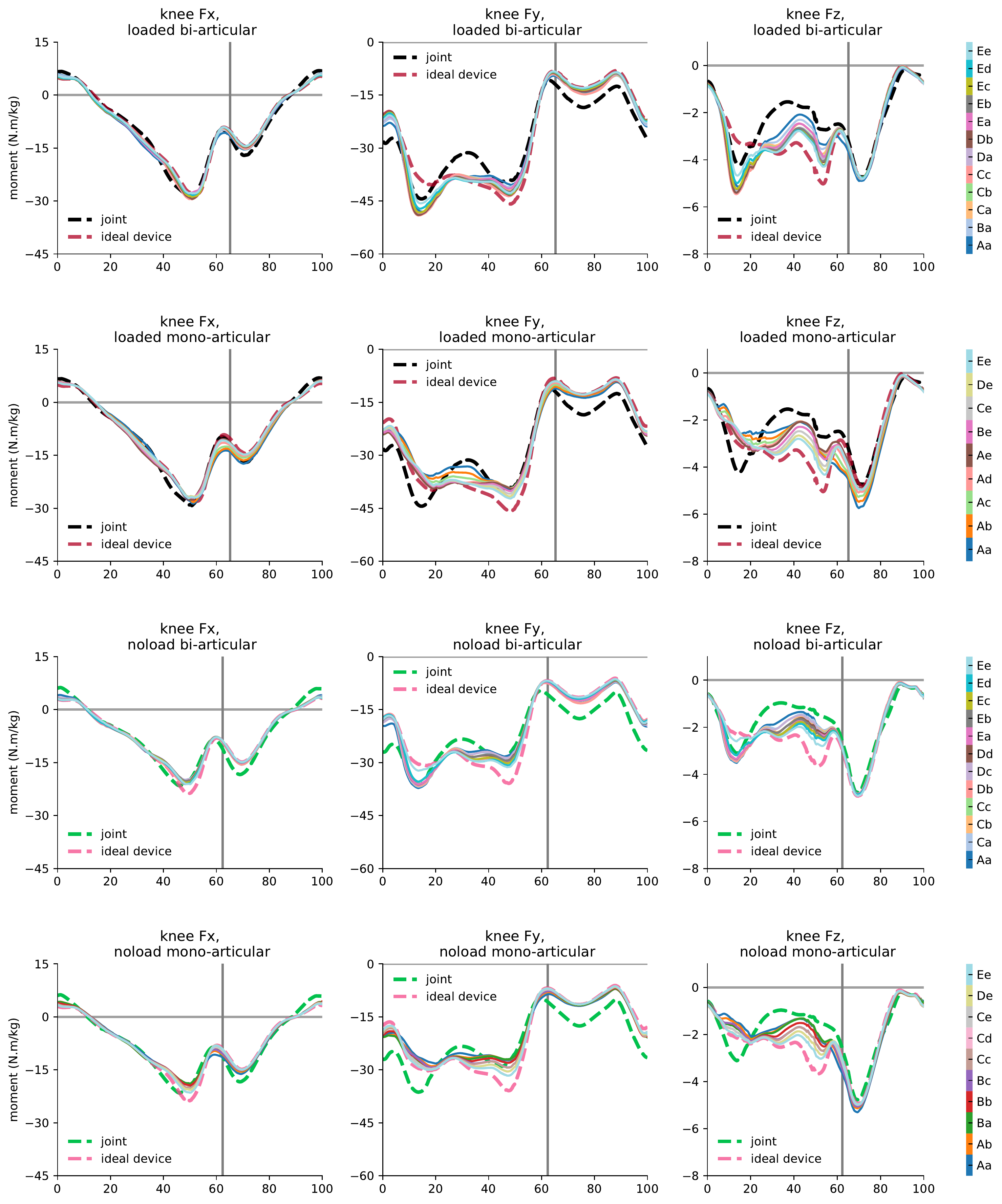}
	\vspace{1mm}
	\caption{\small{\textbf{Optimal devices effect on joint reaction forces of the knee joint.}} The reaction forces of the knee joint in anterior-posterior ($F_x$), compressive ($F_y$), and medial-lateral ($F_z$) directions. The color bars represent the reaction forces of subjects assisted by constrained optimal exoskeletons. The black and green profiles represent the reaction forces of unassisted subjects in {\it loaded} and {\it noload} conditions, respectively. The curves are averaged over 7 subjects with 3 trials and normalized by subject mass.}
	\label{Fig_Knee_ReactionForces_Paretofront}
\end{figure*}
\begin{figure*}[h!]
	\centering
	\includegraphics[width=\linewidth]{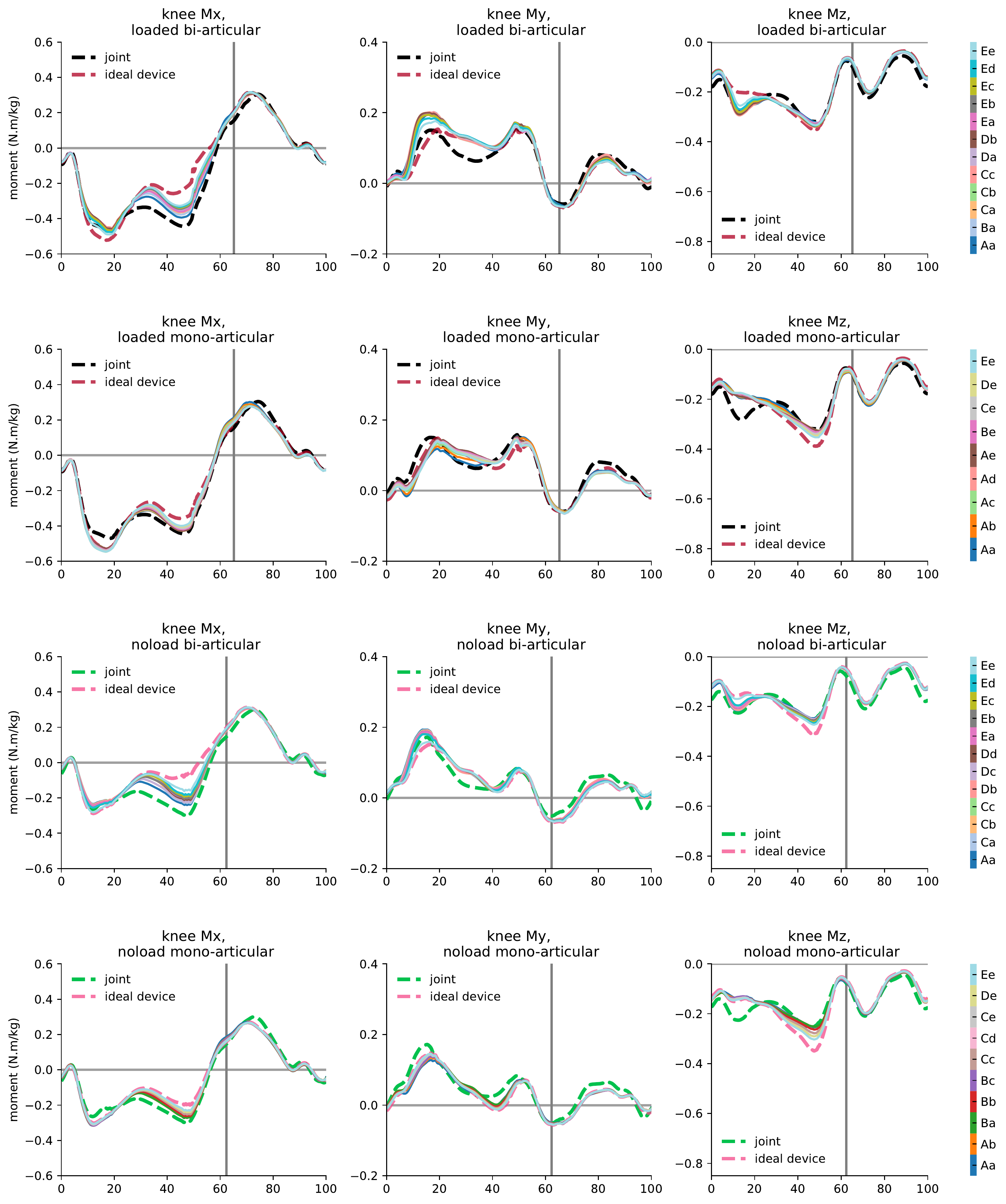}
	\vspace{1mm}
	\caption{\small{\textbf{Optimal devices effect on joint reaction moments of the knee joint.}} The reaction moments of the knee joint in adduction-abduction ($M_x$), internal-external rotation ($M_y$), and medial-lateral ($M_z$) directions. The color bars represent the reaction forces of subjects assisted by constrained optimal exoskeletons. The black and green profiles represent the reaction moments of unassisted subjects in {\it loaded} and {\it noload} conditions, respectively. The curves are averaged over 7 subjects with 3 trials and normalized by subject mass.}
	\label{Fig_Knee_ReactionMoments_Paretofront}
\end{figure*}
\begin{figure*}[h!]
	\centering
	\includegraphics[width=\linewidth]{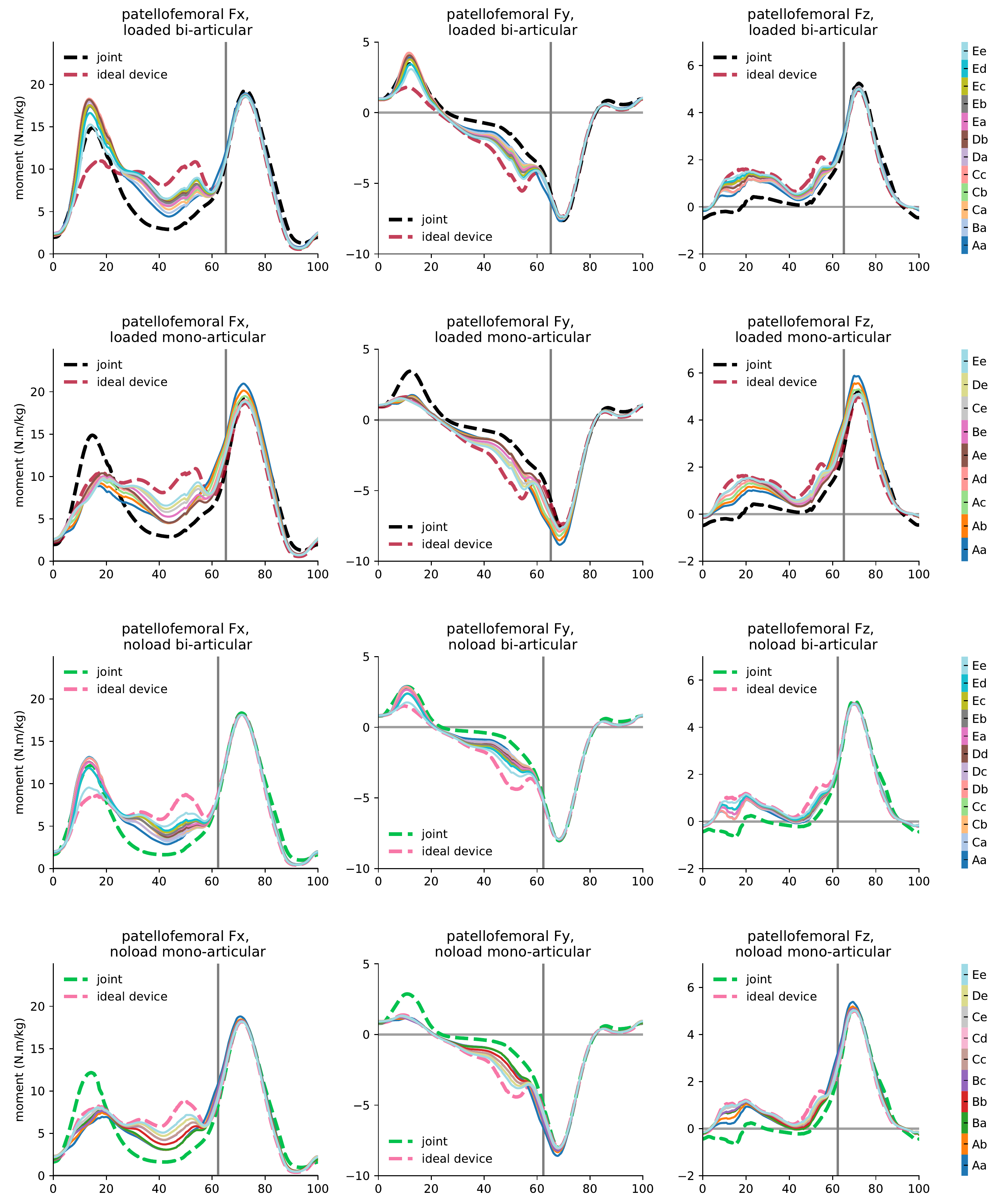}
	\vspace{1mm}
	\caption{\small{\textbf{Optimal devices effect on joint reaction forces of the patellofemoral joint.}} The reaction forces of the patellofemoral joint in anterior-posterior ($F_x$), compressive ($F_y$), and medial-lateral ($F_z$) directions. The color bars represent the reaction forces of subjects assisted by constrained optimal exoskeletons. The black and green profiles represent the reaction forces of unassisted subjects in {\it loaded} and {\it noload} conditions, respectively. The curves are averaged over 7 subjects with 3 trials and normalized by subject mass.}
	\label{Fig_Patellofemoral_ReactionForces_Paretofront}
\end{figure*}
\begin{figure*}[h!]
	\centering
	\includegraphics[width=\linewidth]{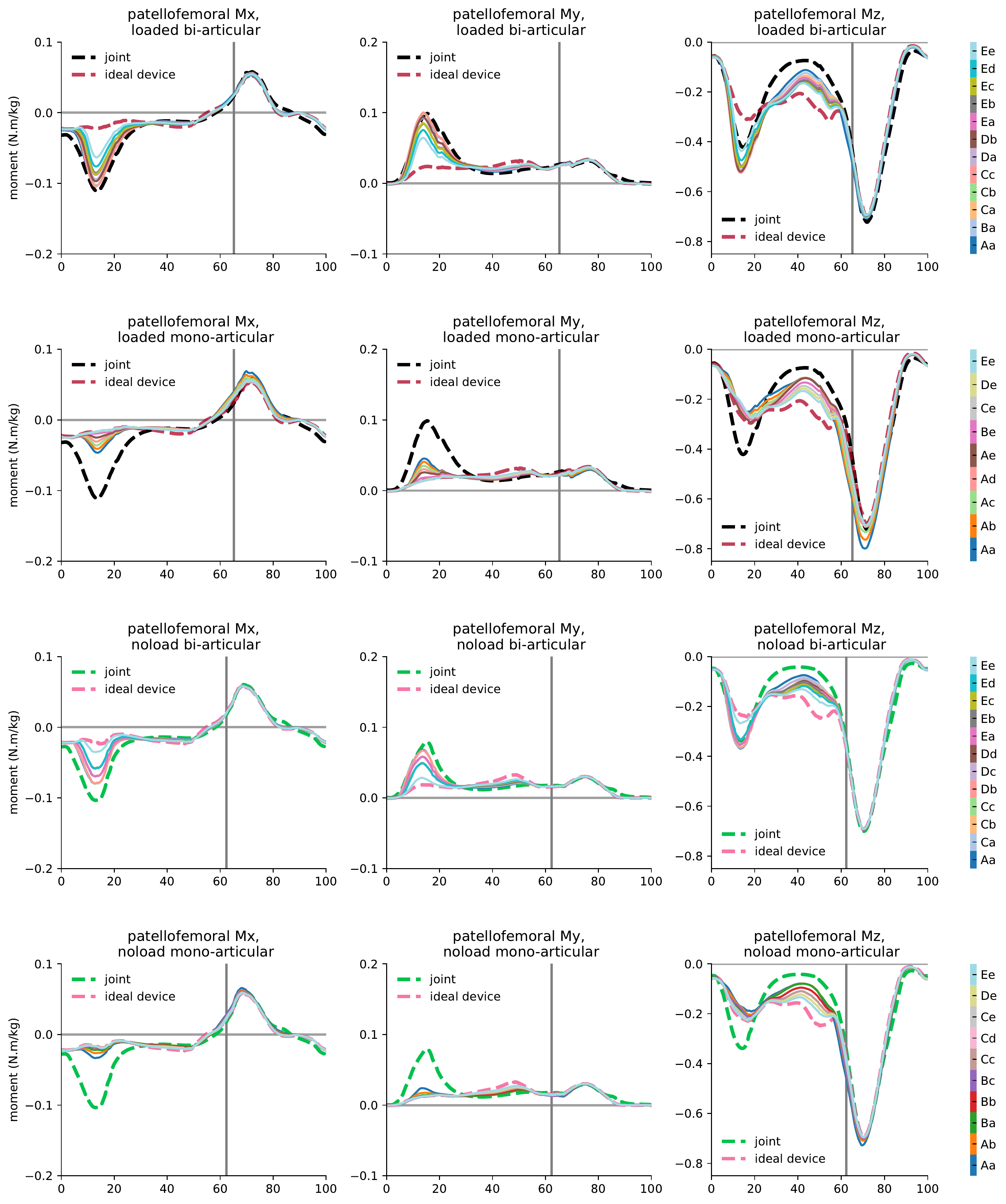}
	\vspace{1mm}
	\caption{\small{\textbf{Optimal devices effect on joint reaction moments of the patellofemoral joint.}} The reaction moments of the patellofemoral joint in adduction-abduction ($M_x$), internal-external rotation ($M_y$), and medial-lateral ($M_z$) directions. The color bars represent the reaction moments of subjects assisted by constrained optimal exoskeletons. The black and green profiles represent the reaction forces of unassisted subjects in {\it loaded} and {\it noload} conditions, respectively. The curves are averaged over 7 subjects with 3 trials and normalized by subject mass.}
	\label{Fig_Patellofemoral_ReactionMoments_Paretofront}
\end{figure*}

\newpage
\clearpage
\section*{S4 Appendix. -- Torque profiles and muscles activities of selected non-dominated exoskeletons} 
\bigskip
\bigskip
\bigskip
\bigskip
\nolinenumbers
\renewcommand{\thefigure}{C\arabic{figure}}
\setcounter{figure}{0}
\begin{figure*}[ht]   
	\centering
	\includegraphics[width=\linewidth]{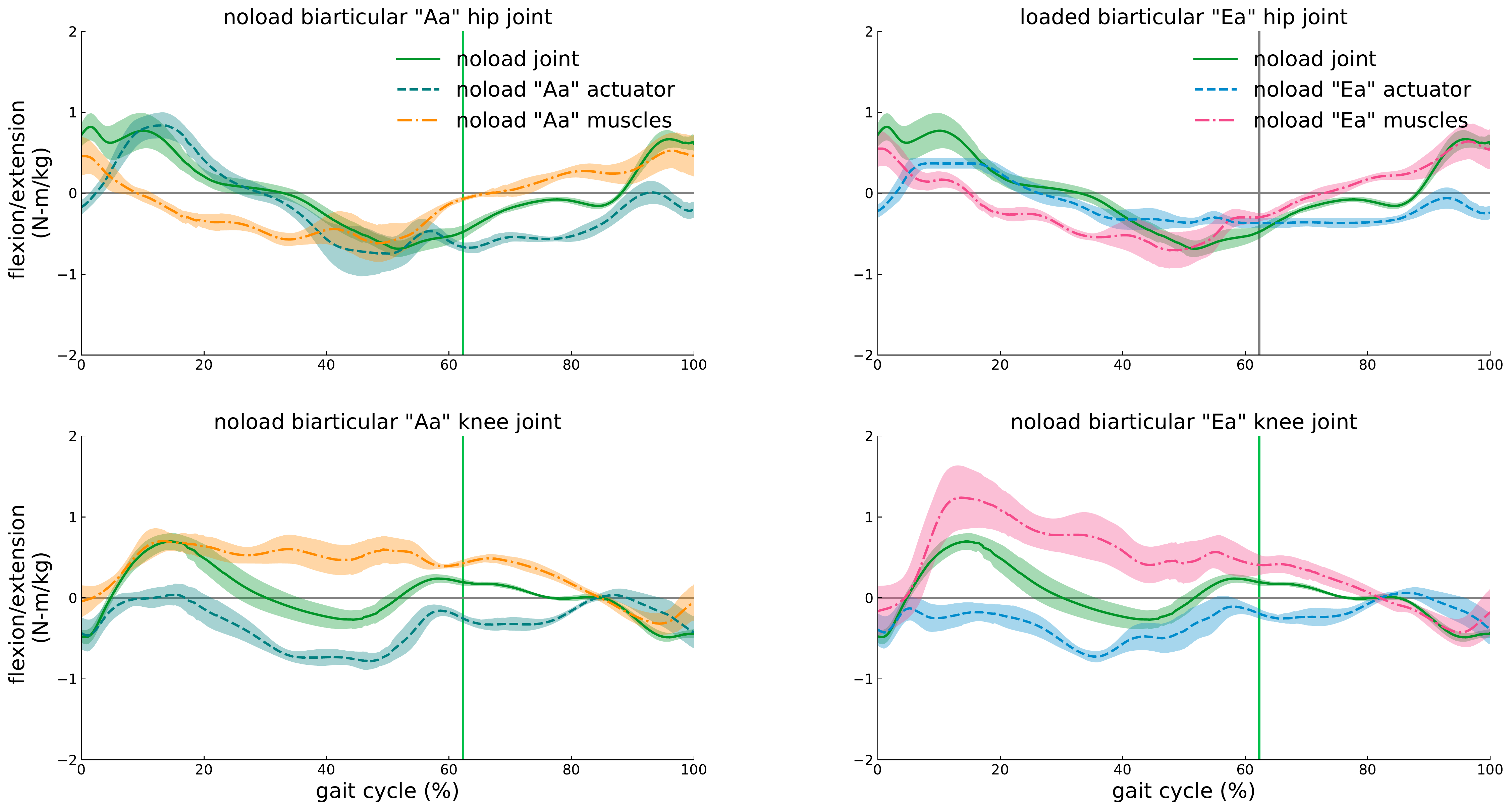}
	\vspace{1mm}
	\caption{\small{\textbf{Torque profiles of biarticular "Aa" and "Ea" exoskeletons in {\it noload} condition.}}.}
	\label{Fig_TorqueProfiles_Biarticular}
\end{figure*}
\begin{figure*}[ht]   
	\centering
	\includegraphics[width=\linewidth]{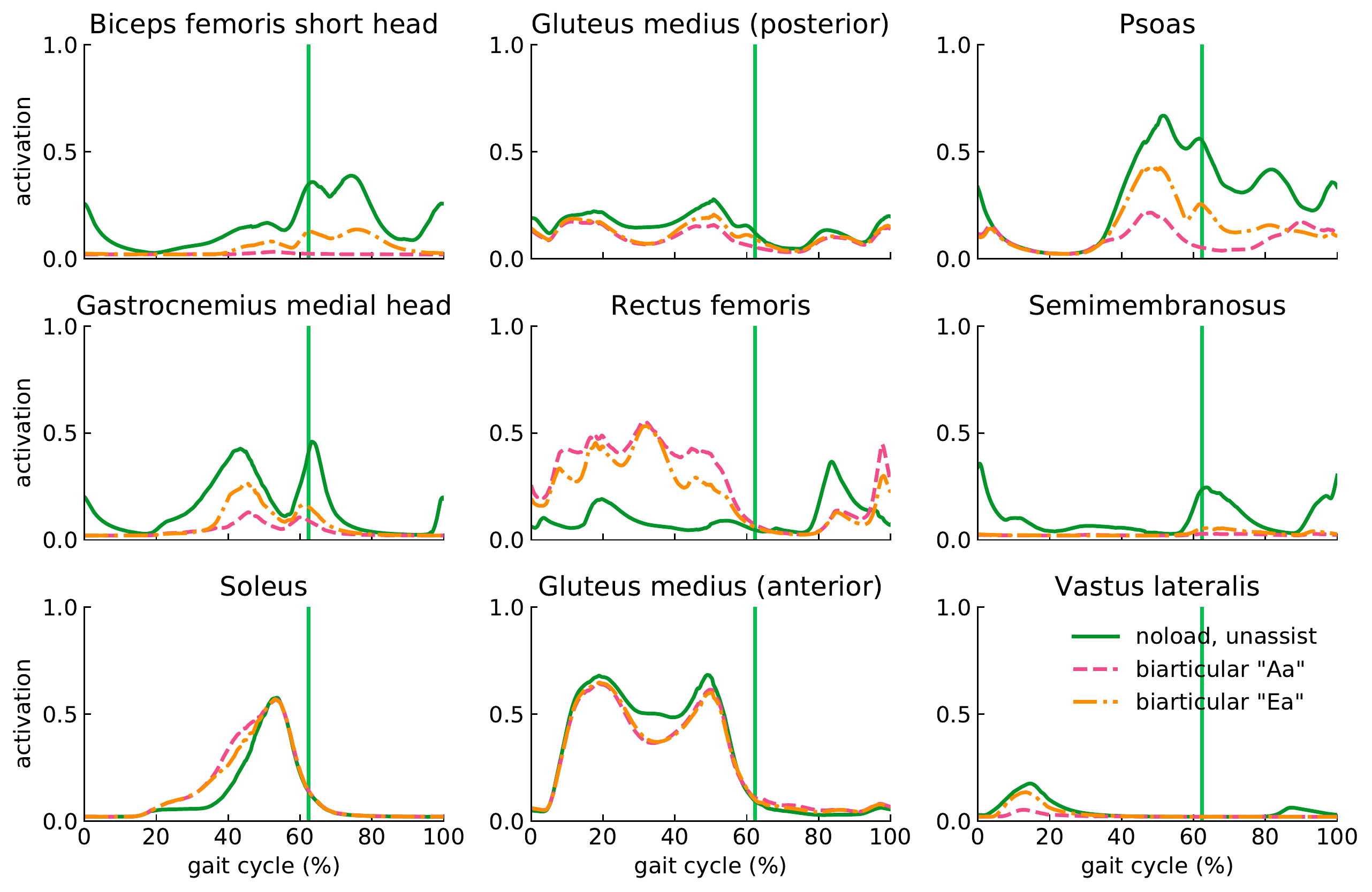}
	\vspace{1mm}
	\caption{\small{\textbf{Muscle activation of nine representative muscles of subjects assisted by biarticular "Aa" and "Ea" exoskeletons in {\it loaded} condition.  }}}
	\label{Fig_MuscleActivation_Biarticular}
\end{figure*}
\begin{figure*}[ht]   
	\centering
	\includegraphics[width=\linewidth]{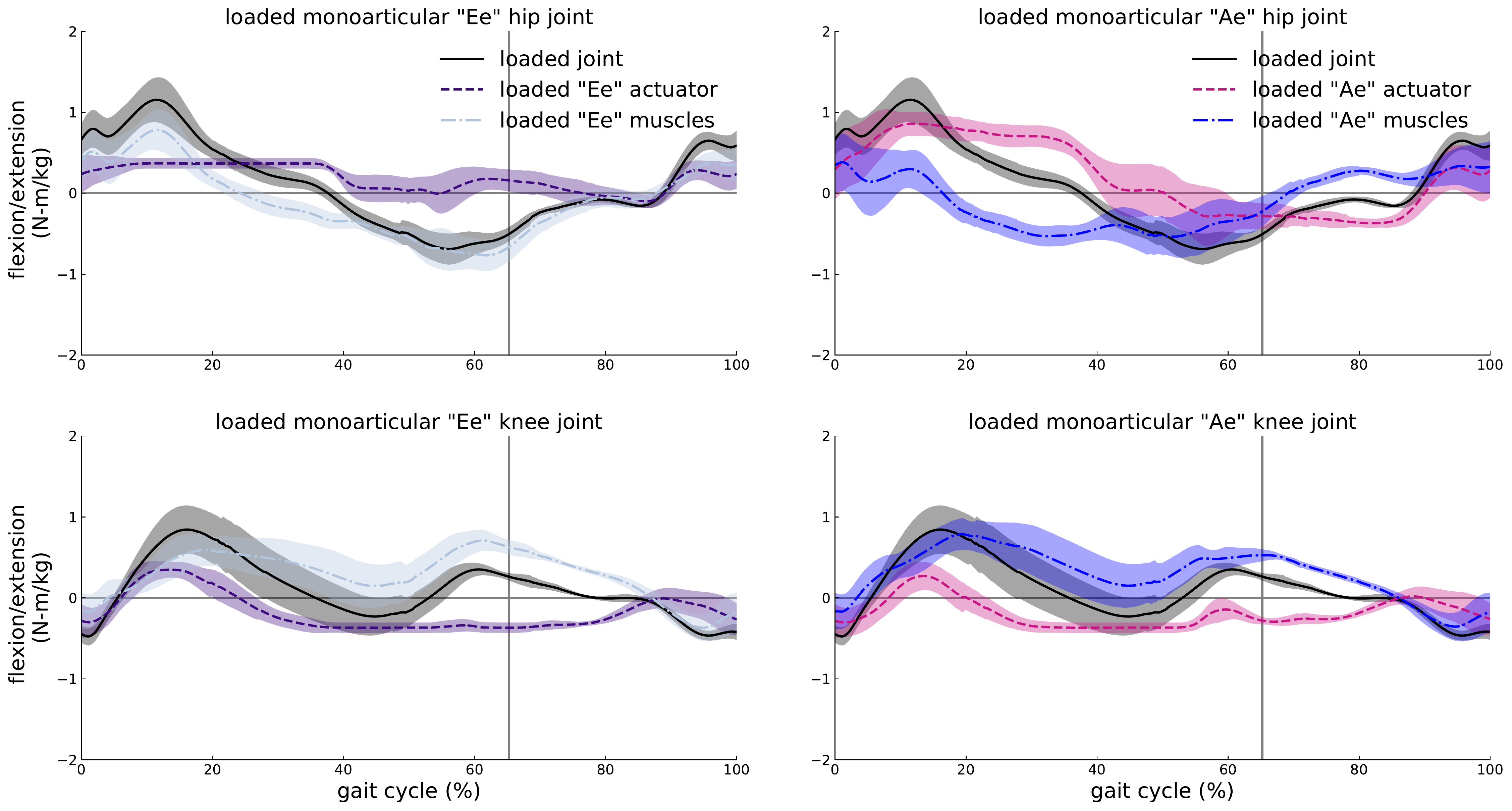}
	\vspace{1mm}
	\caption{\small{\textbf{Torque profiles of monoarticular "Ae" and "Ee" exoskeletons in {\it noload} condition.}}}
	\label{Fig_TorqueProfiles_Monoarticular}
\end{figure*}
\begin{figure*}[ht]   
	\centering
	\includegraphics[width=\linewidth]{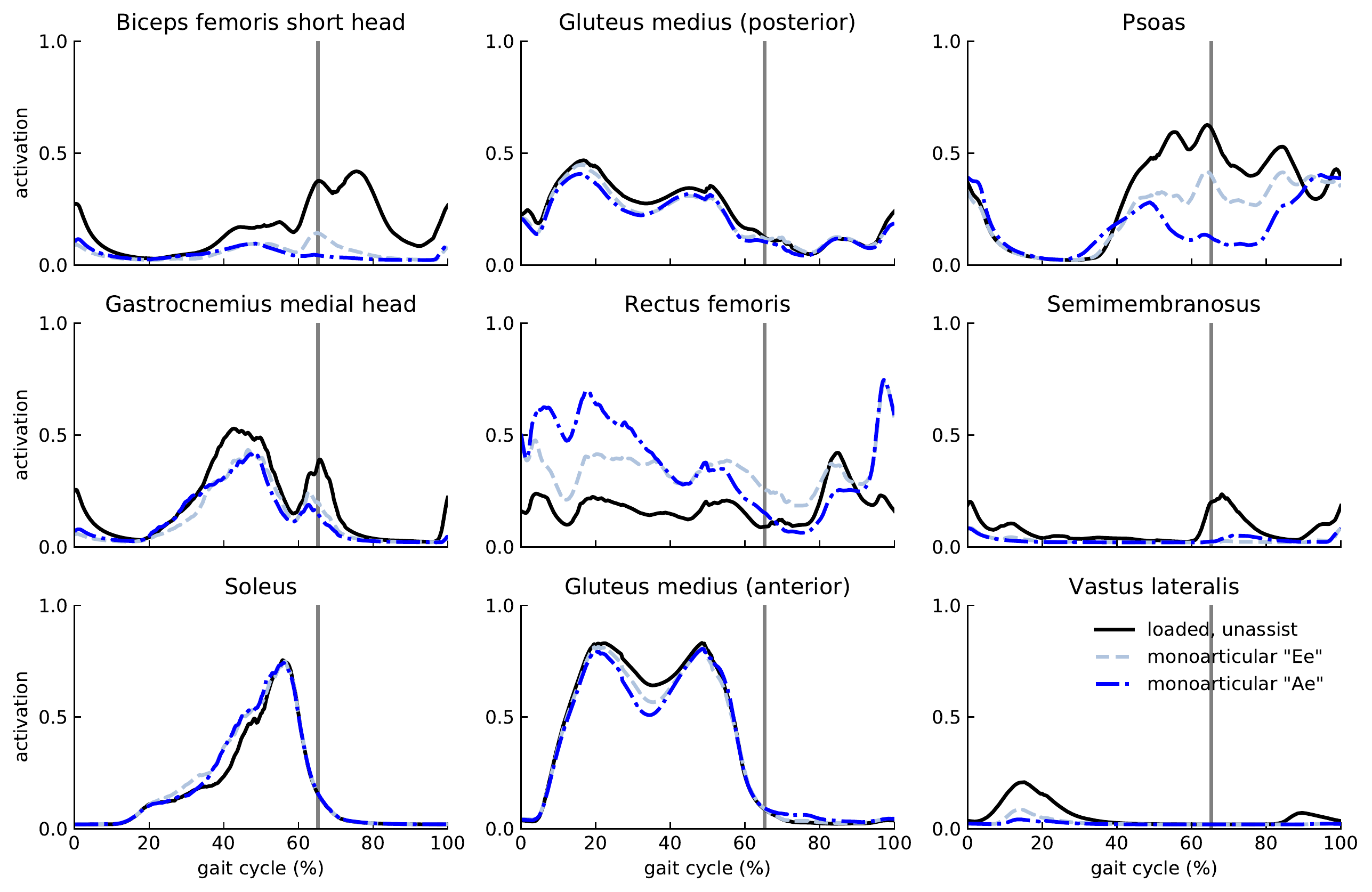}
	\vspace{1mm}
	\caption{\small{\textbf{Muscle activation of nine representative muscles of subjects assisted by monoarticular "Aa" and "Ea" exoskeletons in {\it loaded} condition.  }}}
	\label{Fig_MuscleActivation_Monoarticular}
\end{figure*}
\newpage
\clearpage
%
%
%
\bibliographystyle{plos2015.bst}
\bibliography{Bibliography_v3}

\end{document}